\documentclass{scrartcl}
\usepackage{natbib}
\usepackage{enumitem}
\usepackage{mathtools}

\usepackage{tikz}
\usepackage{tikz-3dplot}
\usetikzlibrary{decorations.pathreplacing}

\usepackage{tikz}
\usetikzlibrary{decorations.pathreplacing}
\usetikzlibrary{fadings}
\usetikzlibrary{shapes.geometric}
\usepackage{subcaption}

\usepackage{amsmath}

\usepackage{longtable}

\usepackage{amsfonts}
\usepackage{amssymb}
\usepackage{dsfont}
\usepackage[T1]{fontenc}
\usepackage[latin1]{inputenc}
\usepackage{epsfig}
\usepackage{graphicx}

\usepackage[boxed, lined]{algorithm2e}
\usepackage{float}
\usepackage{comment}

\newcommand{\E}{\mathbf{E}}

\newcommand{\bk}{\mathbf{k}}

\newcommand{\ba}{\mathbf{a}}
\newcommand{\bb}{\mathbf{b}}

\newcommand{\bd}{\mathbf{d}}

\newcommand{\bn}{\mathbf{n}}

\newcommand{\bs}{\mathbf{s}}
\newcommand{\bx}{\mathbf{x}}
\newcommand{\bX}{\mathbf{X}}

\newcommand{\bw}{\mathbf{w}}

\newcommand{\bz}{\mathbf{z}}
\newcommand{\bM}{\mathbf{M}}

\newcommand{\D}{{\mathcal{D}}}
\newcommand{\A}{{\mathcal{A}}}

\newcommand{\F}{{\mathcal{F}}}
\newcommand{\G}{{\mathcal{G}}}

\newcommand{\Nu}{{\mathcal{N}}}

\newcommand{\N}{\mathbb{N}}

\newcommand{\R}{\mathbb{R}}
\newcommand{\Z}{\mathbb{Z}}
\newcommand{\Rd}{\mathbb{R}^d}
\usepackage{bbm}
\newcommand{\IND}{\mathbbm{1}}
\usepackage{bm}
\newcommand{\balpha}{\pmb{\alpha}}
\newcommand{\btheta}{\pmb{\theta}}

\newcommand{\beq}{\begin{eqnarray*}}

\newcommand{\eeq}{\end{eqnarray*}}

\newcommand{\beqm}{\begin{eqnarray}}

\newcommand{\eeqm}{\end{eqnarray}}

\newtheorem{theorem}{Theorem}

\newtheorem{lemma}{Lemma}
\newtheorem{definition}{Definition}

\newtheorem{remark}{Remark}

\let\oldremark\remark
\renewcommand{\remark}{\oldremark\normalfont}

\DeclareMathOperator*{\argmin}{arg\,min}
\DeclareMathOperator{\sgn}{sgn}
\DeclareMathOperator{\VC}{VCdim}
\newcommand{\EXP}{{\mathbf E}}
\newcommand{\PROB}{{\mathbf P}}

\renewcommand{\bf}{\normalfont \bfseries}
\renewcommand{\it}{\normalfont \itshape}

\usepackage{natbib}

\allowdisplaybreaks
\begin{document}
\begin{center}
	
	{\LARGE \bf
Analysis of convolutional neural network image classifiers in a hierarchical max-pooling model with additional local pooling
	}
	\vspace{0.5cm}

 Benjamin Walter\footnote{\label{note1}Funded by the Deutsche Forschungsgemeinschaft (DFG, German
 	Research Foundation)  - Projektnummer 449102119.}
	\\

	{\it 
		Fachbereich Mathematik, Technische Universit\"at Darmstadt,
		Schlossgartenstr. 7, 64289 Darmstadt, Germany,
		email:
		bwalter@mathematik.tu-darmstadt.de}

\end{center}
\vspace{0.5cm}

\begin{center}
  \today
\end{center}
\vspace{0.5cm}

\noindent
{\bf Abstract}\\
Image classification is considered, and a hierarchical
max-pooling model with additional local pooling is
introduced. Here the additional local pooling enables the
hierachical model to combine parts of the image which have
a variable relative distance towards each other.
Various convolutional neural network image classifiers
are introduced and compared in view of their rate
of convergence. The finite sample size performance of the
estimates is analyzed by applying them to simulated and real
data.

\vspace*{0.2cm}


\vspace*{0.2cm}


\section{Introduction}
\label{se1}
\subsection{Deep convolutional networks}
Deep learning, i.e., estimation of a functional relationship
by a deep neural network, belongs nowadays to the most
successful
and most widely used methods in machine learning, see, e.g., \cite{Schmidhuber2015}
and the literature cited therein.
In many applications the most
successful networks are deep convolutional networks.
E.g., since 2012, the ImageNet Large Scale Visual Recognition
Challenge (ILSVRC) was won each year by deep convolutional neural
networks (cf., \cite{Russakovsky2015}).%

Deep convolutional neural networks can be considered
as a special case of deep feedforward neural networks, where
symmetry constraints are imposed on the weights of the networks. For deep feedforward neural networks, recently a large number
of impressive rate of convergence results have been shown.
E.g., in \cite{KoKr17}, \cite{Bauer2019}, \cite{SchmidtHieber2020}, \cite{KoLa2021}
and
\cite{Suzuki2019}
it was shown that these networks achieve in nonparametric regression
a dimension reduction
and are hence able to circumvent the so--called curse of
dimemensionality in case that the regression function can be written
as a composition of functions where each function depends only an
a few variables.

Recently, a similar result was shown in
\cite{KoKrWa2020}
for image classification by
deep
convolutional neural networks. The main idea
there was to assume that the a posteriori probability
satisfies a hierarchical max-pooling model (see Definition 1), where
a hierarchical model is defined which combines recursively
subparts of the image and where the a posteriori probability
is given by the maximum value which one obtains if one applies
this hierarchical model to all possible parts of the
image. This model mimics the approach of a human,
who classifies a
subpart of an image by combining hierarchically
several decisions concerning parts of this subpart of the image, 
by surveying all subparts of the whole image and
by estimating the probability of a class by the
maximum of the probability of all subparts.
Under this assumption it was shown that properly
defined deep convolutional neural networks, which
use several convolutional layers together with
a final max-pooling layer and which are defined
as a plug-in decision rule corresponding to a least
squares estimate of the a posteriori probability, achieve
in case of a smooth a posteriori probability a rate of
convergence which is independent of the dimension
of the image (and hence are able to circumvent the
curse of dimensionality).
In Kohler and Langer (2020)
it was shown that
a corresponding result also holds for
deep convolutional neural networks
defined by minimizing the cross-entropy loss.

In practice, more general network architectures are used. In addition to convolutional layers, these also contain so-called pooling layers (cf., e.g., Krizhevsky et al. (2012) and Simonyan and Zisserman (2014)). These layers reduce the output of a previous convolutional layer by summarizing local neighborhoods. This is done, for example, by taking the maximum, the average, or by passing only a single particular value from the local region. The goal of this article is to mathematically analyze the performance of such more general network architectures, i.e., we want to show that these network architectures also achieve a dimension reduction in an even more realistic model for image classification problems (compared to the above hierarchical max-pooling model).%
\subsection{Main results in this article}
In this paper we extend the above hierachical max-pooling
model (see Definition 1) such that it becomes more realistic for applications.
In the hierachical max-pooling model the relative distances between different
parts of some fixed level of the hierarchical model are fixed.
This is not the case in the real world, where we can decide
e.g. whether a part of an image contains a face
by dividing the part into four subparts and determine  whether
each of the subparts contains corresponding parts of a face (e.g. two eyes, a nose, two ears or a mouth). Here it is not
important that, for example, two eyes and a mouth have fixed distances between them,
instead, we can vary their positions locally without affecting our decision.
In order to include this in our model, we use
a kind of local max-pooling, where locally parts of the image are
combined by replacing the function values in a small neighborhood
by the maximal occuring value. In this way we reduce the resolution
of our image and the hierarchical model combines decisions on more
abstract levels. 
We introduce various topologies of convolutional
neural networks suitable for this more complex model, which
contain several convolutional and several local pooling layers
and a final max-pooling layer, and analyze
the rate of convergence of the  misclassification risk of the
corresponding plug-in image classification rule towards
the optimal  misclassification risk. Here we are able
to show that the rate of convergence does again not depend
on the dimension of the image. Surprisingly we are able to show the
same rate of convergence also for an image classifier based on
a convolutional neural network using only one subsampling layer.
These results provide a theoretical explanation of why the use of general convolutional neural network architectures (which contain some kind of local pooling layers) is reasonable in image classification problems.
 The finite sample size performance of the
estimates is analyzed by applying them to simulated and real
data.
\subsection{Discussion of related results}
Convolutional neural networks have achieved remarkable success in image recognition applications, see e.g. \cite{LeCun1998}, \cite{LeCun2015}, \cite{Krizhevsky2012}, \cite{Simonyan2014} and the literature cited therein.

Unfortunately, as mentioned in \cite{Rawat2017}, ``a theoretical justification for their successes is still lacking''. In fact, there are only a few papers that study convolutional neural networks from a theoretical perspective.
 Some papers, see, e.g., \cite{Oono2019} and the literature cited therein, use that suitably defined convolutional neural networks can mimic deep fully connected neural networks and therefore achieve similar rate of convergence results. However, this approach does not reveal situations in which convolutional neural networks are superior to fully connected feedforward neural networks, as is specifically the case in many image classification problems.
\cite{Lin2019} treat generalization bounds for convolutional neural networks. That the gradient descent finds the global minimum of the empirical risk with quadratic loss function is shown in \cite{Du2018} for various neural network architectures. The number of neurons per layer of the neural networks used there is at least polynomial in sample size, thus the neural networks are over-parametrized. As shown by a counterexample in \cite{KoKr2021}, overparameterized neural networks generally do not generalize well. Very interesting approximation results for deep convolutional neural networks were obtained by \cite{Yarotsky2018}. Since these results were achieved in an abstract setting, it is unclear how they can be applied.
\cite{Zhou2020} provides some further approximation results for convolutional neural networks.

As already described in more detail above, \cite{KoKrWa2020} obtained a rate of convergence result for convolutional neural networks in image classification problems. Assuming that the a posteriori probability satisfies a generalized hierarchical max-pooling model with smoothness constraints (see Definition 1 in \cite{KoKrWa2020}), they achieved a rate of convergence for suitably defined convolutional neural networks which is independent of the input image dimension. Here, the convolutional neural network plug-in classifier was defined by minimizing the empirical risk with the quadratic loss function. In \cite{KoLa2020}, a corresponding rate of convergence result was obtained by minimizing the cross-entropy loss. In both papers, the convolutional neural networks do not contain pooling layers. A theoretical analysis of local max-pooling and average-pooling layers was presented in \cite{Boureau2010}. Nevertheless, these results do not help to theoretically identify situations in which pooling layers are beneficial. 

For regression estimates based on standard deep feedforward neural networks, there are quite a few impressive rate of convergence results under compository assumptions on the structure of the regression function. Here it was shown that these estimates achieve a dimension reduction (cf., \cite{KoKr17}, \cite{Bauer2019}, \cite{SchmidtHieber2020}, \cite{KoLa2021} and \cite{Suzuki2019}).
Results concerning estimation by neural networks of piecewise smooth regression functions with partitions having rather general
smooth boundaries have been obtained by \cite{Imaizumi2019}.
\cite{Eckle2019} and \cite{KoKrLa2019} showed a connection of least squares regression estimates based on standard deep neural networks and the form of multivariate adaptive regression splines (MARS), where suitably defined standard deep neural networks can achieve a similar rate of convergence.

Convergence rate results concerning classification problems using standard deep neural networks were obtained by \cite{Kim2014} und \cite{Hu2020}.

\subsection{Notation}
Throughout the paper we use the following notation:
$\N$, $\N_0$, $\Z$, $\R$ and $\R_+$ are 
the sets of natural numbers, natural numbers including $0$,
integers, real numbers and nonnegative real numbers, respectively.
For $z \in \R$, we denote
the smallest integer greater than or equal to $z$ by
$\lceil z \rceil$.
Let $D \subseteq \R^d$ and let $f:\R^d \rightarrow \R$ be a real-valued
function defined on $\R^d$.
We write $\bx = \arg \min_{\bz \in D} f(\bz)$ if
$\min_{\bz \in \D} f(\bz)$ exists and if
$\bx$ satisfies
$\bx \in D$ and $f(\bx) = \min_{\bz \in \D} f(\bz)$.
For $f:\R^d \rightarrow \R$
\[
\|f\|_\infty = \sup_{\bx \in \R^d} |f(\bx)|
\]
is its supremum norm, and the supremum norm of $f$
on a set $A \subseteq \R^d$ is denoted by
\[
\|f\|_{A,\infty} = \sup_{\bx \in A} |f(\bx)|.
\]
Let $p=q+s$ for some $q \in \N_0$ and $0< s \leq 1$.
A function $f:\R^d \rightarrow \R$ is called
$(p,C)$-smooth, if for every $\balpha=(\alpha_1, \dots, \alpha_d) \in
\N_0^d$
with $\sum_{j=1}^d \alpha_j = q$ the partial derivative
$\frac{
	\partial^q f
}{
	\partial x_1^{\alpha_1}
	\dots
	\partial x_d^{\alpha_d}
}$
exists and satisfies
\[
\left|
\frac{
	\partial^q f
}{
	\partial x_1^{\alpha_1}
	\dots
	\partial x_d^{\alpha_d}
}
(\bx)
-
\frac{
	\partial^q f
}{
	\partial x_1^{\alpha_1}
	\dots
	\partial x_d^{\alpha_d}
}
(\bz)
\right|
\leq
C
\cdot
\| \bx-\bz \|^s
\]
for all $\bx,\bz \in \R^d$.

Let $\F$ be a set of functions $f:\Rd \rightarrow \R$,
let $\bx_1, \dots, \bx_n \in \Rd$ and set $\bx_1^n=(\bx_1,\dots,\bx_n)$.
A finite collection $f_1, \dots, f_N:\Rd \rightarrow \R$
is called an $\varepsilon$-- cover of $\F$ on $\bx_1^n$
if for any $f \in \F$ there exists  $i \in \{1, \dots, N\}$
such that
\[
\frac{1}{n} \sum_{k=1}^n |f(\bx_k)-f_i(\bx_k)| < \varepsilon.
\]
The $\varepsilon$--covering number of $\F$ on $\bx_1^n$
is the  size $N$ of the smallest $\varepsilon$--cover
of $\F$ on $\bx_1^n$ and is denoted by $\Nu_1(\varepsilon,\F,\bx_1^n)$.

For $z \in \R$ and $\beta>0$ we define
$T_\beta z = \max\{-\beta, \min\{\beta,z\}\}$. If $f:\R^d \rightarrow
\R$
is a function and $\F$ is a set of such functions, then we set
\[
(T_{\beta} f)(\bx)=
T_{\beta} \left( f(\bx) \right)
\quad \mbox{and} \quad
T_{\beta} \mathcal{F}
=
\left\{
T_{\beta} f
\quad : \quad
f \in \mathcal{F}
\right\}.
\]

Let $I$ be a nonempty and finite index set. For $A\subseteq\R$ and $\bx\in\R^d$ we use the notations
\[
A^{I}=\{(a_i)_{i\in I} : a_i\in A~(i\in I)\}.
\]
and
\[
\bx_I=(x_i)_{i\in I}.
\]
For $M\subset\R^d$ and $\bx\in\R^d$ we define
\[
\bx+M=\{\bx+\bz : \bz\in M\}.
\]
\subsection{Outline}
The outline of the paper is as follows:
In Section \ref{se2} the
hierarchical max-pooling model with additional local max-pooling
is introduced. The various convolutional neural networks
analyzed in this paper are described in Section \ref{se3}.
The main result is presented in Section \ref{se4} and proven
in the Supplement. Section \ref{se5} and Section \ref{se6} contain the application
of the estimates to simulated and real data.

\section{Hierarchical max-poling models for image classification}
\label{se2}
In this article we analyze image classification in the following statistical
setting:
Let
$d_1,d_2 \in \N$ and let
$(\bX,Y)$, $(\bX_1,Y_1)$, \dots, $(\bX_n,Y_n)$
be independent and identically distributed random variables
with values in
\[
[0,1]^{
\{1, \dots, d_1\} \times \{1, \dots, d_2\}
  } \times \{0,1\}.
\]
We describe a (random)
image from (random) class $Y \in \{0,1\}$ by a (random) matrix $X$
with $d_1$ columns and $d_2$ rows, which contains at position $(i,j)$
the grey scale value of the pixel of the
image at the corresponding position.

Set
\[
\D_n = \left\{
(\bX_1,Y_1), \dots, (\bX_n,Y_n)
\right\}.
\]
In the sequel we consider the problem of constructing
a classifier
\[
f_n=f_n(\cdot, \D_n):  [0,1]^{\{1, \dots, d_1\} \times \{1, \dots, d_2\}}
\rightarrow \{0,1\}
\]
such that the misclassification risk
\[
\PROB\{ f_n(\bX) \neq Y | \D_n\}
\]
of this classifier is as small as possible.

Let
\begin{equation}
\label{se1eq2}
\eta(\bx) = \PROB\{ Y=1|\bX=\bx\}
\quad
( \bx \in [0,1]^{
\{1, \dots, d_1\} \times \{1, \dots, d_2\}
  })
\end{equation}
be the so--called a posteriori probability of class 1. Then
we have
\[
\min_{f:
  [0,1]^{\{1, \dots, d_1\} \times \{1, \dots, d_2\}}
  \rightarrow \{0,1\}
}
  \PROB\{ f(\bX) \neq Y \}
  =
    \PROB\{ f^*(\bX) \neq Y \},
    \]
    where
\[
f^*(\bx)=
\begin{cases}
  1, & \mbox{if } \eta(\bx) > \frac{1}{2} \\
  0, & \mbox{elsewhere}
  \end{cases}
\]
is the so--called Bayes classifier
(cf., e.g., Theorem 2.1 in \cite{Devroye1996}).
We will use plug-in classifiers of the form
\[
f_n(\bx)=
\begin{cases}
  1, & \mbox{if } \eta_n(\bx) \geq \frac{1}{2} \\
  0, & \mbox{elsewhere}
  \end{cases}
\]
where
\[
\eta_n(\cdot)=\eta_n(\cdot,\D_n):
[0,1]^{\{1, \dots, d_1\} \times \{1, \dots, d_2\}}
\rightarrow \R
\]
is an estimate of the a posteriori probability (\ref{se1eq2}).

Our aim is to derive a bound on the expected difference of
the misclassification risk of $f_n$ and the optimal
misclassification risk, i.e., we want to derive an upper bound on
\begin{eqnarray*}
  &&
\EXP\left\{
\PROB\{ f_n(\bX) \neq Y | \D_n\}
-
\min_{f:
  [0,1]^{\{1, \dots, d_1\} \times \{1, \dots, d_2\}}
  \rightarrow \{0,1\}
}
\PROB\{ f(\bX) \neq Y \}
\right\}
\\
&&
  =
\PROB\{ f_n(\bX) \neq Y \}
-
    \PROB\{ f^*(\bX) \neq Y \}.
\end{eqnarray*}

In Kohler, Krzy\.zak and Walter (2020)
the following model for the a posteriori probability was used to
derive an upper bound on the above difference.

\begin{definition}
  \label{de1}
{\bf a)}
We say that
$m: [0,1]^{\{1, \dots, d_1\} \times \{1, \dots, d_2\}} \rightarrow \R$
satisfies a {\bf max-pooling model with index set}
\[
I \subseteq \{0, \dots, d_1-1\} \times \{0, \dots, d_2-1\},
\]
if there exist a function $f:[0,1]^{(1,1)+I} \rightarrow \R$ such that
\[
m(\bx)=
\max_{
  (i,j) \in \Z^2 \, : \,
  (i,j)+I \subseteq \{1, \dots, d_1\} \times \{1, \dots, d_2\}
}
f\left(
x_{(i,j)+I}
\right)
\quad
(x \in [0,1]^{\{1, \dots, d_1\} \times \{1,
  \dots, d_2\}}).
\]

\noindent
    {\bf b)}
    Let $I=\{0, \dots, 2^l-1\} \times \{0, \dots, 2^l-1\}$
    for some $l \in \N_0$.
    We say that
\[
f:[0,1]^{\{1, \dots, 2^l\} \times \{1, \dots, 2^l\}} \rightarrow \R
\]
 satisfies a
    {\bf hierarchical model of level $l$},
    if there exist functions
    \[
    g_{k,s}: \R^4 \rightarrow [0,1]
    \quad (k=1, \dots, l, s=1, \dots, 4^{l-k} )
    \]
    such that we have
    \[
f=f_{l,1}
    \]
    for some
    $f_{k,s} :[0,1]^{\{1, \dots, 2^k\} \times \{1, \dots, 2^k\}} \rightarrow \R$ recursively defined by
    \begin{eqnarray*}
    f_{k,s}(x)&=&g_{k,s} \big(
    f_{k-1,4 \cdot (s-1)+1}(x_{
\{1, \dots, 2^{k-1}\} \times \{1, \dots, 2^{k-1}\}
    })
    , \\
        &&
        \hspace*{1cm}
        f_{k-1,4 \cdot (s-1)+2}(x_{
\{2^{k-1}+1, \dots, 2^k\} \times \{1, \dots, 2^{k-1}\}
        }), \\
        &&
        \hspace*{1cm}
        f_{k-1,4 \cdot (s-1)+3}(x_{
\{1, \dots, 2^{k-1}\} \times \{2^{k-1}+1, \dots, 2^k\}
        }), \\
        &&
        \hspace*{1cm}
        f_{k-1,4 \cdot s}(x_{
\{2^{k-1}+1, \dots, 2^k\} \times \{2^{k-1}+1, \dots, 2^k\}
        })
    \big)
\\
&&
\hspace*{6cm}
\left(
x \in
[0,1]^{
\{
1, \dots, 2^k
\}
\times
\{ 1, \dots, 2^k
\}
}
\right)
    \end{eqnarray*}
    for $k=2, \dots, l, s=1, \dots,4^{l-k}$,
    and
    \[
 f_{1,s}(
x_{1,1},x_{1,2},x_{2,1},x_{2,2}
)= g_{1,s}(x_{1,1},x_{1,2},x_{2,1},x_{2,2})
\quad
( x_{1,1},x_{1,2},x_{2,1},x_{2,2} \in [0,1])
 \]
 for $s=1, \dots, 4^{l-1}$.

 \noindent
     {\bf c)}
     We say that
     $m: [0,1]^{\{1, \dots, d_1\} \times \{1, \dots, d_2\}} \rightarrow \R$
     satisfies a {\bf hierarchical max-pooling model of level $l$
      } (where $2^l \leq \min\{ d_1,d_2\}$),
     if $m$ satisfies a max-pooling model with index set
     \[
I=\{0, \dots, 2^l-1\} \times \{0, \dots 2^l-1\}
     \]
     and the function
     $f:[0,1]^{ \{ 1, \dots, 2^l \} \times \{1, \dots, 2^l\}} \rightarrow \R$ in the definition of this
     max-pooling model satisfies a hierarchical model
     with level $l$.

    \noindent
     {\bf d)}
Let $p=q+s$ for some $q \in \N_0$ and $s \in (0,1]$, and let $C>0$.
     We say that a  hierarchical max-pooling model is $(p,C)$--smooth
     if all functions $g_{k,s}$ in its definition are $(p,C)$--smooth.
  \end{definition}
\noindent
{\bf Remark 1.}
In Kohler, Krzy\.zak, and Walter (2020), they also introduced a generalization of the hierarchical max-pooling model so that it includes functions from several such models. In order to simplify our notation, we restrict ourselves here to the non-generalized definition of the hierarchical max-pooling model. However, our results can easily be extended to this generalization. \\

We want to extend the above model by some further structural assumptions. To motivate these structural assumptions, we first put our model in a new abstract framework. As in Definition 1, we assume that the decision about the class of an image is made hierarchically. That is, decisions about neighboring small subparts of the image are combined into decisions about larger subparts of the image. In the above setting, the function $f_{k,s}:[0,1]^{\{1, \dots, 2^k\} \times \{1, \dots, 2^k\}} \rightarrow[0,1]$ makes such a decision for a $2^k\times2^k$-sized rectangular subpart of the input image by combining decisions from smaller parts of the subpart. If we now consider the function $f_{k,s}$ for all possible subparts of the input image $\bx\in[0,1]^{\{1,\dots,d_1\}\times\{1,\dots,d_2\}}$, the function yields us a new representation of the input image, which is given by a function
\[y_{k,s}:[0,1]^{\{1,\dots,d_1\}\times\{1,\dots,d_2\}}\rightarrow[0,1]^{\{1,\dots,d_1(k)\}\times\{1,\dots,d_2(k)\}}\]
with 
\begin{equation}
	\label{fmap}
	y_{k,s}(\bx)=\big(f_{k,s}(\bx_{\{i,\dots,i+2^k-1\}\times\{j,\dots,j+2^k-1\}})\big)_{(i,j)\in\{1,\dots,d_1(k)\}\times\{1,\dots,d_2(k)\}}
\end{equation}
for $\bx\in[0,1]^{\{1,\dots,d_1\}\times\{1,\dots,d_2\}}$ and dimensions 
\begin{equation}
	\label{dim1}
	d_1(k)=d_1-2^k+1\mbox{ and }d_2(k)=d_2-2^k+1.
\end{equation}
The choice \eqref{dim1} of the dimensions $d_1(k)$ and $d_2(k)$ ensures that all subparts are considered that are entirely contained in the input image.
We call the representation \eqref{fmap} of the input image {\bf feature map} of level $k\in\{1,\dots,l\}$ of the feature $s\in\{1,\dots,b_k\}$, where we assume that the number of possible features in each level is bounded by $b_k\in\N$.
Here neighboring parts of a feature map correspond to neighboring parts of the input image and in each level each feature map describes whether locally a special kind
of object is contained in the image or not.
We can then rewrite and slightly generalize parts a)--c) of Definition 1 as follows:
\begin{definition}
	\label{de1}
	Let $d_1,d_2,l\in\N$ and let $b_0,\dots,b_{l}\in\N$ with $b_0=b_l=1$ and set $\bb=(b_1,\dots,b_{l-1})$.

\noindent
{\bf a)}
For $k=1,\dots,l$ we recursively define the dimensions
\begin{equation}
	\label{d1eq1}
	d_1(k)= d_1(k-1)-2^{k-1}
	\mbox{ and }
	d_2(k)=	d_2(k-1)-2^{k-1},
\end{equation}
where we set $d_1(0)=d_1$ and $d_2(0)=d_2$. 
We say that 
\[y:[0,1]^{\{1, \dots, d_1\} \times \{1, \dots, d_2\}}\rightarrow\R^{\{1,\dots,d_1(l)\}\times\{1,\dots,d_2(l)\}}\]
is a {\bf feature map} \textbf{of level} $l$ {\bf with feature constraint} $\bb$,
if there exist functions
\[g_{k,s}:\R^4\rightarrow[0,1]\quad(k=1,\dots,l,s=1,\dots,b_k)\]
such that we have
\[y=y_{l,1}\]
for some $y_{k,s}:[0,1]^{\{1,\dots,d_1\}\times\{1,\dots,d_2\}}\rightarrow[0,1]^{\{1,\dots,d_1(k)\}\times\{1,\dots,d_2(k)\}}$ recursively defined as follows:
\begin{enumerate}
	\item 
	We set
	\[
	y_{0,1}(\bx)=\bx.
	\]
	\item
	We define recursively feature maps 
	\[
	y_{k,s}:[0,1]^{\{1,\dots,d_1\}\times\{1,\dots,d_2\}}\rightarrow[0,1]^{\{1,\dots,d_1(k)\}\times\{1,\dots,d_2(k)\}}
	\]
	of level $k$ by
	\begin{align*}
		&\big(y_{k,s}(\bx)\big)_{(i,j)}=g_{k,s}\Big(
		\big(y_{k-1,r_1(k,s)}(\bx)\big)_{(i,j)},
		\big(y_{k-1,r_2(k,s)}(\bx)\big)_{(i+2^{k-1},j)},
		\\&\hspace{3.5cm}
		\big(y_{k-1,r_3(k,s)}(\bx)\big)_{(i,j+2^{k-1})},\big(y_{k-1,r_4(k,s)}(\bx)\big)_{(i+2^{k-1},j+2^{k-1})}\Big)
	\end{align*}
	for $k=1,\dots,l$, $s=1,\dots,b_k$, $(i,j)\in\{1,\dots,d_1(k)\}\times\{1,\dots,d_2(k)\}$, 
	$r_1(k,s)$, $r_2(k,s)$, $r_3(k,s)$, $r_4(k,s)\in\{1,\dots,b_{k-1}\}.$
\end{enumerate}

\noindent
{\bf b)}
We say that $m: [0,1]^{\{1, \dots, d_1\} \times \{1, \dots, d_2\}} \rightarrow[0,1]$ satisfies a \textbf{hierarchical max-pooling model of level} $l$ {\bf with feature constraint} $\bb$, if there exist a function
\[y:[0,1]^{\{1,\dots,d_1\}\times\{1,\dots,d_2\}}\rightarrow[0,1]^{\{1,\dots,d_1(l)\}\times\{1,\dots,d_2(l)\}}\]
which is a feature map of level $l$ with feature constraint $\bb$
such that
\[m(\bx)=
\max_{
	(i,j)\in\{1,\dots,d_1(l)\}\times\{1,\dots,d_2(l)\}}
\big(y(\bx)\big)_{(i,j)}.\]

\noindent
	
\end{definition}
\noindent
{\bf Remark 2.}
	The above model is equivalent to the hierarchical max-pooling model of level $l$ from definition 1 in case we choose $r_i(k,s)=4\cdot(s-1)+i$ for $i=1,\dots,4$ and $b_k=4^{l-k}$ for $k\in\{1,\dots,l\}$.

\noindent
{\bf Remark 3.}
	We have stated a recursive definition \eqref{d1eq1} of the dimensions \eqref{dim1}, because the connection with the following definition then becomes more obvious.\\

Now we can add another structural assumption to our model. For this purpose, consider an example in which a human is supposed to decide whether a given image contains a face or not. The human then surveys the image to see if it contains two eyes, a mouth, and a nose whose positions are approximately in a certain relationship to each other. The exact positions of these objects in relation to each other are not crucial. Therefore, we can assume that we can summarize local neighborhoods of a feature map into the information whether the whole local neighborhood contains the corresponding object. 
We do this by replacing the values  of disjoint rectangular local neighborhoods of a feature map with their maximum occuring values.
 
For a given feature map $y_{k,s}$ of level $k$ with dimensions
$d_1(k)$ and $d_2(k)$ we define rectangular neighborhoods of size $n_k\in\N$ as follows:

For $(i,j)\in\{1,\dots,\left\lceil d_1(k)/n_k\right\rceil\}\times\{1,\dots,\left\lceil d_2(k)/n_k\right\rceil\}$ we define neighborhoods
\begin{equation}
	\label{NeighbH}
	\begin{split}
		N^{(k)}_{(i,j)}&=\Big(\{(i-1)\cdot n_k+1,\dots,i\cdot n_k\}\times\{(j-1)\cdot n_k+1,\dots,j\cdot n_k\}\Big)\\
		&\hspace{0.5cm}\cap\Big(\{1,\dots,d_1(k)\}\times\{1,\dots,d_2(k)\}\Big)
	\end{split}
\end{equation}
and for the given $y_{k,s}$ we define a {\bf feature map with local max-pooling} of level $k$ by a function 
\[z_{k,s}:[0,1]^{\{1,\dots,d_1\}\times\{1,\dots,d_2\}}\rightarrow[0,1]^{\left\{1,\dots,\left\lceil\frac{d_1(k)}{n_k}\right\rceil\right\}\times\left\{1,\dots,\left\lceil\frac{d_2(k)}{n_k}\right\rceil\right\}},\]
which satisfies
\[
\big(z_{k,s}(\bx)\big)_{(i,j)}=
\max_{(i_2,j_2)\in N^{(k)}_{(i,j)}}
\big(y_{k,s}(\bx)\big)_{(i_2,j_2)}
\]
for $(i,j)\in\{1,\dots,\lceil d_1(k)/n_k\rceil\}\times\{1,\dots,\lceil d_2(k)/n_k\rceil\}$.
\begin{figure}[H]
	\label{fig2}
	\centering
	\begin{tikzpicture}[decoration={brace,mirror,amplitude=7},scale=0.45]	
		\foreach \i in{0,2,4,6,8,10,12}
		\foreach \j in{2,4,6,8,10}
		{{
				\fill [color=gray,opacity=0.2] (\i+0.15,\j+0.15) rectangle (\i+2-0.15,\j+2-0.15);
				\draw [color=black,opacity=0.5] (\i+0.15,\j+0.15) rectangle (\i+2-0.15,\j+2-0.15);
		}}
		\foreach \c in {7}{
			\draw [decorate,black,very thick] (17,5.8-\c) --node[below=6]{$z_{k,s}(\bx)$} (27,5.8-\c);}
		\draw [decorate,black,very thick] (0.2,0.8) --node[below=6]{$y_{k,s}(\bx)$} (13.8,0.8);
		\fill [color=black,opacity=0.48] (0+0.25,2+0.25) rectangle (0+0.75,2+0.75);
		\fill [color=black,opacity=0.36] (0+0.25,3+0.25) rectangle (0+0.75,3+0.75);
		\fill [color=black,opacity=0.11] (0+0.25,4+0.25) rectangle (0+0.75,4+0.75);
		\fill [color=black,opacity=0.87] (0+0.25,5+0.25) rectangle (0+0.75,5+0.75);
		\fill [color=black,opacity=0.47] (0+0.25,6+0.25) rectangle (0+0.75,6+0.75);
		\fill [color=black,opacity=0.03] (0+0.25,7+0.25) rectangle (0+0.75,7+0.75);
		\fill [color=black,opacity=0.89] (0+0.25,8+0.25) rectangle (0+0.75,8+0.75);
		\fill [color=black,opacity=0.64] (0+0.25,9+0.25) rectangle (0+0.75,9+0.75);
		\fill [color=black,opacity=0.64] (0+0.25,10+0.25) rectangle (0+0.75,10+0.75);
		\fill [color=black,opacity=0.43] (0+0.25,11+0.25) rectangle (0+0.75,11+0.75);
		\fill [color=black,opacity=0.9] (1+0.25,2+0.25) rectangle (1+0.75,2+0.75);
		\fill [color=black,opacity=0.01] (1+0.25,3+0.25) rectangle (1+0.75,3+0.75);
		\fill [color=black,opacity=0.01] (1+0.25,4+0.25) rectangle (1+0.75,4+0.75);
		\fill [color=black,opacity=0.27] (1+0.25,5+0.25) rectangle (1+0.75,5+0.75);
		\fill [color=black,opacity=0.2] (1+0.25,6+0.25) rectangle (1+0.75,6+0.75);
		\fill [color=black,opacity=0.57] (1+0.25,7+0.25) rectangle (1+0.75,7+0.75);
		\fill [color=black,opacity=0.75] (1+0.25,8+0.25) rectangle (1+0.75,8+0.75);
		\fill [color=black,opacity=0.11] (1+0.25,9+0.25) rectangle (1+0.75,9+0.75);
		\fill [color=black,opacity=0.07] (1+0.25,10+0.25) rectangle (1+0.75,10+0.75);
		\fill [color=black,opacity=0.56] (1+0.25,11+0.25) rectangle (1+0.75,11+0.75);
		\fill [color=black,opacity=0.67] (2+0.25,2+0.25) rectangle (2+0.75,2+0.75);
		\fill [color=black,opacity=0.94] (2+0.25,3+0.25) rectangle (2+0.75,3+0.75);
		\fill [color=black,opacity=0.41] (2+0.25,4+0.25) rectangle (2+0.75,4+0.75);
		\fill [color=black,opacity=0.66] (2+0.25,5+0.25) rectangle (2+0.75,5+0.75);
		\fill [color=black,opacity=0.56] (2+0.25,6+0.25) rectangle (2+0.75,6+0.75);
		\fill [color=black,opacity=0.45] (2+0.25,7+0.25) rectangle (2+0.75,7+0.75);
		\fill [color=black,opacity=0.96] (2+0.25,8+0.25) rectangle (2+0.75,8+0.75);
		\fill [color=black,opacity=0.68] (2+0.25,9+0.25) rectangle (2+0.75,9+0.75);
		\fill [color=black,opacity=0.35] (2+0.25,10+0.25) rectangle (2+0.75,10+0.75);
		\fill [color=black,opacity=0.07] (2+0.25,11+0.25) rectangle (2+0.75,11+0.75);
		\fill [color=black,opacity=0.01] (3+0.25,2+0.25) rectangle (3+0.75,2+0.75);
		\fill [color=black,opacity=0.94] (3+0.25,3+0.25) rectangle (3+0.75,3+0.75);
		\fill [color=black,opacity=0.88] (3+0.25,4+0.25) rectangle (3+0.75,4+0.75);
		\fill [color=black,opacity=0.71] (3+0.25,5+0.25) rectangle (3+0.75,5+0.75);
		\fill [color=black,opacity=0.04] (3+0.25,6+0.25) rectangle (3+0.75,6+0.75);
		\fill [color=black,opacity=0.38] (3+0.25,7+0.25) rectangle (3+0.75,7+0.75);
		\fill [color=black,opacity=0.31] (3+0.25,8+0.25) rectangle (3+0.75,8+0.75);
		\fill [color=black,opacity=0.67] (3+0.25,9+0.25) rectangle (3+0.75,9+0.75);
		\fill [color=black,opacity=0.78] (3+0.25,10+0.25) rectangle (3+0.75,10+0.75);
		\fill [color=black,opacity=0.96] (3+0.25,11+0.25) rectangle (3+0.75,11+0.75);
		\fill [color=black,opacity=0.21] (4+0.25,2+0.25) rectangle (4+0.75,2+0.75);
		\fill [color=black,opacity=0.42] (4+0.25,3+0.25) rectangle (4+0.75,3+0.75);
		\fill [color=black,opacity=0.5] (4+0.25,4+0.25) rectangle (4+0.75,4+0.75);
		\fill [color=black,opacity=0.11] (4+0.25,5+0.25) rectangle (4+0.75,5+0.75);
		\fill [color=black,opacity=0.7] (4+0.25,6+0.25) rectangle (4+0.75,6+0.75);
		\fill [color=black,opacity=0.98] (4+0.25,7+0.25) rectangle (4+0.75,7+0.75);
		\fill [color=black,opacity=0.72] (4+0.25,8+0.25) rectangle (4+0.75,8+0.75);
		\fill [color=black,opacity=0.67] (4+0.25,9+0.25) rectangle (4+0.75,9+0.75);
		\fill [color=black,opacity=0.51] (4+0.25,10+0.25) rectangle (4+0.75,10+0.75);
		\fill [color=black,opacity=1.0] (4+0.25,11+0.25) rectangle (4+0.75,11+0.75);
		\fill [color=black,opacity=0.92] (5+0.25,2+0.25) rectangle (5+0.75,2+0.75);
		\fill [color=black,opacity=0.67] (5+0.25,3+0.25) rectangle (5+0.75,3+0.75);
		\fill [color=black,opacity=0.92] (5+0.25,4+0.25) rectangle (5+0.75,4+0.75);
		\fill [color=black,opacity=0.13] (5+0.25,5+0.25) rectangle (5+0.75,5+0.75);
		\fill [color=black,opacity=0.25] (5+0.25,6+0.25) rectangle (5+0.75,6+0.75);
		\fill [color=black,opacity=0.9] (5+0.25,7+0.25) rectangle (5+0.75,7+0.75);
		\fill [color=black,opacity=0.3] (5+0.25,8+0.25) rectangle (5+0.75,8+0.75);
		\fill [color=black,opacity=0.13] (5+0.25,9+0.25) rectangle (5+0.75,9+0.75);
		\fill [color=black,opacity=0.57] (5+0.25,10+0.25) rectangle (5+0.75,10+0.75);
		\fill [color=black,opacity=0.73] (5+0.25,11+0.25) rectangle (5+0.75,11+0.75);
		\fill [color=black,opacity=0.56] (6+0.25,2+0.25) rectangle (6+0.75,2+0.75);
		\fill [color=black,opacity=0.66] (6+0.25,3+0.25) rectangle (6+0.75,3+0.75);
		\fill [color=black,opacity=0.45] (6+0.25,4+0.25) rectangle (6+0.75,4+0.75);
		\fill [color=black,opacity=0.66] (6+0.25,5+0.25) rectangle (6+0.75,5+0.75);
		\fill [color=black,opacity=0.2] (6+0.25,6+0.25) rectangle (6+0.75,6+0.75);
		\fill [color=black,opacity=0.57] (6+0.25,7+0.25) rectangle (6+0.75,7+0.75);
		\fill [color=black,opacity=0.28] (6+0.25,8+0.25) rectangle (6+0.75,8+0.75);
		\fill [color=black,opacity=0.38] (6+0.25,9+0.25) rectangle (6+0.75,9+0.75);
		\fill [color=black,opacity=0.78] (6+0.25,10+0.25) rectangle (6+0.75,10+0.75);
		\fill [color=black,opacity=0.55] (6+0.25,11+0.25) rectangle (6+0.75,11+0.75);
		\fill [color=black,opacity=0.27] (7+0.25,2+0.25) rectangle (7+0.75,2+0.75);
		\fill [color=black,opacity=0.07] (7+0.25,3+0.25) rectangle (7+0.75,3+0.75);
		\fill [color=black,opacity=0.61] (7+0.25,4+0.25) rectangle (7+0.75,4+0.75);
		\fill [color=black,opacity=0.19] (7+0.25,5+0.25) rectangle (7+0.75,5+0.75);
		\fill [color=black,opacity=0.34] (7+0.25,6+0.25) rectangle (7+0.75,6+0.75);
		\fill [color=black,opacity=0.82] (7+0.25,7+0.25) rectangle (7+0.75,7+0.75);
		\fill [color=black,opacity=0.71] (7+0.25,8+0.25) rectangle (7+0.75,8+0.75);
		\fill [color=black,opacity=0.54] (7+0.25,9+0.25) rectangle (7+0.75,9+0.75);
		\fill [color=black,opacity=0.27] (7+0.25,10+0.25) rectangle (7+0.75,10+0.75);
		\fill [color=black,opacity=0.98] (7+0.25,11+0.25) rectangle (7+0.75,11+0.75);
		\fill [color=black,opacity=0.9] (8+0.25,2+0.25) rectangle (8+0.75,2+0.75);
		\fill [color=black,opacity=0.25] (8+0.25,3+0.25) rectangle (8+0.75,3+0.75);
		\fill [color=black,opacity=0.38] (8+0.25,4+0.25) rectangle (8+0.75,4+0.75);
		\fill [color=black,opacity=0.19] (8+0.25,5+0.25) rectangle (8+0.75,5+0.75);
		\fill [color=black,opacity=0.71] (8+0.25,6+0.25) rectangle (8+0.75,6+0.75);
		\fill [color=black,opacity=0.89] (8+0.25,7+0.25) rectangle (8+0.75,7+0.75);
		\fill [color=black,opacity=0.52] (8+0.25,8+0.25) rectangle (8+0.75,8+0.75);
		\fill [color=black,opacity=0.55] (8+0.25,9+0.25) rectangle (8+0.75,9+0.75);
		\fill [color=black,opacity=0.32] (8+0.25,10+0.25) rectangle (8+0.75,10+0.75);
		\fill [color=black,opacity=0.7] (8+0.25,11+0.25) rectangle (8+0.75,11+0.75);
		\fill [color=black,opacity=0.05] (9+0.25,2+0.25) rectangle (9+0.75,2+0.75);
		\fill [color=black,opacity=0.98] (9+0.25,3+0.25) rectangle (9+0.75,3+0.75);
		\fill [color=black,opacity=0.68] (9+0.25,4+0.25) rectangle (9+0.75,4+0.75);
		\fill [color=black,opacity=0.23] (9+0.25,5+0.25) rectangle (9+0.75,5+0.75);
		\fill [color=black,opacity=0.93] (9+0.25,6+0.25) rectangle (9+0.75,6+0.75);
		\fill [color=black,opacity=0.1] (9+0.25,7+0.25) rectangle (9+0.75,7+0.75);
		\fill [color=black,opacity=0.55] (9+0.25,8+0.25) rectangle (9+0.75,8+0.75);
		\fill [color=black,opacity=0.96] (9+0.25,9+0.25) rectangle (9+0.75,9+0.75);
		\fill [color=black,opacity=0.13] (9+0.25,10+0.25) rectangle (9+0.75,10+0.75);
		\fill [color=black,opacity=0.41] (9+0.25,11+0.25) rectangle (9+0.75,11+0.75);
		\fill [color=black,opacity=0.5] (10+0.25,2+0.25) rectangle (10+0.75,2+0.75);
		\fill [color=black,opacity=0.8] (10+0.25,3+0.25) rectangle (10+0.75,3+0.75);
		\fill [color=black,opacity=0.92] (10+0.25,4+0.25) rectangle (10+0.75,4+0.75);
		\fill [color=black,opacity=0.14] (10+0.25,5+0.25) rectangle (10+0.75,5+0.75);
		\fill [color=black,opacity=0.97] (10+0.25,6+0.25) rectangle (10+0.75,6+0.75);
		\fill [color=black,opacity=0.39] (10+0.25,7+0.25) rectangle (10+0.75,7+0.75);
		\fill [color=black,opacity=0.04] (10+0.25,8+0.25) rectangle (10+0.75,8+0.75);
		\fill [color=black,opacity=1.0] (10+0.25,9+0.25) rectangle (10+0.75,9+0.75);
		\fill [color=black,opacity=0.84] (10+0.25,10+0.25) rectangle (10+0.75,10+0.75);
		\fill [color=black,opacity=0.02] (10+0.25,11+0.25) rectangle (10+0.75,11+0.75);
		\fill [color=black,opacity=0.8] (11+0.25,2+0.25) rectangle (11+0.75,2+0.75);
		\fill [color=black,opacity=0.11] (11+0.25,3+0.25) rectangle (11+0.75,3+0.75);
		\fill [color=black,opacity=0.53] (11+0.25,4+0.25) rectangle (11+0.75,4+0.75);
		\fill [color=black,opacity=0.43] (11+0.25,5+0.25) rectangle (11+0.75,5+0.75);
		\fill [color=black,opacity=0.84] (11+0.25,6+0.25) rectangle (11+0.75,6+0.75);
		\fill [color=black,opacity=0.98] (11+0.25,7+0.25) rectangle (11+0.75,7+0.75);
		\fill [color=black,opacity=0.41] (11+0.25,8+0.25) rectangle (11+0.75,8+0.75);
		\fill [color=black,opacity=0.21] (11+0.25,9+0.25) rectangle (11+0.75,9+0.75);
		\fill [color=black,opacity=0.61] (11+0.25,10+0.25) rectangle (11+0.75,10+0.75);
		\fill [color=black,opacity=0.69] (11+0.25,11+0.25) rectangle (11+0.75,11+0.75);
		\fill [color=black,opacity=0.06] (12+0.25,2+0.25) rectangle (12+0.75,2+0.75);
		\fill [color=black,opacity=0.61] (12+0.25,3+0.25) rectangle (12+0.75,3+0.75);
		\fill [color=black,opacity=0.26] (12+0.25,4+0.25) rectangle (12+0.75,4+0.75);
		\fill [color=black,opacity=0.79] (12+0.25,5+0.25) rectangle (12+0.75,5+0.75);
		\fill [color=black,opacity=0.43] (12+0.25,6+0.25) rectangle (12+0.75,6+0.75);
		\fill [color=black,opacity=0.18] (12+0.25,7+0.25) rectangle (12+0.75,7+0.75);
		\fill [color=black,opacity=0.98] (12+0.25,8+0.25) rectangle (12+0.75,8+0.75);
		\fill [color=black,opacity=0.58] (12+0.25,9+0.25) rectangle (12+0.75,9+0.75);
		\fill [color=black,opacity=0.14] (12+0.25,10+0.25) rectangle (12+0.75,10+0.75);
		\fill [color=black,opacity=0.89] (12+0.25,11+0.25) rectangle (12+0.75,11+0.75);
		\fill [color=black,opacity=0.09] (13+0.25,2+0.25) rectangle (13+0.75,2+0.75);
		\fill [color=black,opacity=0.0] (13+0.25,3+0.25) rectangle (13+0.75,3+0.75);
		\fill [color=black,opacity=0.38] (13+0.25,4+0.25) rectangle (13+0.75,4+0.75);
		\fill [color=black,opacity=0.25] (13+0.25,5+0.25) rectangle (13+0.75,5+0.75);
		\fill [color=black,opacity=0.27] (13+0.25,6+0.25) rectangle (13+0.75,6+0.75);
		\fill [color=black,opacity=0.07] (13+0.25,7+0.25) rectangle (13+0.75,7+0.75);
		\fill [color=black,opacity=0.34] (13+0.25,8+0.25) rectangle (13+0.75,8+0.75);
		\fill [color=black,opacity=0.92] (13+0.25,9+0.25) rectangle (13+0.75,9+0.75);
		\fill [color=black,opacity=0.62] (13+0.25,10+0.25) rectangle (13+0.75,10+0.75);
		\fill [color=black,opacity=0.99] (13+0.25,11+0.25) rectangle (13+0.75,11+0.75);
		\draw [decorate,black,semithick] (0,11.9) --node[left=4,scale=0.8]{$N^{(k)}_{(1,1)}$} (0,10.1);
		\draw [<->,black,thick] (0.5,12.7) --node[above=0.2,scale=0.9]{distance of $2^3$} (8.5,12.7);
		\draw[color=black,thick] (0.5,12.1) -- (0.5,12.5);
		\draw[color=black,thick] (8+0.5,12.1) -- (8+0.5,12.5);
		\fill[color=gray,opacity=0.4] (0.2,11.8)--(1.8,11.8)--(19.5-2,13.5-7)--(1.8,10.2)--(0.2,10.2)--(0.2,11.8);
		\fill[color=gray,opacity=0.4] (0.2+12,11.8)--(1.8+12,11.8)--(19.5+9-2,13.5-7)--(1.8+12,10.2)--(0.2+12,10.2)--(0.2+12,11.8);
		\fill[color=gray,opacity=0.4] (0.2+12,11.8-8)--(1.8+12,11.8-8)--(19.5+9-2,13.5-7-6)--(1.8+12,10.2-8)--(0.2+12,10.2-8)--(0.2+12,11.8-8);
		\fill[color=gray,opacity=0.4] (0.2,11.8-8)--(1.8,11.8-8)--(19.5-2,13.5-7-6)--(1.8,10.2-8)--(0.2,10.2-8)--(0.2,11.8-8);
		\fill [color=black,opacity=0.9] (17,0) rectangle (18,1);
		\fill [color=black,opacity=0.87] (17,1.5) rectangle (18,2.5);
		\fill [color=black,opacity=0.57] (17,3) rectangle (18,4);
		\fill [color=black,opacity=0.89] (17,4.5) rectangle (18,5.5);
		\fill [color=black,opacity=0.64] (17,6) rectangle (18,7);
		\fill [color=black,opacity=0.94] (18.5,0) rectangle (19.5,1);
		\fill [color=black,opacity=0.88] (18.5,1.5) rectangle (19.5,2.5);
		\fill [color=black,opacity=0.56] (18.5,3) rectangle (19.5,4);
		\fill [color=black,opacity=0.96] (18.5,4.5) rectangle (19.5,5.5);
		\fill [color=black,opacity=0.96] (18.5,6) rectangle (19.5,7);
		\fill [color=black,opacity=0.92] (20,0) rectangle (21,1);
		\fill [color=black,opacity=0.92] (20,1.5) rectangle (21,2.5);
		\fill [color=black,opacity=0.98] (20,3) rectangle (21,4);
		\fill [color=black,opacity=0.72] (20,4.5) rectangle (21,5.5);
		\fill [color=black,opacity=1.0] (20,6) rectangle (21,7);
		\fill [color=black,opacity=0.66] (23,0) rectangle (24,1);
		\fill [color=black,opacity=0.66] (23,1.5) rectangle (24,2.5);
		\fill [color=black,opacity=0.82] (23,3) rectangle (24,4);
		\fill [color=black,opacity=0.71] (23,4.5) rectangle (24,5.5);
		\fill [color=black,opacity=0.98] (23,6) rectangle (24,7);
		\fill [color=black,opacity=0.98] (21.5,0) rectangle (22.5,1);
		\fill [color=black,opacity=0.68] (21.5,1.5) rectangle (22.5,2.5);
		\fill [color=black,opacity=0.93] (21.5,3) rectangle (22.5,4);
		\fill [color=black,opacity=0.96] (21.5,4.5) rectangle (22.5,5.5);
		\fill [color=black,opacity=0.7] (21.5,6) rectangle (22.5,7);
		\fill [color=black,opacity=0.8] (24.5,0) rectangle (25.5,1);
		\fill [color=black,opacity=0.92] (24.5,1.5) rectangle (25.5,2.5);
		\fill [color=black,opacity=0.98] (24.5,3) rectangle (25.5,4);
		\fill [color=black,opacity=1.0] (24.5,4.5) rectangle (25.5,5.5);
		\fill [color=black,opacity=0.84] (24.5,6) rectangle (25.5,7);
		\fill [color=black,opacity=0.61] (26,0) rectangle (27,1);
		\fill [color=black,opacity=0.79] (26,1.5) rectangle (27,2.5);
		\fill [color=black,opacity=0.43] (26,3) rectangle (27,4);
		\fill [color=black,opacity=0.98] (26,4.5) rectangle (27,5.5);
		\fill [color=black,opacity=0.99] (26,6) rectangle (27,7);
		goo
		\draw [<->,black,thick] (17.5,7.8) --node[above=0.2,scale=0.9]{distance of $\frac{2^3}{2}=4$} (23.5,7.8);
		\draw[color=black,thick] (17+0.5,7.2) -- (17+0.5,7.6);
		\draw[color=black,thick] (23+0.5,7.2) -- (23+0.5,7.6);
	\end{tikzpicture}
	\caption{llustration of the local max-pooling of a feature map with neighborhood size $n_k=2$.}
\end{figure}
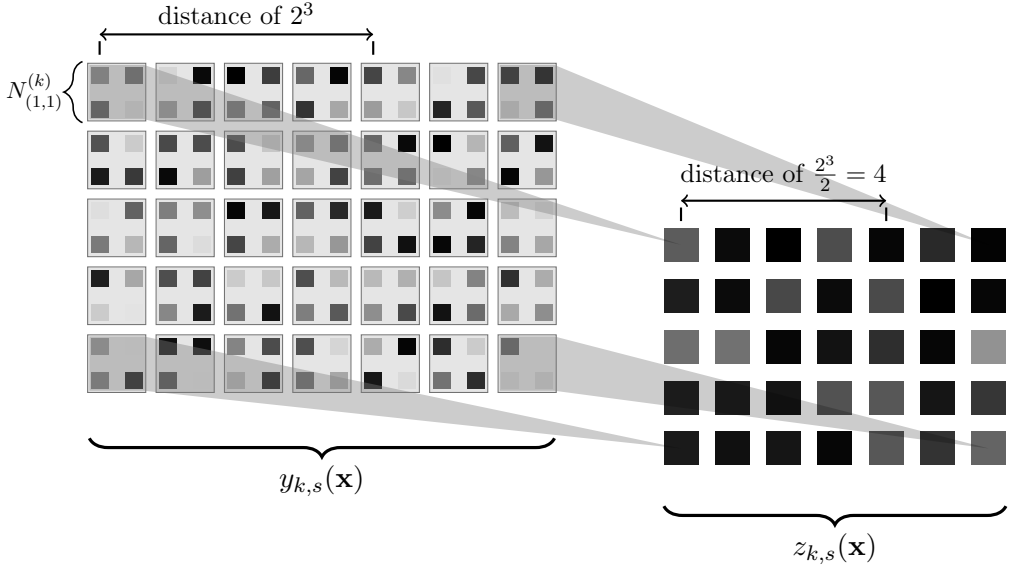
As can be seen in Figure 1, the distances of the underlying subparts of the input image, which are hierarchically combined in Definition 2, are reduced in the resulting feature map with local max-pooling.
Therefore, we introduce the parameter $\delta_k$ in the following definition, which describes the distance $2^{k}$ of the underlying subparts adapted to the feature maps with local max-pooling of level $k$.
\begin{definition}
  \label{de2}
	Let $d_1,d_2,l\in\N$ and  $n_0,n_1,\dots,n_{l}\in\{2^0,2^1,\dots,2^{l-1}\}$ with $n_0=n_l=1$, 
	\begin{equation}
		\label{de3eq1}
		\prod_{i=1}^{k}n_i\leq 2^{k}
	\end{equation}
	for $k\in\{1,\dots,l-1\}$ and
	\begin{equation}
		\label{de3eq2}
		\min\{d_1,d_2\}\geq2^l+\prod_{k=1}^{l-1}n_k-1.
	\end{equation}
	Furthermore, let $b_0,\dots,b_{l}\in\N$ with $b_0=b_l=1$ and set $\bb=(b_1,\dots,b_{l-1})$ and $\bn=(n_1,\dots,n_{l-1})$.
		
	\noindent
{\bf a)}
For $k=1,\dots,l$ we set $\delta_{k-1}={2^{k-1}}/{\prod_{i=0}^{k-1}n_i}$
and define recursively the dimensions
\[
d_1(k)=	\left\lceil
\frac{d_1(k-1)}{n_{k-1}}
\right\rceil
-\delta_{k-1}
\mbox{ and }
d_2(k)=	\left\lceil
\frac{d_2(k-1)}{n_{k-1}}
\right\rceil
-\delta_{k-1},
\]
where we set $d_1(0)=d_1$ and $d_2(0)=d_2$.
We say that 
\[z:[0,1]^{\{1, \dots, d_1\} \times \{1, \dots, d_2\}}\rightarrow[0,1]^{\left\{1,\dots,d_1(l)\right\}\times\left\{1,\dots,d_2(l)\right\}}\]
satisfies a \textbf{hierarchical model of level} $l$ {\bf with feature constraint} $\bb$ \textbf{and local max-pooling parameter} $\bn$,
if there exist functions
\[g_{k,s}:\R^4\rightarrow[0,1]\quad(k=1,\dots,l,s=1,\dots,b_s)\]
such that we have
\[z=z_{l,1}\]
for some
\[
z_{k,s}:[0,1]^{\{1, \dots, d_1\} \times \{1, \dots, d_2\}}\rightarrow[0,1]^{\left\{1,\dots,\left\lceil\frac{d_1(k)}{n_k}\right\rceil\right\}\times\left\{1,\dots,\left\lceil\frac{d_2(k)}{n_k}\right\rceil\right\}}
\]
recursively defined as follows:
\begin{enumerate}
	\item 
	We set
	\[
	z_{0,1}(\bx)=\bx.
	\]
	\item
	We use a hierarchical model to define recursively feature maps 
	\[
	y_{k,s}:[0,1]^{\{1,\dots,d_1\}\times\{1,\dots,d_2\}}\rightarrow[0,1]^{\{1,\dots,d_1(k)\}\times\{1,\dots,d_2(k)\}}
	\]
    for $k=1,\dots,l$ by
	\begin{align*}
		&\big(y_{k,s}(\bx)\big)_{(i,j)}=g_{k,s}\Big(
		\big(z_{k-1,r_1(k,s)}(\bx)\big)_{(i,j)},
		\big(z_{k-1,r_2(k,s)}(\bx)\big)_{(i+\delta_{k-1},j)},
		\\&\hspace{3.5cm}
		\big(z_{k-1,r_3(k,s)}(\bx)\big)_{(i,j+\delta_{k-1})},\big(z_{k-1,r_4(k,s)}(\bx)\big)_{(i+\delta_{k-1},j+\delta_{k-1})}\Big)
	\end{align*}
for $k=1,\dots,l$, $s=1,\dots,b_k$, $(i,j)\in\{1,\dots,d_1(k)\}\times\{1,\dots,d_2(k)\}$, 
$r_1(k,s)$, $r_2(k,s)$, $r_3(k,s)$, $r_4(k,s)\in\{1,\dots,b_{k-1}\}.$
\item
Next we define the feature maps with local max-pooling by
\[
\big(z_{k,s}(\bx)\big)_{(i,j)}=
\max_{(i_2,j_2)\in N^{(k)}_{(i,j)}}
\big(y_{k,s}(\bx)\big)_{(i_2,j_2)}
\]
for $k=1,\dots,l$, $s=1,\dots,b_k$ and $(i,j)\in\{1,\dots,\lceil d_1(k)/n_k\rceil\}\times\{1,\dots,\lceil d_2(k)/n_k\rceil\}$, where the neighborhoods 
$N_{(i,j)}^{(k)}$ are defined by equation \eqref{NeighbH}.
\end{enumerate}

	\noindent
	{\bf b)}
	We say that $m: [0,1]^{\{1, \dots, d_1\} \times \{1, \dots, d_2\}} \rightarrow[0,1]$ satisfies a \textbf{hierarchical max-pooling model of level} $l$ {\bf with feature constraint} $\bb$ \textbf{and local max-pooling parameter} $\bn$, if there exist a function
	\[z:[0,1]^{\{1, \dots, d_1\} \times \{1, \dots, d_2\}}\rightarrow[0,1]^{\left\{1,\dots,d_1(l)\right\}\times\left\{1,\dots,d_2(l)\right\}}\]
	which satisfies a hierarchical model of level $l$ with feature constraint $\bb$ and local max-pooling parameter $\bn$
	such that
	\[m(\bx)=
	\max_{
		(i,j)\in\{1,\dots,d_1(l)\}\times\{1,\dots,d_2(l)\}}
	\big(z(\bx)\big)_{(i,j)}.\]

    \noindent
{\bf c)}
Let $p=q+s$ for some $q \in \N_0$ and $s \in (0,1]$, and let $C>0$.
We say that a  hierarchical max-pooling model has smoothness constraint $p$
if all functions $g_{k,s}$ in its definition are $(p,C)$--smooth.
\end{definition}
\noindent
{\bf Remark 4.}
	The above model is a generalization of the hierarchical max-pooling model with feature constraint
	in Definition \ref{de1}, since this model is equivalent to the new model
	above in case we choose $n_1=n_2=\dots=n_{l-1}=1$.

\noindent
{\bf Remark 5.}
	Condition \eqref{de3eq1} ensures that the neighborhoods do not become too large relative to the sizes of the underlying subparts of the input image belonging to the corresponding feature map. Condition \eqref{de3eq2} ensures that the dimensions $d_1(k)$ and $d_2(k)$ are greater than zero for all $k\in\{1,\dots,l\}$.

\noindent
{\bf Remark 6.}
	Since we assume $n_l=1$ in the above definition,  we have $z_{l,1}=y_{l,1}$ in part a).\\
\section{Convolutional neural network image classifiers}
\label{se3}
In this section we describe the architecture of the convolutional neural
network, which we use for the definition of our estimates.

The input of our convolutional neural networks is a $[0,1]^{\{1,\dots,d_1\}\times\{1,\dots,d_2\}}$--valued image with the image dimensions $d_1,d_2\in\N$.
In our network architecture we use the so-called ReLU activation function $\sigma:\R\rightarrow\R$, which is defined by
$\sigma(x)=\max\{x,0\}.$
Our convolutional neural network consists of several convolutional blocks, where each convolutional block is followed by a sub-sampling layer. Then a linear layer follows and the output of the convolutional neural network is computed by a global max-pooling layer. Here a convolutional block consists of several convolutional layers.

\subsection{Convolutional blocks}
The convolutional block that we define consists of several convolutional layers.
A convolutional layer receives the output of the previous layer as input, which can be either the output of a sub-sampling layer, a convolutional layer or the input image.
The neurons of a convolutional layer are arranged in planes of equal size, which we call channels (also called feature maps). 
The parameter $k\in\N$ denotes the number of channels in a convolutional layer. %
Each neuron in a channel is tagged with an index, which defines its position in the corresponding plane.
Therefore, each convolutional layer is assigned to an index set $I=\{1,\dots,i_1\}\times\{1,\dots,i_2\}$ for $i_1,i_2\in\N$, which contains the different positions of the $k$ channels. In our construction we will used a so--called
zero-padding, and consequently the dimension of the channels will not
be reduced by the convolutional layer.

The output of a convolutional layer is calculated by a function
\[
{o_{(k',k),M,\bw}}:\R^{I \times\{1,\dots,k'\}}\rightarrow\R^{I\times\{1,\dots,k\}},
\]
where the value $k'\in\N$ correspond to  the number of channels of the previous layer. If the input of the convolutional layer is the input image, we have
$I=\{1,\dots,d_1\}\times\{1,\dots,d_2\}$ and $k'=1$. 
Furthermore, a convolutional layer depends on a filter size $M\in\N$ and
has the following trainable weights:
\begin{enumerate}
	\item A weight matrix (so--called filter)
	\begin{equation}
		\left(
		w_{i,j,s_1,s_2}
		\right)_{
			1 \leq i,j \leq M, s_1 \in \{1, \dots, k'\}, s_2 \in \{1, \dots, k\}
		}
	\end{equation}
\item and bias weights
\begin{equation}
	\left(
	w_{s_2}
	\right)_{
		s_2 \in \{1, \dots, k\}.
	}
\end{equation}
\end{enumerate}
Let
\begin{equation*}
	\bw= \left(
	\left(
	w_{i,j,s_1,s_2}
	\right)_{
		1 \leq i,j \leq M, s_1 \in \{1, \dots, k'\}, s_2 \in \{1, \dots, k\}
	}
	,
	\left(
	w_{s_2}
	\right)_{
		s_2 \in \{1, \dots, k\}
	}         
	\right)
\end{equation*}
be the vector of all weights of the convolutional layer.

For $(i,j)\in I$ and $s_2\in\{1,\dots,k\}$ we compute the output of a convolutional layer by
\begin{align}
	\label{s2eq1}
	&\big(o_{(k',k),M,\bw}(\bx)\big)_{(i,j),s_2}
	=
	\sigma \left(
	\sum_{s_1=1}^{k'}
	\sum_{
		\substack{
			t_1,t_2 \in \{1, \dots, M\}
			\\(i+t_1-1,j+t_2-1)\in I}
	}
	w_{t_1,t_2,s_1,s_2}
	\cdot
	x_{(i+t_1-1,j+t_2-1),s_1}
	+
	w_{s_2}
	\right).
\end{align}
In case $k'=1$ we identify $\R^{I \times\{1\}}$
with $\R^{I }$ and define our convolutional layer  by the function
\[
{o}_{(1,k),M,\bw}:\R^{I}\rightarrow\R^{I\times\{1,\dots,k\}},
\]
\[
\big(o_{(1,k),M,\bw}(\bx)\big)_{(i,j),s_2}
	=
	\sigma \left(
	\sum_{
		\substack{
			t_1,t_2 \in \{1, \dots, M\}
			\\(i+t_1-1,j+t_2-1)\in I}
	}
	w_{t_1,t_2,1,s_2}
	\cdot
	x_{(i+t_1-1,j+t_2-1)}
	+
	w_{s_2}
	\right).
        \]

A convolutional block of size $z\in\N$ with parameters $k$ and $M$ is then defined as a function
\[o^{(z)}_{(k',k),M}:\R^{I\times\{1,\dots,k'\}}\rightarrow\R^{I\times\{1,\dots,k\}}
\]
given by
\begin{equation}
	\label{convblock}
	o_{(k',k),M}^{(z)}=
          o_{(k,k),M, \bw_z}\circ o_{(k,k),M, \bw_{z-1}}\circ\dots\circ o_{(k',k),M,\bw_1}
       ,
\end{equation}
where the $z$ convolutional layers use the different weight vectors
\begin{equation}
	\label{weightsblock}
	\bw_r= \left(
	\left(
	w_{i,j,s_1,s_2}^{(r)}
	\right)_{
		1 \leq i,j \leq M, s_1 \in \{1, \dots, k'\}, s_2 \in \{1, \dots, k\}
	}
	,
	\left(
	w_{s_2}^{(r)}
	\right)_{
		s_2 \in \{1, \dots, k\}
	}         
	\right)
\end{equation}
for $r=1,\dots,z$.
In case $k'=1$ we identify again
$\R^{I\times\{1\}}$ with $\R^I$.

\subsection{Local max-pooling layers}
A local max-pooling layer is an example of a so-called pooling layer. A pooling layer reduces the resolution of the output channels from the previous layer.
This is done by summarizing the outputs of several neurons from a local neighborhood of a channel.
In case of a local max-pooling, the local neighborhood of a output is summarized by the maximum of the outputs in the corresponding neighborhood.
The input of a local max-pooling layer is $\R^{I\times\{1,\dots,k\}}$-valued
for an index set $I=\{1,\dots,i_1\}\times\{1,\dots,i_2\}$ with $i_1,i_2\in\N$ and $k\in\N$. 
The local max-pooling layer depends on a parameter $s\in\N$, which denotes the size of the local neighborhood.
We define a local max-pooling layer as a function
\[f_{max}^{(s)}:\R^{I\times\{1,\dots,k\}}\rightarrow\R^{\left\{1,\dots,\left\lceil\frac{i_1}{s}\right\rceil\right\}\times\left\{1,\dots,\left\lceil\frac{i_2}{s}\right\rceil\right\}\times \{1,\dots,k\}},\]
of the following form:

For $(i,j)\in \left\{1,\dots,\left\lceil\frac{i_1}{s}\right\rceil\right\}\times\left\{1,\dots,\left\lceil\frac{i_2}{s}\right\rceil\right\}$ and $s_2\in\{1,\dots,k\}$ we define
\[\big(f_{max}^{(s)}(\bx)\big)_{(i,j),s_2}=
\max_{(i_2,j_2)\in\big(\{(i-1)\cdot s+1\dots,i\cdot s\}\times\{(j-1)\cdot s+1\dots,j\cdot s\}\big)\cap I}
x_{(i_2,j_2),s_2}
\quad\big(\bx\in\R^{I\times\{1,\dots,k\}}\big).
\]
The local max-pooling layer depends only on the parameter $s\in\N$ and has no trainable weights.
\subsection{Subsampling layers}
The subsampling layer is another example of a pooling layer, which in practice is also performed by a so-called convolutional stride (cf., e.g., \cite{Goodfellow2016}).
Here the outputs of several neurons from a local neighborhood of a channel are replaced by just one of these values corresponding to some fixed position in this neighborhood.
For simplicity here we just choose
the value of the neuron in the left upper corner of the neighborhood.
The input of a subsampling layer is again $\R^{I\times\{1,\dots,k\}}$-valued
for an index set $I=\{1,\dots,i_1\}\times\{1,\dots,i_2\}$ with $i_1,i_2\in\N$ and $k\in\N$, and
the subsampling layer depends again on a parameter $s\in\N$, which denotes the size of the local neighborhood.
We define a subsampling layer as a function
\[f_{sub}^{(s)}:\R^{I\times\{1,\dots,k\}}\rightarrow\R^{\left\{1,\dots,\left\lceil\frac{i_1}{s}\right\rceil\right\}\times\left\{1,\dots,\left\lceil\frac{i_2}{s}\right\rceil\right\}\times \{1,\dots,k\}},\]
of the following form:

For $(i,j)\in \left\{1,\dots,\left\lceil\frac{i_1}{s}\right\rceil\right\}\times\left\{1,\dots,\left\lceil\frac{i_2}{s}\right\rceil\right\}$ and $s_2\in\{1,\dots,k\}$ we define
\[\big(f_{sub}^{(s)}(\bx)\big)_{(i,j),s_2}=
x_{((i-1)\cdot s+1,(j-1)\cdot s+1),s_2}
\quad\big(\bx\in\R^{I\times\{1,\dots,k\}}\big).
\]
The subsampling layer depends only on the parameter $s\in\N$ and has no trainable weights.
\subsection{Output layer}
We now define the output layer of our convolutional neural network, which consists of a global max-pooling layer
applied to a linear combination of the channels of the previous layer.
The output layer depends on 
output bounds $(\tilde{d}_1,\tilde{d}_2)\in I$ and
the trainable output weights
\begin{equation}
	\label{outputweights}
	\bw_{out}=(w_{s})_{
		s \in \{1, \dots, k\}
	},
\end{equation}
where $k\in\N$ denotes the number of channels of the previous convolutional layer.
The output layer is computed by a function 
$f_{out}:\R^{I\times\{1,\dots,k\}}\rightarrow\R$
defined by
\[
f_{out}^{(\tilde{d}_1,\tilde{d}_2)}(\bx)=\max\left\{\sum_{s_2=1}^{k}w_{s_2}\cdot x_{(i,j),s_2} : (i,j)\in\{1,\dots,\tilde{d}_1\}\times\{1,\dots,\tilde{d}_2\}\right\}.
\]

\subsection{Three classes of convolutional neural networks}
We next introduce the three classes of convolutional neural networks
which we investigate in this articel. The difference between these
three classes is that the first uses severeal local max-pooling layers, the
second several subsampling layers and the third uses only one subsampling layer after the convolutional layers.

For the first class our convolutional neural network is given by a function
\[f:[0,1]^{\{1,\dots,d_1\}\times\{1,\dots,d_2\}}\rightarrow\R\]
defined by
\begin{equation}
	\label{cnndef}
f(\bx)
	=f_{out}^{(\tilde{d}_1,\tilde{d}_2)}\circ o^{(z)}_{(k_{L-1},k_L),M_L}\circ f_{max}^{(s_{L-1})}\circ o^{(z)}_{(k_{L-2},k_{L-1}),M_{L-1}}\circ\dots\circ f_{max}^{(s_1)}\circ o^{(z)}_{(1,k_1),M_1}(\bx).
\end{equation}
For our second class of convolutional neural networks this
function is defined by
\begin{equation}
	\label{cnndefb}
f(\bx)
=f_{out}^{(\tilde{d}_1,\tilde{d}_2)}\circ o^{(z)}_{(k_{L-1},k_L),M_L}\circ f_{sub}^{(s_{L-1})}\circ o^{(z)}_{(k_{L-2},k_{L-1}),M_{L-1}}\circ\dots\circ f_{sub}^{(s_1)}\circ o^{(z)}_{(1,k_1),M_1}(\bx).
\end{equation}
For our third class of convolutional neural networks this
function is defined by
\begin{equation}
	\label{cnndefc}
	f(\bx)
	=f_{out}^{(\tilde{d}_1,\tilde{d}_2)}\circ f_{sub}^{(s)}\circ o^{(z)}_{(k_{L-1},k_L),M_L}\circ o^{(z)}_{(k_{L-2},k_{L-1}),M_{L-1}}\circ\dots\circ o^{(z)}_{(1,k_1),M_1}(\bx).
\end{equation}
In all three cases the class of
convolutional neural networks $f$ depends on the parameters
$L$, $\bk=(k_1,\dots,k_L)$, $\bM=(M_1,\dots,M_L)$, $z$
and $\tilde{\bd}=(\tilde{d}_1,\tilde{d}_2)$. The first two classes additionally depend on the parameter $\bs=(s_1,\dots,s_{L-1})$ and the third class on the parameter $s$.
Moreover, the functions of the three classes depend on the weights
\[
\bw=\big((\bw_{s,1},\dots,\bw_{s,z})_{s\in\{1,\dots,L\}},\bw_{out}\big),
\]
where $\bw_{s,r}$ is of the form \eqref{weightsblock} and corresponds to the weight vector of the $r$-th convolutional layer in the $s$-th convolutional block and $\bw_{out}$ denotes the weights of the output layer \eqref{outputweights}.
We define the three classes of convolutional neural networks by
\begin{align*}
	&\F_1(\btheta_1)=\big\{f : f \text{ is of the form \eqref{cnndef} with parameters }\\
	&\hspace{3cm}
	\btheta_1=(L,\bk,\bM,z,\bs,\tilde{\bd})\text{  and weights }\bw\big\},
\end{align*}
\begin{align*}
	&\F_2(\btheta_2)=\big\{f : f \text{ is of the form \eqref{cnndefb} with parameters }\\
	&\hspace{3cm}
	\btheta_2=(L,\bk,\bM,z,\bs,\tilde{\bd})\text{  and weights }\bw\big\}
\end{align*}
and
\begin{align*}
&\F_3(\btheta_3)=\big\{f : f \text{ is of the form \eqref{cnndefc} with parameters }\\
&\hspace{3cm}
\btheta_3=(L,\bk,\bM,z,s,\tilde{\bd})\text{  and weights }\bw\big\}.
\end{align*}
The different network architectures are illustrated in Figure 2 to 4.
\begin{figure}[h]
	\centering
	\scalebox{.8}{
		\begin{tabular}{l@{}l@{}}
			\begin{tikzpicture}[decoration={brace,mirror,amplitude=7},scale=1.1]
				
				\fill [color=black,opacity=0.1] (0.9,-2.2) rectangle (3.6,1);
				\fill [color=black,opacity=0.1] (4.1,-2.2) rectangle (4.9,1);
				\fill [color=black,opacity=0.1] (5.6,-2.2) rectangle (7.4,1);
				\fill [color=black,opacity=0.1] (8.2,-2.2) rectangle (8.8,1);
				\fill [color=black,opacity=0.1] (9.7,-2.2) rectangle (11.3,1);
				\fill [color=black,opacity=0.1] (11.6,-1) rectangle (13.3,-0.2);
				
				\node at (-0.5,-1.5){\begin{tabular}{c}$\bx$\end{tabular}};
				
				\node at (-0.5,-0.6) [rectangle,minimum size=26,fill=black!25!white,draw=black] (input) {};
				
				\node at (2.2,1.5){\begin{tabular}{c}convolutional\\block\end{tabular}};
				\foreach \i in {0,...,1}
				\foreach \j in {0,...,2}
				\node at (1.5+\i*1.5,0.4-\j*1) [rectangle,minimum size=26,fill=black!25!white,draw=black] (\i\j) {};
				\node at (4.5,-3){\begin{tabular}{c}local\\max-pooling\\layer\end{tabular}};
				\foreach \i in {2}
				\foreach \j in {0,...,2}
				\node at (\i+2.5,0.4-\j*1) [rectangle,minimum size=14,fill=black!25!white,draw=black] (\i\j) {};
				\node at (6.5,1.5){\begin{tabular}{c}convolutional\\block\end{tabular}};
				\foreach \i in {3,...,4}
				\foreach \j in {0,...,2}
				\node at (\i+3,0.4-\j*1) [rectangle,minimum size=14,fill=black!25!white,draw=black] (\i\j) {};
				\node at (8.5,-3){\begin{tabular}{c}local\\max-pooling\\layer\end{tabular}};	
				\foreach \i in {5}
				\foreach \j in {0,...,2}
				\node at (\i+3.5,0.4-\j*1) [rectangle,minimum size=7,fill=black!25!white,draw=black] (\i\j) {};
				\node at (10.5,1.5){\begin{tabular}{c}convolutional\\block\end{tabular}};
				\foreach \i in {6,7}
				\foreach \j in {0,...,2}
				\node at (\i+4,0.4-\j*1) [rectangle,minimum size=7,fill=black!25!white,draw=black] (\i\j) {};
				\foreach \i in {8}
				\foreach \j in {1}
				\node at (\i+4,0.4-\j*1) [rectangle,minimum size=7,fill=black!25!white,draw=black] (\i\j) {};
				\foreach \i in {9}
				\foreach \j in {1}
				\draw[black,fill] (\i+4,0.4-\j*1)  circle [radius=0.04]; 
				\node at (12.5,-1.5){\begin{tabular}{c}output\\layer\end{tabular}};

				\foreach \j in {0,...,2}
				\draw[line width=0.5] (input) edge (0\j);
				\foreach \i in {0,...,2}
				\foreach \j in {0,...,2}
				\draw[line width=0.5] (0\i) edge (1\j);
				\foreach \i in {1}
				\foreach \u in {2}
				\foreach \j in {0,...,2}
				\draw[line width=0.5] (\i\j) edge (\u\j);
				\foreach \i in {2}
				\foreach \u in {3}
				\foreach \j in {0,...,2}
				\foreach \r in {0,...,2}
				\draw[line width=0.5] (\i\j) edge (\u\r);
				\foreach \i in {3}
				\foreach \u in {4}
				\foreach \j in {0,...,2}
				\foreach \r in {0,...,2}
				\draw[line width=0.5] (\i\j) edge (\u\r);
				\foreach \i in {4}
				\foreach \u in {5}
				\foreach \j in {0,...,2}
				\draw[line width=0.5] (\i\j) edge (\u\j);
				\foreach \i in {5}
				\foreach \u in {6}
				\foreach \j in {0,...,2}
				\foreach \r in {0,...,2}
				\draw[line width=0.5] (\i\j) edge (\u\r);
				\foreach \i in {6}
				\foreach \u in {7}
				\foreach \j in {0,...,2}
				\foreach \r in {0,...,2}
				\draw[line width=0.5] (\i\j) edge (\u\r);
				\foreach \i in {7}
				\foreach \u in {8}
				\foreach \j in {0,...,2}
				\foreach \r in {1}
				\draw[line width=0.5] (\i\j) edge (\u\r);
				\draw[line width=0.5] (81) edge (13,-0.6);
			\end{tikzpicture}
		\end{tabular}
	}
	\caption{Illustration of the network architecture of the class $\F_{1}(\btheta_1)$ with $L=3$, $k_1=k_2=k_3=3$ and $z=2$.}
\end{figure}
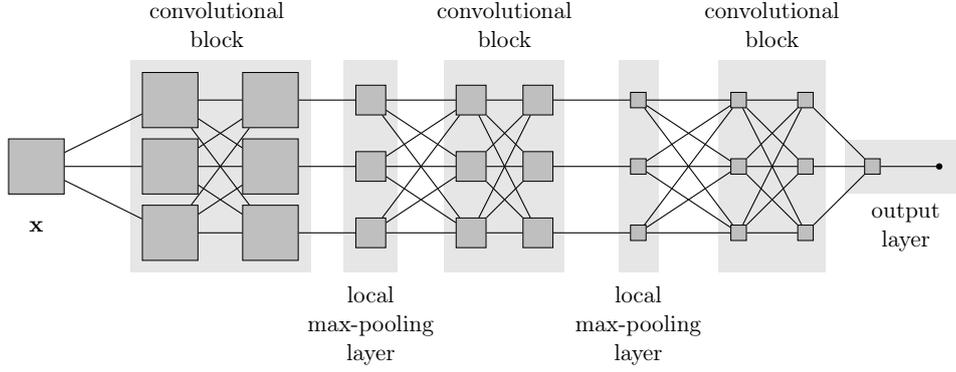
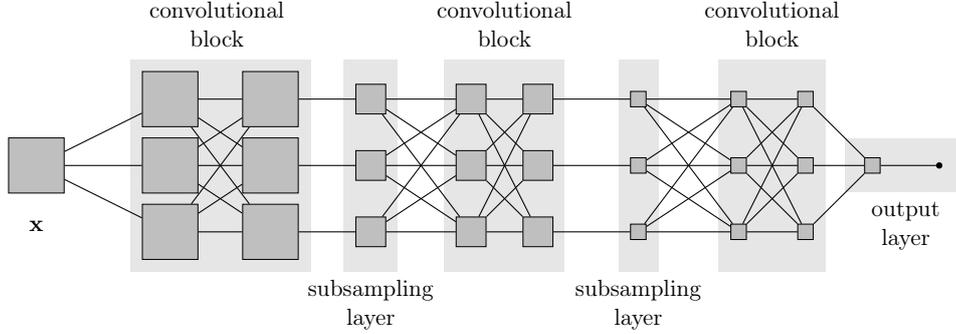
\begin{figure}[h]
	\centering
	\scalebox{.8}{
		\begin{tabular}{l@{}l@{}}
			\begin{tikzpicture}[decoration={brace,mirror,amplitude=7},scale=1.1]
				
				\fill [color=black,opacity=0.1] (0.9,-2.2) rectangle (3.6,1);
				\fill [color=black,opacity=0.1] (4.1,-2.2) rectangle (4.9,1);
				\fill [color=black,opacity=0.1] (5.6,-2.2) rectangle (7.4,1);
				\fill [color=black,opacity=0.1] (8.2,-2.2) rectangle (8.8,1);
				\fill [color=black,opacity=0.1] (9.7,-2.2) rectangle (11.3,1);
				\fill [color=black,opacity=0.1] (11.6,-1) rectangle (13.3,-0.2);
				\node at (-0.5,-1.5){\begin{tabular}{c}$\bx$\end{tabular}};
				
				\node at (-0.5,-0.6) [rectangle,minimum size=26,fill=black!25!white,draw=black] (input) {};
				
				\node at (2.2,1.5){\begin{tabular}{c}convolutional\\block\end{tabular}};
				\foreach \i in {0,...,1}
				\foreach \j in {0,...,2}
				\node at (1.5+\i*1.5,0.4-\j*1) [rectangle,minimum size=26,fill=black!25!white,draw=black] (\i\j) {};
				\node at (4.5,-2.7){\begin{tabular}{c}subsampling\\layer\end{tabular}};
				\foreach \i in {2}
				\foreach \j in {0,...,2}
				\node at (\i+2.5,0.4-\j*1) [rectangle,minimum size=14,fill=black!25!white,draw=black] (\i\j) {};
				\node at (6.5,1.5){\begin{tabular}{c}convolutional\\block\end{tabular}};
				\foreach \i in {3,...,4}
				\foreach \j in {0,...,2}
				\node at (\i+3,0.4-\j*1) [rectangle,minimum size=14,fill=black!25!white,draw=black] (\i\j) {};
				\node at (8.5,-2.7){\begin{tabular}{c}subsampling\\layer\end{tabular}};	
				\foreach \i in {5}
				\foreach \j in {0,...,2}
				\node at (\i+3.5,0.4-\j*1) [rectangle,minimum size=7,fill=black!25!white,draw=black] (\i\j) {};
				\node at (10.5,1.5){\begin{tabular}{c}convolutional\\block\end{tabular}};
				\foreach \i in {6,7}
				\foreach \j in {0,...,2}
				\node at (\i+4,0.4-\j*1) [rectangle,minimum size=7,fill=black!25!white,draw=black] (\i\j) {};
				\foreach \i in {8}
				\foreach \j in {1}
				\node at (\i+4,0.4-\j*1) [rectangle,minimum size=7,fill=black!25!white,draw=black] (\i\j) {};
				\foreach \i in {9}
				\foreach \j in {1}
				\draw[black,fill] (\i+4,0.4-\j*1)  circle [radius=0.04]; 
				\node at (12.5,-1.5){\begin{tabular}{c}output\\layer\end{tabular}};

				\foreach \j in {0,...,2}
				\draw[line width=0.5] (input) edge (0\j);
				\foreach \i in {0,...,2}
				\foreach \j in {0,...,2}
				\draw[line width=0.5] (0\i) edge (1\j);
				\foreach \i in {1}
				\foreach \u in {2}
				\foreach \j in {0,...,2}
				\draw[line width=0.5] (\i\j) edge (\u\j);
				\foreach \i in {2}
				\foreach \u in {3}
				\foreach \j in {0,...,2}
				\foreach \r in {0,...,2}
				\draw[line width=0.5] (\i\j) edge (\u\r);
				\foreach \i in {3}
				\foreach \u in {4}
				\foreach \j in {0,...,2}
				\foreach \r in {0,...,2}
				\draw[line width=0.5] (\i\j) edge (\u\r);
				\foreach \i in {4}
				\foreach \u in {5}
				\foreach \j in {0,...,2}
				\draw[line width=0.5] (\i\j) edge (\u\j);
				\foreach \i in {5}
				\foreach \u in {6}
				\foreach \j in {0,...,2}
				\foreach \r in {0,...,2}
				\draw[line width=0.5] (\i\j) edge (\u\r);
				\foreach \i in {6}
				\foreach \u in {7}
				\foreach \j in {0,...,2}
				\foreach \r in {0,...,2}
				\draw[line width=0.5] (\i\j) edge (\u\r);
				\foreach \i in {7}
				\foreach \u in {8}
				\foreach \j in {0,...,2}
				\foreach \r in {1}
				\draw[line width=0.5] (\i\j) edge (\u\r);
				\draw[line width=0.5] (81) edge (13,-0.6);
			\end{tikzpicture}
		\end{tabular}
	}
	\caption{Illustration of the network architecture of the class $\F_{2}(\btheta_2)$ with $L=3$, $k_1=k_2=k_3=3$ and $z=2$.}
\end{figure}
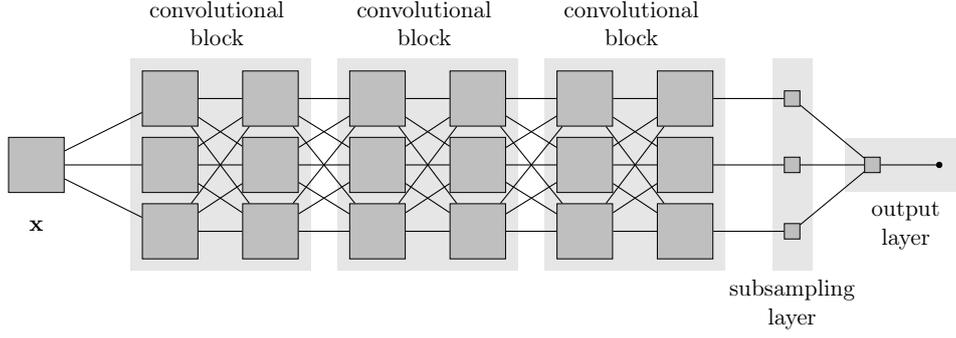
\begin{figure}[h]
	\centering
	\scalebox{.8}{
		\begin{tabular}{l@{}l@{}l@{}}
			\begin{tikzpicture}[decoration={brace,mirror,amplitude=7},scale=1.1]
				\fill [color=black,opacity=0.1] (10.5,-2.2) rectangle (11.1,1);
				\foreach \u in {0,...,2}{
					\fill [color=black,opacity=0.1] (0.9+\u*3.1,-2.2) rectangle (3.6+\u*3.1,1);}
				
				\fill [color=black,opacity=0.1] (11.6,-1) rectangle (13.3,-0.2);
				
				\node at (-0.5,-1.5){\begin{tabular}{c}$\bx$\end{tabular}};
				
				\node at (-0.5,-0.6) [rectangle,minimum size=26,fill=black!25!white,draw=black] (input) {};
				
				\foreach \u in {1,...,3}{
					\node at (2.2+\u*3.1-3.1,1.5){\begin{tabular}{c}convolutional\\block\end{tabular}};}
				\foreach \u in {0,...,2}{
					\foreach \i in {0,...,1}
					\foreach \j in {0,...,2}
					\node at (1.5+\i*1.5+\u*3.1,0.4-\j*1) [rectangle,minimum size=26,fill=black!25!white,draw=black] (\i\j\u) {};}
				\node at (10.8,-2.7){\begin{tabular}{c}subsampling\\layer\end{tabular}};		
				\foreach \i in {7}
				\foreach \j in {0,...,2}
				\node at (\i+3.8,0.4-\j*1) [rectangle,minimum size=6,fill=black!25!white,draw=black] (\i\j) {};

				\foreach \i in {8}
				\foreach \j in {1}
				\node at (\i+4,0.4-\j*1) [rectangle,minimum size=7,fill=black!25!white,draw=black] (\i\j) {};
				\foreach \i in {9}
				\foreach \j in {1}
				\draw[black,fill] (\i+4,0.4-\j*1)  circle [radius=0.04]; 
				\node at (12.5,-1.5){\begin{tabular}{c}output\\layer\end{tabular}};
				\foreach \u in {0,...,2}{
					\foreach \i in {0,...,2}{
						\foreach \j in {0,...,2}{
							\draw[line width=0.5] (0\i\u) edge (1\j\u);}}}
				
				\foreach \i in {0,...,2}{
					\foreach \j in {0,...,2}{
						\draw[line width=0.5] (1\i0) edge (0\j1);}}
				\foreach \i in {0,...,2}{
					\foreach \j in {0,...,2}{
						\draw[line width=0.5] (1\i1) edge (0\j2);}}
				\foreach \j in {0,...,2}{
					\draw[line width=0.5] (input) edge (0\j0);}
				\foreach \j in {0,...,2}{
					\draw[line width=0.5] (1\j2) edge (7\j);}
				\foreach \j in {0,...,2}{
					\draw[line width=0.5] (7\j) edge (81);}				
				\draw[line width=0.5] (81) edge (13,-0.6);	
			\end{tikzpicture}
		\end{tabular}
	}
	\caption{
		Illustration of the network architecture of the class $\F_{3}(\btheta_3)$ with $L=3$, $k_1=k_2=k_3=3$ and $z=2$.}
	\label{fig:traditional-convolutional-network}
\end{figure}
\subsection{Convolutional neural network image classifiers}
\label{subse37}
We are now ready to define the convolutional neural network
image classifiers which we will analyze in this article.
For $j \in \{1,2,3\}$ define the least squares estimate of
$\eta(\bx)=\EXP\{Y=1|\bX=\bx\}$
by
\begin{equation}
	\eta_n^{(j)} = \argmin_{f \in \F_j(\btheta_j)}
	\frac{1}{n} \sum_{i=1}^n |Y_i - f(\bX_i)|^2.
	\label{eq:lqp}
\end{equation}
Then our estimate $f_n^{(j)}$ is for $j \in \{1,2,3\}$ defined by
\[
f_n^{(j)}(\bx)=
\begin{cases}
	1, & \mbox{if } \eta_n(\bx) \geq \frac{1}{2} \\
	0, & \mbox{elsewhere}.
\end{cases}
\]

\section{Main result}
\label{se4}
Our main result is the following theorem, in which we present an upper bound on the distance between the expected misclassification
risk of our plug-in classifier and the optimal misclassification risk
in case
that the a posteriori probability satisfies a hierarchical max-pooling
model with a local max-pooling parameter
for the three topologies of the convolutional neural networks
which we have introduced in Section \ref{se3}.

\begin{theorem}
	\label{th1}
	Let $d_1, d_2 \in \N$, let $n\in\N$ with $n>1$, and
	let $(\bX,Y)$, $(\bX_1,Y_1)$, \dots, $(\bX_n,Y_n)$
	be independent and identically distributed
	$[0,1]^{\{1, \dots, d_1\} \times \{1, \dots, d_2\}} \times \{0,1\}$-valued
	random variables. Assume that the a posteriori probability
	$\eta(\bx)=\PROB\{Y=1|\bX=\bx\}$ satisfies
	a hierarchical max-pooling model of finite order $d^*$ and level $l$ with feature constraint $\bb=(b_1,\dots,b_{l-1})$ and
        local max-pooling parameter $\bn=(n_1,\dots,n_{l-1})$ and smoothness constraint $p\in[1,\infty)$.
	Assume that %
	the image dimensions satisfy
	\begin{equation}
		\label{th1eq1}
		d_1=2^l\cdot m_1-1\quad d_2=2^l\cdot m_2-1
	\end{equation}
	for some $m_1,m_2\in\N\setminus\{1\}$.
        We set
        	\[
        	L_n =\left\lceil
        	c_1 \cdot n^{\frac{4}{2 \cdot (2\cdot p+4)}}
        	\right\rceil
        	\]
	for $c_1>0$ sufficiently large
	and select the parameters of our convolutional neural network function classes $\F_1\left(\btheta_1\right)$, $\F_2\left(\btheta_2\right)$ and $\F_3\left(\btheta_3\right)$ as follows:
	
	We set 
	\[
	z=\max\{b_1,\dots,b_{l-1}\}\cdot(L_n+1),\quad L=l,\quad\tilde{\bd}=(\tilde{d}_1,\tilde{d}_2)=\left(\frac{d_1-2^{l}+1}{\prod_{i=1}^{l-1}n_i},\frac{d_2-2^{l}+1}{\prod_{i=1}^{l-1}n_i}\right)
	\]
	and for $r\in\{1,\dots,L\}$ we choose
	\[
	k_r=2\cdot\max\{b_1,\dots,b_{l-1}\} + c_{2},\quad M_{r}=\frac{2^{r-1}}{\prod_{i=0}^{r-1}n_i}+1,\quad\bar{M}_r=2^{r-1}+1
	\]
	for $c_{2}\in \N$ sufficiently large and set $\bs=(n_1,\dots,n_{L-1})$,  
	and $s=n_1\cdot\ldots\cdot n_{L-1}$.
	Furthermore, we set
	\[\bar{z}=z+3\cdot\max\{k_1,\dots,k_L\}\cdot\max\{\log_2(n_1),\dots,\log_2(n_{L-1})\},\]
	$
	{\bk}=(k_1,\dots,k_L),
	$
	$
	\bar{\bk}=(2\cdot k_1+4,\dots,2\cdot k_L+4),
	$ ${\bM}=({M}_1,\dots,{M}_L)$, $\bar{\bM}=(\bar{M}_1,\dots,\bar{M}_L)$, 
	\[
	\btheta_1=(L,\bk,\bM,z,\bs,\tilde{\bd}),\quad\btheta_2=(L,\bar{\bk},{\bM},\bar{z},\bs,\tilde{\bd})\quad\text{and}\quad\btheta_3=(L,\bar{\bk},\bar{\bM},\bar{z},s,\tilde{\bd})
	\]
	We define the estimates $f_n^{(j)}$ $(j=1,2,3)$ as in Subsection \ref{subse37}. Then we have
	\begin{align*}
		&\PROB\{f_n^{(j)}(\bX) \neq Y\}
		-
		\min_{f: [0,1]^{\{1, \dots, d_1\} \times \{1, \dots, d_2\}} \rightarrow \{0,1\}}
		\PROB\{f(\bX) \neq Y\}\\
		&\leq
		c_{3}
		\cdot
		\sqrt{\log(d_1\cdot d_2)}
		\cdot
		(\log n)^2
		\cdot
n^{- \frac{p}{2\cdot p+4}},
	\end{align*}
	for all $j\in\{1,2,3\}$ and some constant $c_{3}>0$ which does not depend on $d_1$, $d_2$ and $n$.
\end{theorem}
\noindent
The proof is available in the supplement.\\

\noindent
{\bf Remark 7.}
	The rate of convergence in Theorem 1 does not depend on the image dimensions $d_1$ and $d_2$, hence under the assumptions on the a posteriori probability from Theorem 1 our convolutional neural network image classifiers are able to circumvent the curse of dimensionality.

\noindent
{\bf Remark 8.}
	The above parameters $L_n$, $L$, $\bM$, $z$, $\bar{z}$, $\bs$ and $s$  depend on the smoothness, respectively the level, or neighborhood sizes of the model of the a posteriori probability, which may be unknown in applications.
	Nevertheless, we can choose these parameters in a data-dependent way, e.g., by using a splitting of the sample approach as described in the next section.
For the classes $\F_1(\btheta_1)$ and $\F_2(\btheta_2)$, the number of possibilities for the
parameter
$\bs=(s_1,\dots,s_{L-1})$ is equal to the number of possible sizes of the local neighborhoods of the model of the a posteriori probability $n_1,\dots,n_{l-1}$. Since the function class $\F_3(\btheta_3)$ has only one subsampling layer and its parameter $s$ is calculated as the product of the neighborhood sizes $n_1,\dots,n_{l-1}$, the class has the advantage over the other two classes that there are much less possibilities for parameter combinations.

\noindent
{\bf Remark 9.}
The constant $c_3$ in Theorem 1 is different for the different function classes $\F_1(\btheta_1)$, $\F_2(\btheta_2)$ and $\F_3(\btheta_3)$. Since we show in Lemma 5 that $\F_1(\btheta_1)\subset\F_2(\btheta_2)\subset\F_3(\btheta_3)$ for parameters $\btheta_1$, $\btheta_2$ und $\btheta_3$ choosen as in Theorem 1, our proof of Theorem 1 yields
\begin{align*}
	&\PROB\{f_n^{(j)}(\bX) \neq Y\}
	-
	\min_{f: [0,1]^{\{1, \dots, d_1\} \times \{1, \dots, d_2\}} \rightarrow \{0,1\}}
	\PROB\{f(\bX) \neq Y\}\\
	&\leq
	c_{3}^{(j)}
	\cdot
	\sqrt{\log(d_1\cdot d_2)}
	\cdot
	(\log n)^2
	\cdot
	n^{- \frac{p}{2\cdot p+4}},
\end{align*}
for some constants $c_3^{(j)}>0$ for all $j\in\{1,2,3\}$ satisfying $c_3^{(1)}\leq c_3^{(2)}\leq c_3^{(3)}$.

\noindent
{\bf Remark 10.}
We can make the above assumption \eqref{th1eq1} for the image dimensions, since any image dimensions can be made into the above form by an appropriate zero padding.
\section{Application to simulated data}
\label{se5}
To analyze the finite sample size performance of our newly introduced image classifiers, we apply them to simulated and real data 
and compare their results with the classifier from  \cite{KoKrWa2020}, which has no pooling layers.
First, we describe how to generate the synthetic image data sets. A synthetically  generated image data set is given by finitely many realizations 
\[\mathcal D_N=\{(\bx_1,y_1),(\bx_2,y_2),\dots,(\bx_N,y_N)\}\]
 of a $[0,1]^{\{1,\dots,31\}\times\{1,\dots,31\}}\times\{0,1\}$-valued random variable $(\bX,Y)$. As described in Section 1 the matrix $\bX$ contains at position $(i,j)$ the grey scale value of the pixel of the image at the corresponding position and the value of $Y$ denotes the class of the image. In this example we use $d_1=31$ and $d_2=31$ for the image dimensions.
 Both classes of our synthetically generated images consist of the same geometric objects, namely a circle, an equilateral triangle and a square. The only difference between the two classes are the relative positions of the geometric objects to each other. 
First we determine the label $Y$ from a uniform distribution on $\{0,1\}$ and then determine a random image of the corresponding class.
We now describe how the images of the two classes are created. 
After the objects are theoretically defined on the cube $[0,31]^2$, they are downsampled to the grid $\{1,\dots,31\}\times\{1,\dots,31\}$ using the Python package \textit{Pillow}.

We start by randomly choosing the area and grey scale values of the three objects. For each object, the area is choosen independently and is uniformly distributed on the interval $[20,40]$. We determine the grey scale values of the three objects by randomly permuting the list $(0, \frac{1}{3}, \frac{2}{3})$ of three grey scale values.
Next, we explain how the positions of the objects in relation to each other are determined and then how this leads to the specific positions within the image area.
 For both classes, the objects are arranged approximately as corners of a large square, whose area is randomly choosen by the uniform distribution on the interval $[80,160]$.
For both classes, the bottom right corner defines the position of the square. The position of the circle is defined by the upper left corner for class 0 and by the lower left corner for class 1. Lastly, the position of the triangle is defined for class 0 by the lower left corner and for class 1 by the upper right corner.
 Since the objects should only have approximately these positions to each other, the positions are chosen randomly by a uniform distribution on a disk located on the corresponding corners, where the radius of the disk is a third of a side length of the large square.
  Once the positions to each other are determined in this way, 
  they
  are moved collectively within the image area by a uniform distribution on the 
  restricted image area so that all objects lie completely within the image. This leads us to the desired positions. As a last step, we add a scaled standard normally distributed noise value to each pixel value and truncate the resulting values to the interval $[0,1]$. We have shown some example images in Figure 5 and Figure 6.
\begin{figure}[h]
	\centering
	\includegraphics[width=.08\textwidth]{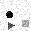}
	\includegraphics[width=.08\textwidth]{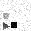}
	\includegraphics[width=.08\textwidth]{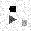}
	\includegraphics[width=.08\textwidth]{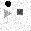}
	\includegraphics[width=.08\textwidth]{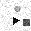}
	\includegraphics[width=.08\textwidth]{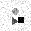}
	\includegraphics[width=.08\textwidth]{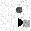}
	\includegraphics[width=.08\textwidth]{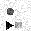}
	\includegraphics[width=.08\textwidth]{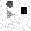}
	\includegraphics[width=.08\textwidth]{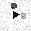}
	\\[\smallskipamount]
	\includegraphics[width=.08\textwidth]{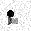}
	\includegraphics[width=.08\textwidth]{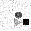}
	\includegraphics[width=.08\textwidth]{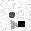}
	\includegraphics[width=.08\textwidth]{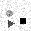}
	\includegraphics[width=.08\textwidth]{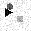}
	\includegraphics[width=.08\textwidth]{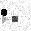}
	\includegraphics[width=.08\textwidth]{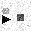}
	\includegraphics[width=.08\textwidth]{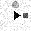}
	\includegraphics[width=.08\textwidth]{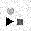}
	\includegraphics[width=.08\textwidth]{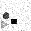}
	\caption{The two rows show images of class 0 of our classification task.}\label{fig:foobar}
\end{figure}
\begin{figure}[h]
	\centering
	\includegraphics[width=.08\textwidth]{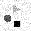}
	\includegraphics[width=.08\textwidth]{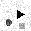}
	\includegraphics[width=.08\textwidth]{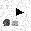}
	\includegraphics[width=.08\textwidth]{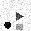}
	\includegraphics[width=.08\textwidth]{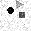}
	\includegraphics[width=.08\textwidth]{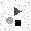}
	\includegraphics[width=.08\textwidth]{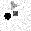}
	\includegraphics[width=.08\textwidth]{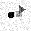}
	\includegraphics[width=.08\textwidth]{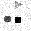}
	\includegraphics[width=.08\textwidth]{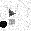}
	\\[\smallskipamount]
	\includegraphics[width=.08\textwidth]{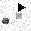}
	\includegraphics[width=.08\textwidth]{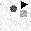}
	\includegraphics[width=.08\textwidth]{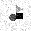}
	\includegraphics[width=.08\textwidth]{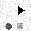}
	\includegraphics[width=.08\textwidth]{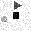}
	\includegraphics[width=.08\textwidth]{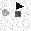}
	\includegraphics[width=.08\textwidth]{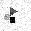}
	\includegraphics[width=.08\textwidth]{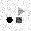}
	\includegraphics[width=.08\textwidth]{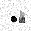}
	\includegraphics[width=.08\textwidth]{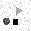}
	\caption{The two rows show images of class 1 of our classification task.}\label{fig:foobar}
\end{figure}%

Due to the fact that all of our classifiers depend on parameters that influence their behavior,  we choose these parameters data-dependently by splitting of the sample.
For this purpose we proceed as follows: We split our sample into a learning sample of size $n_{learn}=\left\lfloor\frac{4}{5}\cdot n\right\rfloor$ and a testing sample of size $n_{test}=n-n_{learn}$, and then use the learning sample to train our classifiers several times with the different choices for the parameters and use the testing sample to select the classifiers that minimize the empirical misclassification risk. Finally, we train the selected classifiers on the entire training set, consisting of the $n$ data points.

We slightly simplify the classifier presented in \cite{KoKrWa2020} by selecting it from class $\F_3(\btheta)$, setting the neighborhood size of the subsampling layer equal to 1, and adjusting the output bounds. Therefore, we set $\F_4(\btheta)\coloneqq\F_3(\btheta)$ and denote the classifier by $f_n^{(4)}$.
We now describe how precisely we choose the parameters of all classifiers, which may depend on the level $l$ and the local max-pooling parameter $\bn=(n_1,\dots,n_{l-1})$ of the hierarchical max-pooling model with local max-pooling parameter.
We adaptively choose $l\in\{3,4\}$ and $n_1,\dots,n_{l-1}\in\{2^0,2^1,\dots,2^{l-1}\}$ such that
	\begin{equation*}
	n_r\in\{2^0,\dots,2^r\}\quad\text{and}\quad\prod_{i=0}^{r}n_i\leq 2^{r}
\end{equation*}
is satisfied for all
$r\in\{1,\dots,l-1\}$ (where $n_0=1$).
Furthermore, we adaptively choose the number of channels in each convolutional layer from $k\in\{2,4,8\}$ and the number of layers in each convolutional block from $z\in\{1,2,3\}$. According to Theorem 1, the remaining parameters result respectively from the values for $l$ and $n_1,\dots,n_{l-1}$ as shown in Table 1. 
The parameters then are given by
\[
\btheta_j=(L,(k,\dots,k),\bM^{(j)},z,\bs^{(j)},\tilde{\bd}^{(j)})
\]
for our classifiers $f_n^{(j)}\in\F_j(\btheta_j)$ for $j\in\{1,2,3,4\}$.
\setlength{\tabcolsep}{5pt}
\begin{table}[h]
	\centering
	\arraycolsep=5pt\def\arraystretch{2.2}
	$\begin{array}{|c|c|c|c|c|}
		\hline
		j & L &\bM^{(j)} & \bs^{(j)}& \tilde{\bd}^{(j)}\\
		\hline
				{1} & l & \frac{2^{r-1}}{\prod_{i=0}^{r-1}n_i}+1\quad(r=1,\dots,L) &(n_1,\dots,n_{L-1})  & \left(\left\lceil\frac{d_1-2^{l}+1}{\prod_{i=1}^{l-1}n_i}\right\rceil,\left\lceil\frac{d_2-2^{l}+1}{\prod_{i=1}^{l-1}n_i}\right\rceil\right)\\
	{2} & l &\frac{2^{r-1}}{\prod_{i=0}^{r-1}n_i}+1\quad(r=1,\dots,L) &(n_1,\dots,n_{L-1}) &  \left(\left\lceil\frac{d_1-2^{l}+1}{\prod_{i=1}^{l-1}n_i}\right\rceil,\left\lceil\frac{d_2-2^{l}+1}{\prod_{i=1}^{l-1}n_i}\right\rceil\right)\\

	{3} & l &2^{r-1}+1\quad(r=1,\dots,L)   & n_1\cdot\ldots\cdot n_{L-1}& \left(\left\lceil\frac{d_1-2^{l}+1}{\prod_{i=1}^{l-1}n_i}\right\rceil,\left\lceil\frac{d_2-2^{l}+1}{\prod_{i=1}^{l-1}n_i}\right\rceil\right)\\

	{4} & l &{2^{r-1}+1}\quad(r=1,\dots,L) &  1& \left({d_1-2^{l}+1},{d_2-2^{l}+1}\right)\\		
	\hline	
	\end{array}$
\caption{Definition of the parameters that depend on the adaptively chosen values $l$ and $n_1,\dots,n_{l-1}$.}
\label{table1}
\end{table}
\noindent
We want to avoid overparameterization, so we only use the parameter combinations such that the total number of weights does not exceed the training sample size $n$.
For minimizing the empirical $L_2$ risk \eqref{eq:lqp}, we use the optimizer \textit{Adam} from the \textit{Keras} library.

For the evaluation of our classifiers we use their empirical misclassification risk on new independent test data, which is defined by
\begin{equation}
	\epsilon_{N}(f_n)=\frac{1}{N}\sum_{k=1}^{N}\IND_{\{f_n(\bx_{n+k})\neq y_{n+k}\}},%
	\label{eq:ac}
\end{equation}
where $f_n$ is the considered classifier based on the training set. The data points \[(\bx_{n+1},y_{n+1}),\dots,(\bx_{n+N},y_{n+N})\] are newly generated independent realizations of the random variable $(\bX,Y)$ with ${N=10^5}$. 
Table \ref{table2} shows the median and interquartile range (IQR) of the empirical misclassification risk \eqref{eq:ac} of our four estimates of 25 runs on 25 different independently generated data sets $\{(\bx_1,y_1),\dots,(\bx_{n+N},y_{n+N})\}$.
\begin{table}[H]
	\centering
	\begin{tabular}{|c|c|c|c|c|}
		\hline
		\textit{sample size} & $n=200$ & $n=400$ \\
		\hline
		\textit{approach} & median (IQR) & median (IQR)\\
		\hline
		\textit{$f_n^{(1)}$} & \textbf{0.15 (0.06)}  & \textbf{0.04 (0.05)} \\
		\textit{$f_n^{(2)}$} & 0.18 (0.19)  & 0.06 (0.03) \\
		\textit{$f_n^{(3)}$} & 0.49 (0.05)  & 0.27 (0.17)\\
		\textit{$f_n^{(4)}$} & 0.43 (0.08)  & 0.29 (0.20) \\
		\hline
	\end{tabular}
	\caption{Median and interquartile range of the empirical misclassification risk $\epsilon_N\left(f_n^{(i)}\right)$ for $i=1,2,3,4$.}
	\label{table2}
\end{table}
We observe that our convolutional neural network classifier $f_n^{(1)}$ with several local max-pooling layers outperforms the other convolutional neural network classifiers $f_n^{(2)}$, $f_n^{(3)}$ and $f_n^{(4)}$. Also its relative improvement with increasing sample size is larger than the improvement for the other classifiers. 
This supports
Remark 9, that the classifier $f_n^{(1)}$ has a convergence rate better than the classifier $f_n^{(2)}$ by some constant factor. We have the same observation concerning the relative improvement of the empirical misclassification risk with increasing sample size for the classifiers $f_n^{(2)}$ and $f_n^{(3)}$. The classifier without local pooling layers $f_n^{(4)}$ has the smallest relative improvement with increasing sample size, which suggests that it also has the slowest rate of convergence.
\section{Application to real images}
\label{se6}
In this section, we test our classifiers using two different datasets containing real images to show their practical relevance. We follow \cite{Kim2018} in our choice of datasets and training data.
We use the CIFAR-10 dataset (cf., \cite{Krizhevsky2009}) and the SVHN dataset (cf., \cite{Netzer2011}).
Both datasets contain 10 different classes of real images consisting of $32\times32$ pixels.
We have reduced the size of the images from $32\times32$ pixels to $31\times31$ pixels by removing the last column and row of the original images (because of our condition on the image dimensions from Theorem 1) and  because the images are in color, we have converted them to grey scale.
Since our theory is based on a binary image classification problem, we decide to use only two classes of each dataset, which we think best fit our model (only approximate relative distances of features are crucial for correct classification) and, on the other hand, represent a hard classification task.
For the CIFAR-10 dataset we consider the two classes `\textit{dog}' and `\textit{cat}' and for the SVHN dataset of house numbers we consider the classes `$4$' and `$9$'.
The original data sets then reduce to $10,000$ training images and $N=2,000$ test images in the case of the CIFAR-10 data set. For the SVHN dataset, we get $12,117$ training images and $N=4,118$ test images.

\begin{figure}[h]
	\centering
	\includegraphics[width=.08\textwidth]{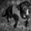}
	\includegraphics[width=.08\textwidth]{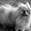}
	\includegraphics[width=.08\textwidth]{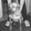}
	\includegraphics[width=.08\textwidth]{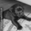}
	\includegraphics[width=.08\textwidth]{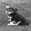}
	\includegraphics[width=.08\textwidth]{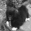}
	\includegraphics[width=.08\textwidth]{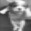}
	\includegraphics[width=.08\textwidth]{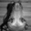}
	\includegraphics[width=.08\textwidth]{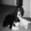}
	\includegraphics[width=.08\textwidth]{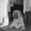}
	\\[\smallskipamount]
	\includegraphics[width=.08\textwidth]{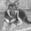}
	\includegraphics[width=.08\textwidth]{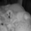}
	\includegraphics[width=.08\textwidth]{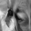}
	\includegraphics[width=.08\textwidth]{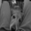}
	\includegraphics[width=.08\textwidth]{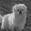}
	\includegraphics[width=.08\textwidth]{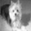}
	\includegraphics[width=.08\textwidth]{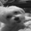}
	\includegraphics[width=.08\textwidth]{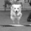}
	\includegraphics[width=.08\textwidth]{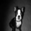}
	\includegraphics[width=.08\textwidth]{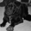}
	\\[\smallskipamount]
	\includegraphics[width=.08\textwidth]{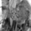}
	\includegraphics[width=.08\textwidth]{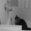}
	\includegraphics[width=.08\textwidth]{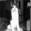}
	\includegraphics[width=.08\textwidth]{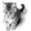}
	\includegraphics[width=.08\textwidth]{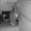}
	\includegraphics[width=.08\textwidth]{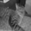}
	\includegraphics[width=.08\textwidth]{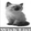}
	\includegraphics[width=.08\textwidth]{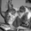}
	\includegraphics[width=.08\textwidth]{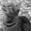}
	\includegraphics[width=.08\textwidth]{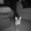}
	\\[\smallskipamount]
	\includegraphics[width=.08\textwidth]{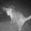}
	\includegraphics[width=.08\textwidth]{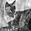}
	\includegraphics[width=.08\textwidth]{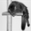}
	\includegraphics[width=.08\textwidth]{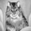}
	\includegraphics[width=.08\textwidth]{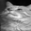}
	\includegraphics[width=.08\textwidth]{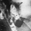}
	\includegraphics[width=.08\textwidth]{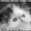}
	\includegraphics[width=.08\textwidth]{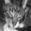}
	\includegraphics[width=.08\textwidth]{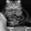}
	\includegraphics[width=.08\textwidth]{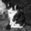}
	\caption{The first two rows show some images of the dogs and the lower two rows show images of the cats of the grey scaled CIFAR-10 data set.}\label{fig:foobar}
\end{figure}
\begin{figure}[h]
	\centering
	\includegraphics[width=.08\textwidth]{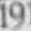}
	\includegraphics[width=.08\textwidth]{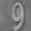}
	\includegraphics[width=.08\textwidth]{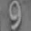}
	\includegraphics[width=.08\textwidth]{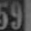}
	\includegraphics[width=.08\textwidth]{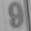}
	\includegraphics[width=.08\textwidth]{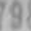}
	\includegraphics[width=.08\textwidth]{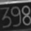}
	\includegraphics[width=.08\textwidth]{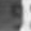}
	\includegraphics[width=.08\textwidth]{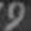}
	\includegraphics[width=.08\textwidth]{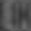}
	\\[\smallskipamount]
	\includegraphics[width=.08\textwidth]{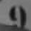}
	\includegraphics[width=.08\textwidth]{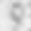}
	\includegraphics[width=.08\textwidth]{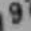}
	\includegraphics[width=.08\textwidth]{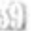}
	\includegraphics[width=.08\textwidth]{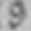}
	\includegraphics[width=.08\textwidth]{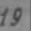}
	\includegraphics[width=.08\textwidth]{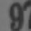}
	\includegraphics[width=.08\textwidth]{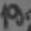}
	\includegraphics[width=.08\textwidth]{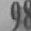}
	\includegraphics[width=.08\textwidth]{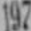}
	\\[\smallskipamount]
	\includegraphics[width=.08\textwidth]{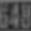}
	\includegraphics[width=.08\textwidth]{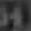}
	\includegraphics[width=.08\textwidth]{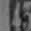}
	\includegraphics[width=.08\textwidth]{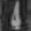}
	\includegraphics[width=.08\textwidth]{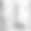}
	\includegraphics[width=.08\textwidth]{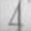}
	\includegraphics[width=.08\textwidth]{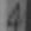}
	\includegraphics[width=.08\textwidth]{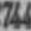}
	\includegraphics[width=.08\textwidth]{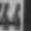}
	\includegraphics[width=.08\textwidth]{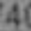}
	\\[\smallskipamount]
	\includegraphics[width=.08\textwidth]{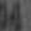}
	\includegraphics[width=.08\textwidth]{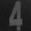}
	\includegraphics[width=.08\textwidth]{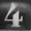}
	\includegraphics[width=.08\textwidth]{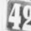}
	\includegraphics[width=.08\textwidth]{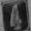}
	\includegraphics[width=.08\textwidth]{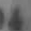}
	\includegraphics[width=.08\textwidth]{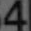}
	\includegraphics[width=.08\textwidth]{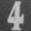}
	\includegraphics[width=.08\textwidth]{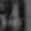}
	\includegraphics[width=.08\textwidth]{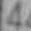}
	\caption{The first two rows show some images of the nines and the lower two rows show some images of the fours of the grey scaled SVHN data set.}\label{fig:foobar}
\end{figure}
\noindent
We choose the parameter grids of our four classifiers as in Section 5 and also choose $n_{learn}=\left\lfloor\frac{4}{5}\cdot n\right\rfloor$ and $n_{test}=n-n_{learn}$. 
We randomly select $n$ training images from the $10,000$ and $12,117$ original training images, respectively and evaluate our classifiers using the corresponding $N$ test images and repeat this several times. Table \ref{table3} then shows the %
median and interquartile range (IQR) of 25 runs
of the empirical misclassification risk \eqref{eq:ac} of 25 runs (where the $n$ training images are drawn with replacement).
\begin{table}[h]
	\centering
	\begin{tabular}{|c|c|c|c|c|}
		\hline
		\textit{data} &  \multicolumn{2}{c|}{CIFAR-10}& \multicolumn{2}{c|}{SVHN}\\
		\hline
		\textit{sample size} & $n=200$ & $n=400$ & $n=200$ & $n=400$\\
		\hline
		\textit{approach} & median (IQR) & median (IQR)& median (IQR)& median(IQR)\\
		\hline
		\textit{$f_n^{(1)}$} & {0.48 } (0.02)&{0.47} (0.03) &  \textbf{0.31 (0.08)}&\textbf{0.25 (0.07)} \\
		\textit{$f_n^{(2)}$} & 0.50 (0.02)&0.48 (0.03)& 0.35 (0.05)& 0.28 (0.04)\\
		\textit{$f_n^{(3)}$} &0.48 (0.03)&{0.47} (0.03) &  0.40 (0.03)&0.38 (0.04) \\
		\textit{$f_n^{(4)}$} &\textbf{0.47 (0.04)}& \textbf{0.45 (0.03)} &  0.39 (0.02)&0.39 (0.03) \\
		\hline
	\end{tabular}
	\caption{Median and interquartile range (IQR) of the empirical misclassification risk $\epsilon_N\left(f_n^{(i)}\right)$ for $i=1,2,3,4$ based on the presented grey scaled subsets of the CIFAR-10 and SVHN data sets.}
	\label{table3}
\end{table}
For the CIFAR-10 dataset, the classifier $f_n^{(4)}$ outperforms the other classifiers, even though we obtain similar errors for all four approaches.
Considering the SVHN dataset, the classifier $f_n^{(1)}$ outperforms the other approaches, where the errors of all classifiers relative to each other behaving roughly as for the simulated data (the classifiers $f_n^{(1)}$ and $f_n^{(2)}$ perform much better than the classifiers $f_n^{(3)}$ and $f_n^{(4)}$ and also have a better relative improvement with increasing sample size). Hence, we assume that, at least for the SVHN dataset, our assumption of local max-pooling on the a posteriori probability seems plausible.
\section{Summary}
In this paper, we analyzed various convolutional neural network image classifiers in a hierarchical max-pooling model with local max-pooling. This model extends the hierarchical max-pooling model of \cite{KoKrWa2020} by allowing features of an image that are combined hierarchically to have variable relative distances towards each other. We introduced three convolutional neural network architectures and investigated the convergence rate of the distance between the expected misclassification risk of the classifier and the optimal misclassification risk in each case. All three classes of convolutional neural networks include some kind of local pooling and achieve the same convergence rate in our model, which does not depend on the image dimensions, thus circumventing the curse of dimensionality. Hence, we provide a theoretical explanation why some kind of local pooling is useful in some image classification applications. By applying our classifiers to simulated and real data and analyzing their finite sample size performance, we were also able to support our theoretical results.
\setcitestyle{numbers} 
\bibliographystyle{kohler}
\bibliography{Literatur}
\setcitestyle{authoryear} 
 
 \newpage
 \begin{center}
 	
 	{\LARGE \bf
 		Supplementary material to ``Analysis of convolutional neural network image classifiers in a hierarchical max-pooling model with additional local pooling''
 	}
 \end{center}
 \appendix
 \noindent
 The supplement contains the complete proof of Theorem 1 and all auxiliary results necessary for it.
 \label{se7}
 In the proof of Theorem \ref{th1} we proceed as in the proof of
 Theorem 1 in Kohler, Krzy\.zak and Walter (2020), i.e., we
 relate the misclassification error of our plug-in estimate
 to the $L_2$ error of the corresponding least squares estimates,
 bound its error via empirical process theory, and derive
 bounds on the approximation error and the covering number.
 The bound on the covering number is a straightforward
 extension of the corresponding bound in 
 Kohler, Krzy\.zak and Walter (2020),
 but the analysis of the approximation error needs some
 technical difficult extensions of the results from 
 Kohler, Krzy\.zak and Walter (2020).
 
 \section{Proof of Theorem 1}
 W.l.o.g. we assume that $n$ is so large that $c_{4} \cdot \log n \geq
 2$
 holds for some constant $c_4>0$.
 Then $z\geq1/2$ holds if and only if $T_{c_{4} \cdot \log n} z\geq1/2$
 holds, and consequently we have
 \[
 f_n^{(j)}(\bx)=
 \begin{cases}
 	1, & \mbox{if } T_{c_{4} \cdot \log n} \eta_n^{(j)}(\bx) \geq \frac{1}{2} \\
 	0, & \mbox{elsewhere}
 \end{cases}
 \]
 $(j \in \{1,2,3\})$.
 Because of Lemma \ref{le10} from Section E we then have
 \begin{equation*}
 	\PROB\{f_n(\bX)^{(j)} \neq Y\}
 	-
 	\PROB\{ f^*(\bX) \neq Y\}
 	\leq
 	2 \cdot
 	\sqrt{
 		\EXP \left\{
 		\int |T_{c_{4} \cdot \log n}\eta_n^{(j)}(\bx)-\eta(\bx)|^2 \, \PROB_{\bX}(d\bx)
 		\right\}
 	}
 \end{equation*}
 for $j\in\{1,2,3\}$
 and therefore it suffices to show
 \[
 \EXP \int |T_{c_{4} \cdot \log n}\eta_n^{(j)}(\bx)-\eta(\bx)|^2  \PROB_X(d\bx)
 \leq
 c_{11}
 \cdot
 \log(d_1\cdot d_2)
 \cdot
 (\log n)^4
 \cdot
 n^{- \frac{2p}{2p+4}}
 \]
 for $j \in \{1,2,3\}$.
 Lemma \ref{le5} from Section B yields us 
 \begin{equation}
 	\label{ptheq1}
 	\F_1(\btheta_1)\subset\F_2(\btheta_2)\subset\F_3(\btheta_3).
 \end{equation}
 and since the covering number and the infimum are monotonic, we get together with Lemma \ref{le11} from Section E
 \begin{eqnarray*}
 	&&
 	\EXP \int |T_{c_{4} \cdot \log n} \eta_n^{(j)}(\bx)-\eta(\bx)|^2  \PROB_\bX(d\bx)
 	\\
 	&&
 	\leq
 	\frac{c_{12} \cdot (\log(n))^2 \cdot \sup_{\bx_1^n } \left(\log\left(
 		\mathcal{N}_1 \left(\frac{1}{n\cdot c_{4} \log(n)}, T_{c_{4} \log(n)} \F_{3}(\btheta_3), \bx_1^n\right)
 		\right)+1\right)}{n}\notag\\
 	&&\quad + 2 \cdot \inf_{f \in \F_{1}(\btheta_1)} \int |f(\bx)-\eta(\bx)|^2 {\PROB}_\bX (d\bx)
 \end{eqnarray*}
for $j\in\{1,2,3\}$.
 Application of our bound on the covering number from Lemma \ref{le9} from Section D yields
 \begin{eqnarray*}
 	&&
 	\frac{c_{5} \cdot (\log(n))^2 \cdot \sup_{\bx_1^n} \left(\log\left(
 		\mathcal{N}_1 \left(\frac{1}{n\cdot c_{4} \log(n)}, T_{c_{4} \log(n)} \F_{3}(\btheta_3), \bx_1^n\right)
 		\right)+1\right)}{n}
 	\\
 	&&
 	\leq c_{6} \cdot \frac{
 		\log(d_1 \cdot d_2)
 		\cdot (\log n)^3 \cdot z^2 \cdot \log z}{n}
 	\\
 	&&
 	\leq c_{7}\cdot\log(d_1\cdot d_2)\cdot(\log n)^4\cdot n^{-\frac{2\cdot p}{2\cdot p+4}}
 	.
 \end{eqnarray*}
 Next we derive a bound on the approximation error
 \[
 \inf_{f \in  \F_1(\btheta_1)} \int |f(\bx)-\eta(\bx)|^2 {\PROB}_\bX (d\bx).
 \]
 For $k \in \{1, \dots, l\}$ and $s \in \{1, \dots, b_k\}$ let
 $g_{net,k,s}: \R^4 \rightarrow \R$ be the neural network from Kohler and Langer (2021) (cf., Lemma \ref{le12} from Section E) with $L_n$ layers and $c_2$ neurons per layer
 which satisfies
 \[
 \| g_{k,s}-g_{net,k,s}\|_{[-2,2]^4, \infty}
 \leq c_{8} \cdot L_n^{-\frac{2p}{4}}.
 \]
 Since the functions $g_{k,s}$ are $[0,1]$--valued, we are able to choose $c_1$ in the definition of $L_n$ sufficiently large such that
 \[\|g_{net,k,s}\|_{[-2,2]^4,\infty}\leq1+c_{8} \cdot L_n^{-\frac{2p}{4}}\leq2\]
 for all $k \in \{1, \dots, l\}$ and $s \in \{1, \dots, b_k\}$.
 We define $\bar{m} \in \F_1(\btheta_1)$ as in Lemma \ref{le7} from Section C. Then Lemma
 \ref{le6} and Lemma \ref{le7} from Section C
 imply
 \begin{eqnarray*}
 	\inf_{f \in \F_1(\btheta_1)} \int |f(\bx)-\eta(\bx)|^2 {\PROB}_\bX (d\bx)
 	&\leq&
 	c_{9} \cdot \max_{k \in \{1, \dots, l\}, s \in \{1, \dots,b_k\}}
 	\| g_{k,s}-g_{net,k,s}\|_{[-2,2]^4,\infty}^2
 	\\
 	&\leq & c_{10} \cdot L_n^{-p}.
 \end{eqnarray*}
 Putting in the value of $L_n$ and summarizing
 the above results, the proof is complete.
 
 \hfill $\Box$
 \section{A connection between several classes of convolutional neural networks}
 The aim of this section is to prove equation \eqref{ptheq1}, which we used in the proof of Theorem 1. In order to show that we may represent a convolutional neural network with several max-pooling layers (as in $\F_1(\btheta_1)$) as a convolutional neural network with several subsampling layers (as in $\F_2(\btheta_2)$) we will use the following result.
 \begin{lemma}
 	\label{le1}
 	Let $k,i_1,i_2\in\N$, set $I=\{1,\dots,i_1\}\times\{1,\dots,i_2\}$, let
 	\[f:\R^{\{1,\dots,d_1\}\times\{1,\dots,d_2\}}\rightarrow\R_+^{I\times\{1,\dots,2\cdot k+4\}}\]
 	be a function. Let $n\in\N_0$ and let $M\in\N$ with $M\geq 2^{n-1}+1$.
 	Then there exist a convolutional block 
 	\[
 	o^{(z)}_{(2\cdot k+4,2\cdot k+4),M}:\R^{I\times\{1,\dots,2\cdot k+4\}}\rightarrow\R^{I\times\{1,\dots,2\cdot k+4\}}
 	\]
 	defined as in Subsection 3.2 with
 	$z=3\cdot n\cdot k$ layers
 	such that
 	\[
 	\Big(\big(f_{sub}^{(2^n)}\circ o^{(z)}_{(2\cdot k+4,2\cdot k+4),M}\circ f\big)(\bx)\Big)_{(i,j),s}
 	=\Big(\big(f_{max}^{(2^n)}\circ f\big)(\bx)\Big)_{(i,j),s}
 	\]
 	holds for all $s\in\{1,\dots,k\}$, $\bx\in[0,1]^{\{1,\dots,d_1\}\times\{1,\dots,d_2\}}$ and $(i,j)\in\left\{1,\dots,\left\lceil{i_1}/{2^n}\right\rceil\right\}\times\left\{1,\dots,\left\lceil{i_2}/{2^n}\right\rceil\right\}$.
 \end{lemma}
 In the proof of Lemma \ref{le1} we will apply following
 auxiliary result.
 \begin{lemma}
 	\label{le2}
 	Let $g_{net}:\R^4\rightarrow\R$ be a
 	standard feedforward
 	neural network with $L_{net}\in\N$ hidden layers and
 	$r_{net}\in\N$ neurons per hidden layer (see Subsection 3.1).
 	Let
 	\[f:\R^{\{1,\dots,d_1\}\times\{1,\dots,d_2\}}\rightarrow\R^{I\times\{1,\dots,t+r_{net}\}}\]
 	be a function, where $d_1,d_2,i_1,i_2,t\in \N$ and $I=\{1,\dots,i_1\}\times\{1,\dots,i_2\}$.
 	Furthermore, let $s_{1},\dots,s_{5} \in \{1,
 	\dots, t\}$ and $\delta\in\N$.
 	Then there exist a convolutional block 
 	\[o_{(t+r_{net},t+r_{net}),\delta+1}^{(L_{net}+1)}:\R^{I\times\{1,\dots,t+r_{net}\}}\rightarrow\R^{I\times\{1,\dots,t+r_{net}\}}\]
 	defined as in Subsection 3.2 with arbitrary weights in channels 
 	\[\{1,\dots,t\}\setminus\{s_5\}\]
 	such that
 	\begin{equation}
 		\label{le2eq1}
 		\begin{split}
 			&
 			\Big(\big(o^{(L_{net}+1)}_{(t+r_{net},t+r_{net}),\delta+1}\circ f\big)(\bx)\Big)_{(i,j),s_{5}}
 			\\
 			&
 			=
 			\sigma\Big(g_{net} \Big(
 			\big(f(\bx)\big)_{(i,j),s_1},\big(f(\bx)\big)_{(i+\delta,j),s_2},
 			\big(f(\bx)\big)_{(i,j+\delta),s_3},
 			\big(f(\bx)\big)_{(i+\delta,j+\delta),s_4}\Big)\Big)
 		\end{split}
 	\end{equation}
 	for all $(i,j)\in I$ and all $\bx\in[0,1]^{\{1,\dots,d_1\}\times\{1,\dots,d_2\}}$
 	, where we set $\big(f(\bx)\big)_{(i,j),s}=0$ for $(i,j)\notin I$.
 	
 \end{lemma}
 \noindent 
 {\bf Proof.}
 We assume that the standard feedforward neural network $g_{net}$ is given by
 \[
 g_{net}(\bx) = \sum_{i=1}^{r_{net}} w_{1,i}^{(L_{net})}g_i^{(L_{net})}(\bx) + w_{1,0}^{(L_{net})},
 \]
 where $g_i^{(L_{net})}$ is recursively defined by
 \[
 g_i^{(r)}(\bx) = \sigma\left(\sum_{j=1}^{r_{net}} w_{i,j}^{(r-1)} g_j^{(r-1)}(\bx) + w_{i,0}^{(r-1)} \right)
 \]
 for
 $i \in \{1,\dots,r_{net}\}$,
 $r \in \{2, \dots, L_{net}\}$,
 and
 \[
 g_i^{(1)}(\bx) = \sigma \left(\sum_{j=1}^4 w_{i,j}^{(0)} x^{(j)} +
 w_{i,0}^{(0)} \right)
 \quad (i \in \{1, \dots, r_{net}\}).
 \]
 We now choose the weights
 of the convolutional block 
 \[o_{(t+r_{net},t+r_{net}),\delta+1}^{(L_{net}+1)}=\big(o_{(t+r_{net},t+r_{net}),\delta+1,\bw_{L_{net}+1}}\circ\dots\circ o_{(t+r_{net},t+r_{net}),\delta+1,\bw_1}\big)\]
 by using the weights of $g_{net}$. As in Section 3.1, the weights of $o_{(t+r_{net},t+r_{net}),\delta+1}^{(L_{net}+1)}$ are given by
 \[
 \bw_r= \left(
 \left(
 w^{(r)}_{i,j,s_{2,1},s_{2,2}}
 \right)_{
 	1 \leq i,j \leq\delta+1, s_{2,1} \in \{1, \dots,t+r_{net}\}, s_{2,2} \in \{1, \dots,t+r_{net}\}
 }
 ,
 \left(
 w^{(r)}_{s_{2,2}}
 \right)_{
 	s_{2,2} \in \{1, \dots,t+r_{net}\}
 }         
 \right)
 \]
 for $r=1,\dots,L_{net}+1$. In the sequel we set 
 \[o^{(r)}=o_{(t+r_{net},t+r_{net}),\delta+1,\bw_{r}}\circ\dots\circ o_{(t+r_{net},t+r_{net}),\delta+1,\bw_{1}}\]
 for $r=1,\dots,L_{net}+1$ and $\big(f(\bx)\big)_{(i,j),s}=0$ if $(i,j)\notin I$.
 In the first layer in channel $t+i$ we set
 \[w_{t_1,t_2,s,t+i}^{(1)}=0\]
 for all $t_1,t_2\notin\{1,\delta+1\}$ and all $s\notin\{s_1,\dots,s_4\}$ and choose the only nonzero weights by
 \begin{equation*}
 	\begin{split}
 		&w_{1,1,s_{1},t+i}^{(1)}=w_{i,1}^{(0)},\\
 		&w_{1,\delta+1,s_{3},t+i}^{(1)}=w_{i,3}^{(0)},
 	\end{split}
 	\quad\quad
 	\begin{split}
 		&w_{\delta+1,1,s_{2},t+i}^{(1)}=w_{i,2}^{(0)},\\
 		&w_{\delta+1,\delta+1,s_{4},t+i}^{(1)}=w_{i,4}^{(0)},
 	\end{split}
 	\label{eq:}
 \end{equation*}
 and $w_{t+i}^{(1)}=w_{i,0}^{(0)}$ for $i\in\{1, \dots,r_{net}\}$.
 Then we have
 \begin{align}
 	\label{le6eq8}
 	\begin{split}
 		&\big((o^{(1)}\circ f)(\bx)\big)_{(i_2,j_2),t+i}\\
 		&=
 		\sigma \left(
 		\sum_{s=1}^{t+r_{net}}
 		\sum_{\substack{t_1,t_2 \in \{1, \dots,\delta+1\}
 				\\
 				(i_2+t_1-1,j_2+t_2-1)\in I
 			}
 		}
 		w_{t_1,t_2,s,t+i}^{(1)}
 		\cdot
 		(f(\bx))_{(i_2+t_1-1,j_2+t_2-1),s}
 		+
 		w_{t+i}^{(1)}
 		\right)\\
 		&=
 		\sigma \Bigg(
 		w_{i,1}^{(0)}
 		\cdot
 		(f(\bx))_{(i_2,j_2),s_1}
 		+w_{i,2}^{(0)}
 		\cdot
 		(f(\bx))_{(i_2+\delta,j_2),s_2}+w_{i,3}^{(0)}
 		\cdot
 		(f(\bx))_{(i_2,j_2+\delta),s_3}\\
 		&\hspace{7cm}+w_{i,4}^{(0)}
 		\cdot
 		(f(\bx))_{(i_2+\delta,j_2+\delta),s_4}
 		+
 		w_{i,0}^{(0)}
 		\Bigg)\\
 		&=g_{i}^{(1)}
 		\Big(
 		(f(\bx))_{(i_2,j_2),s_1}, (f(\bx))_{(i_2+\delta,j_2),s_2}, 
 		(f(\bx))_{(i_2,j_2+\delta),s_3}, (f(\bx))_{(i_2+\delta,j_2+\delta),s_4}
 		\Big)
 	\end{split}
 \end{align}
 for all $i\in\{1,\dots,r_{net}\}$, $(i_2,j_2)\in I$ and all $\bx\in[0,1]^{\{1,\dots,d_1\}\times\{1,\dots,d_2\}}$.
 In layers $r \in \{2, \dots,L_{net}\}$ in channel $t+i$ we set
 \[w_{t_1,t_2,s,t+i}^{(r)}=0\]
 for all $(t_1,t_2)\neq(1,1)$ and all $s\in\{1,\dots,t\}$
 and choose the only nonzero weights by
 \[
 w_{1,1,t+j,t+i}^{(r)}=w_{i,j}^{(r-1)},\quad w_{t+i}^{(r)}=w_{i,0}^{(r-1)}
 \quad
 (j \in  \{1, \dots, r_{net}\})
 \]
 for $i\in\{1,\dots,r_{net}\}$.
 Thus we obtain
 \begin{align*}
 	&\big(o^{(r)}\circ f(\bx)\big)_{(i_2,j_2),t+i}\\
 	&=\sigma\Bigg(
 	\sum_{j=1}^{r_{net}}w_{i,j}^{(r-1)}\cdot\big(o^{(r-1)}\circ f(\bx)\big)_{(i_2,j_2),t+j}+w_{i,0}^{(r-1)}\Bigg)
 \end{align*}
 for all $i\in\{1,\dots,r_{net}\}$, $r\in\{2,\dots,L_{net}\}$, $(i_2,j_2)\in I$ and all $\bx\in[0,1]^{\{1,\dots,d_1\}\times\{1,\dots,d_2\}}$. Then we get by equation \eqref{le6eq8} and the definition of $g_{i}^{(r)}$ that
 \begin{align*}
 	&\big(o^{(r)}\circ f(\bx)\big)_{(i_2,j_2),t+i}\\
 	&=g_{i}^{(r)}
 	\Big(
 	(f(\bx))_{(i_2,j_2),s_1}, (f(\bx))_{(i_2+\delta,j_2),s_2}, 
 	(f(\bx))_{(i_2,j_2+\delta),s_3}, (f(\bx))_{(i_2+\delta,j_2+\delta),s_4}
 	\Big)
 \end{align*}
 for all $i\in\{1,\dots,r_{net}\}$, $r\in\{2,\dots,L_{net}\}$, $(i_2,j_2)\in I$ and all $\bx\in[0,1]^{\{1,\dots,d_1\}\times\{1,\dots,d_2\}}$.
 Now in layer $L_{net}+1$ in channel $s_{5}$ we set
 \[w_{t_1,t_2,s,s_{5}}^{(L_{net}+1)}=0\]
 for all $(t_1,t_2)\neq(1,1)$ and all $s\in\{1,\dots,t+r_{net}\}$
 and choose the only nonzero weights by
 \[
 w_{1,1,t+i,s_{5}}^{(L_{net}+1)}=w_{1,i}^{(L_{net})}\quad\mbox{ and }\quad w_{s_{5}}^{(L_{net}+1)}=w_{1,0}^{(L_{net})}
 \]
 for $i\in  \{1, \dots, r_{net}\}$.
 Consequently, we obtain
 \begin{align*}
 	\big(o_{(t+r_{net},t+r_{net}),\delta+1}^{(L_{net}+1)}\circ f(\bx)\big)_{(i_2,j_2),s_{5}}
 	&=
 	\sigma \left(
 	\sum_{i=1}^{r_{net}} w_{1,i}^{(L_{net})}
 	\cdot
 	\big(o^{(L_{net})}\circ f(\bx)\big)_{(i_2,j_2),t+i}
 	+ w_{1,0}^{(L_{net})}
 	\right)
 	\\
 	&=
 	\sigma \Bigg(
 	g_{net}
 	\Big(
 	(f(\bx))_{(i_2,j_2),s_1},
 	(f(\bx))_{(i_2+\delta,j_2),s_2},\\
 	&\hspace{2.5cm}
 	(f(\bx))_{(i_2,j_2+\delta),s_3}, (f(\bx))_{(i_2+\delta,j_2+\delta),s_4}
 	\Big)
 	\Bigg)
 \end{align*}
 for all $(i_2,j_2)\in I$ and all $\bx\in[0,1]^{\{1,\dots,d_1\}\times\{1,\dots,d_2\}}$.
 \hfill $\Box$\\

 \noindent
 {\bf Proof of Lemma \ref{le1}.}
 Without loss of generality we can assume $n>0$ (in the case $n=0$ the assertion is trivial because $f_{sub}^{(1)}=f_{max}^{(1)}$).
 In the proof we will use the fact that for $x\geq0$ we have
 \[\sigma(x)=\max\{x,0\}=x\]
 which enables us to 
 propagate a nonnegative value $x_{(i,j),s_1}\geq0$ computed in a layer of a convolutional neural network in channel $s_1$ at position $(i,j)$ to the next convolutional layer by 
 \begin{equation}
 	\label{ple1eq1}
 	\big(o_{(k',k),M,\bw_1}(\bx)\big)_{(i,j),s_2}=
 	\sigma\Big(
 	x_{(i,j),s_1}
 	\Big)
 	=x_{(i,j),s_1}
 \end{equation}
 with a corresponding weight vector $\bw_1$ whose weights are chosen accordingly from the set $\{0,1\}$ in channel $s_2$.
 Furthermore, we will use an auxiliary network $g_{max}:\R^4\rightarrow\R$ which calculates the maximum of its arguments in case they are nonnegative.
 Therefore we note that
 for $a\geq0$ and $b\in\R$ we have 
 \begin{equation}
 	\label{ple2eq2}
 	\max\{a,b\}=\max\{b-a,0\}+\max\{a,0\}=\sigma(b-a)+\sigma(a).
 \end{equation}
 Then we define the network $g_{max}$ with two hidden layers and at most four neurons per layer by
 \begin{align*}
 	g_{max}(\bx)=\sigma\Big({\sigma(x_{2}-x_{1})+\sigma(x_{1})}-\big({\sigma(x_{4}-x_{3})+\sigma(x_{3})}\big)\Big)+\sigma\Big({\sigma(x_{4}-x_{3})+\sigma(x_{3})}\Big).
 \end{align*}
 For $\bx\in\R_+^{4}$ equation \eqref{ple2eq2} then implies
 \[
 g_{max}(\bx)=\max\{\max\{x_{1},x_{2}\},\max\{x_{3},x_{4}\}\}=\max\{x_{1},x_{2},x_{3},x_{4}\}.
 \]
 The idea is to use Lemma 2 several times with the network $g_{net}=g_{max}$ and with appropriate growing values for $\delta$ to mimic the maximum computation of the local max-pooling layer.
 Since we want to choose the weight matrices $\bw_1,\dots,\bw_z$ of our convolutional block $o^{(z)}_{(2\cdot k+4,2\cdot k+4),M}$ layer by layer we define the function \[o^{(0)}:[0,1]^{\{1,\dots,d_1\}\times\{1,\dots,d_2\}}\rightarrow\R_+^{I\times\{1,\dots,2\cdot k+4\}}\]
 by
 $
 o^{(0)}=f
 $
 and define recursively the functions $o^{(t)}:[0,1]^{\{1,\dots,d_1\}\times\{1,\dots,d_2\}}\rightarrow\R^{I\times\{1,\dots,2\cdot k+4\}}$ by
 \[
 o^{(t)}=o_{(2\cdot k+4,2\cdot k+4),M,\bw_t}\circ o^{(t-1)}
 \]
 for $t=1,\dots,z$.
 We show that we can choose the weights $\bw_1,\dots,\bw_z$ of $o^{(z)}_{(2\cdot k+4,2\cdot k+4),M}$ such that the following property $(\ast)$ holds for all $r\in\{0,\dots,n\}$ by induction on $r$:
 \begin{itemize}
 	\item [($\ast$)]
 	For all $s\in\{1,\dots,k\}$, $(i,j)\in I$ and $\bx\in[0,1]^{\{1,\dots,d_1\}\times\{1,\dots,d_2\}}$ it holds that
 	\begin{equation*}
 		\label{propl8}
 		\Big(o^{(r\cdot 3\cdot k)}(\bx)\Big)_{(i,j),s}=\max_{(i,j)\in\{i,\dots,i+2^r-1\}\times\{j,\dots,j+2^r-1\}\cap I}\big(f(\bx)\big)_{(i,j),s}.
 	\end{equation*}
 \end{itemize}
 For the case $r=0$ the assertion follows due to the definition of $o^{(0)}$.
 Now assume property ($\ast$) is true for some $r\in\{0,\dots,n-1\}$.
 The idea is to use successively Lemma 2 for the computation of each network 
 \begin{equation}
 	\label{ple8eq3}
 	\begin{split}
 		&\sigma\Bigg({g}_{max}\Big(
 		\big(o^{(r\cdot 3\cdot k)}(\bx)\big)_{(i,j),s},
 		\big(o^{(r\cdot 3\cdot k)}(\bx)\big)_{(i+2^r,j),s},
 		\\&\hspace{2cm}
 		\big(o^{(r\cdot 3\cdot k)}(\bx)\big)_{(i,j+2^r),s},
 		\big(o^{(r\cdot 3\cdot k)}(\bx)\big)_{(i+2^r,j+2^r),s}\Big)\Bigg)
 	\end{split}
 \end{equation} 
 for $s=1,\dots,k$ and store the computed values 
 in the corresponding channels 
 \[1,\dots, k\]
 by propagating computed values to the next layer using equation \eqref{ple1eq1}.
 By equation \eqref{ple1eq1} and because of the induction hypothesis ($\ast$) we can choose the weights 
 \[
 \bw_{r\cdot3\cdot k+1},\dots,\bw_{(r+1)\cdot3\cdot k}
 \]
 of our convolutional block in the channels 
 \[k+1,\dots,2\cdot k\]
 such that
 for all $t\in\{r\cdot3\cdot k+1,\dots,(r+1)\cdot3\cdot k\}$, $s\in\{1,\dots,k\}$ and $(i,j)\in I$ it holds that
 \[\Big(o^{(t)}(\bx)\Big)_{(i,j),k+s}=\Big(o^{(r\cdot3\cdot k)}(\bx)\Big)_{(i,j),s}\]
 for all $\bx\in[0,1]^{\{1,\dots,d_1\}\times\{1,\dots,d_2\}}$.
 
 Now, by using Lemma 2 for each $s\in\{1,\dots,k\}$ with parameters
 \[
 s_m=\IND_{\N\setminus\{1\}}(s)\cdot k+s
 \]
 for $m\in\{1,\dots,4\}$ and $s_5=s$,
 we can calculate the values \eqref{ple8eq3}  in layers 
 \[3\cdot r\cdot k+(s-1)\cdot3+1,\dots,3\cdot r\cdot k+s\cdot3\]
 by choosing corresponding weights 
 in channels 
 \[s,2\cdot k+1,\dots,2\cdot k+4\]
 such that we have
 \begin{align*}
 	&\Big(\big(o^{(3\cdot r\cdot k+3\cdot s)}\circ f\big)(\bx)\Big)_{(i,j),s}\\
 	&=\sigma\Bigg({g}_{max}\Big(
 	\big(o^{(r\cdot 3\cdot k)}(\bx)\big)_{(i,j),s},
 	\big(o^{(r\cdot 3\cdot k)}(\bx)\big)_{(i+2^r,j),s},
 	\\&\hspace{2cm}
 	\big(o^{(r\cdot 3\cdot k)}(\bx)\big)_{(i,j+2^r),s},
 	\big(o^{(r\cdot 3\cdot k)}(\bx)\big)_{(i+2^r,j+2^r),s}\Big)\Bigg)
 \end{align*}
 for all $\bx\in[0,1]^{\{1,\dots,d_1\}\times\{1,\dots,d_2\}}$, $s\in\{1,\dots,k\}$ and $(i,j)\in I$, where we set 
 \[\big(o^{(r\cdot 3\cdot k)}(\bx)\big)_{(i,j),s}=0\]
 for $(i,j)\notin I$.
 Once a value has been saved in layer $3\cdot r\cdot k+3\cdot s$, it will be propagated to the next layer using equation \eqref{ple1eq1} such that we have
 \begin{align*}
 	&\Big(\big(o^{(3\cdot(r+1)\cdot k)}\circ f\big)(\bx)\Big)_{(i,j),s}\\
 	&=\sigma\Bigg({g}_{max}\Big(
 	\big(o^{(r\cdot 3\cdot k)}(\bx)\big)_{(i,j),s},
 	\big(o^{(r\cdot 3\cdot k)}(\bx)\big)_{(i+2^r,j),s},
 	\\&\hspace{2.5cm}
 	\big(o^{(r\cdot 3\cdot k)}(\bx)\big)_{(i,j+2^r),s},
 	\big(o^{(r\cdot 3\cdot k)}(\bx)\big)_{(i+2^r,j+2^r),s}\Big)\Bigg)\\
 	&=\max_{(i,j)\in\{i,\dots,i+2^{r+1}-1\}\times\{j,\dots,j+2^{r+1}-1\}\cap I}\big(f(\bx)\big)_{(i,j),s}
 \end{align*}
 for all $(i,j)\in I$ and $s\in\{1,\dots,k\}$, where the last line follows from the induction hypothesis.
 Finally, the assertion follows from the definition of the subsampling layer $f_{sub}$ and the local max-pooling layer $f_{max}$.
 \hfill $\Box$
 
 In order to represent a convolutional neural network with several subsampling layers by a convolutional neural network with a single subsampling layer on top of all convolutional layers (as in $\F_3(\btheta_3)$) we will use the next lemma.

 \begin{lemma}
 	\label{le3}
 	Let $i_1,i_2,n,k,k'\in\N$
 	and set
 	\[
 	I=\{1,\dots,i_1\}\times\{1,\dots,i_2\}\mbox{ and }\tilde{I}=\{1,\dots,\lceil i_1/n\rceil\}\times\{1,\dots,\lceil i_2/n\rceil\}.
 	\]
 	Furthermore, let
 	\[f:\R^{\{1,\dots,d_1\}\times\{1,\dots,d_2\}}\rightarrow\R_+^{I\times\{1,\dots,k'\}}\]
 	be a function and let
 	\[
 	o^{(z)}_{(k',k),M}:\R^{\tilde{I}\times\{1,\dots,k'\}}\rightarrow\R^{\tilde{I}\times\{1,\dots,k\}}
 	\]
 	be a convolutional block
 	defined as in Subsection 3.2 with $z\in\N$ layers and a filter size of $M\in\N$ and fixed weight matrices $\bw_1,\dots,\bw_z$. Then there exist a convolutional block
 	\[
 	\tilde{o}^{(z)}_{(k',k),(M-1)\cdot n+1}:\R^{I\times\{1,\dots,k'\}}\rightarrow\R^{I\times\{1,\dots,k\}}
 	\]
 	such that
 	\[
 	f_{sub}^{(n)}\circ \tilde{o}^{(z)}_{(k',k),(M-1)\cdot n+1}\circ f
 	=
 	o^{(z)}_{(k',k),M}\circ f_{sub}^{(n)}\circ f.
 	\]
 \end{lemma}
 {\bf Proof.}
 The idea is to access in each convolutional layer of $\tilde{o}^{(z)}_{(k',k),(M-1)\cdot n+1}$ only the values of the positions of the previous layer that would remain after applying the subsampling layer $f_{sub}^{(n)}$ to the previous layer.
 We will achieve this by choosing the weight matrices $\tilde{\bw}_1,\dots,\tilde{\bw}_z$ of $\tilde{o}^{(z)}_{(k',k),(M-1)\cdot n+1}$ such that they have non-zero values only at the positions of the index set $J\subset\{1,\dots,(M-1)\cdot n+1\}^2$ defined by
 \[
 J=\{((t_1-1)\cdot n+1,(t_2-1)\cdot n+1) : t_1,t_2\in\{1,\dots,M\}\}.
 \]
 The values of the weights matrices $\tilde{\bw}_1,\dots,\tilde{\bw}_z$ that are non-zero then match the values of the weight matrices $\bw_1,\dots,\bw_z$ of the convolutional block $o^{(z)}_{(k',k),M}$.  
 In the proof we set
 \begin{equation*}
 	k'(t)=
 	\begin{cases}
 		k'&,\text{ if }t=1\\
 		k&,\text{ else}
 	\end{cases}
 \end{equation*}
 and note that the weight matrices $\bw_1,\dots,\bw_z$ of the convolutional block $o^{(z)}_{(k',k),M}$ are of the following form
 \[
 \bw_t= \left(
 \left(
 w^{(t)}_{i,j,s_1,s_2}
 \right)_{
 	1 \leq i,j \leq M, s_1 \in \{1, \dots, k'(t)\}, s_2 \in \{1, \dots, k\}
 }
 ,
 \left(
 w^{(t)}_{s_2}
 \right)_{
 	s_2 \in \{1, \dots, k\}
 }         
 \right)
 \]
 for $t\in\{1,\dots,z\}$. 
 For $(i,j)\in\{1,\dots,(M-1)\cdot n+1\}^2$, $t\in\{1,\dots,z\}$, $s_1\in\{1,\dots,k'(t)\}$, $s_2\in\{1,\dots,k\}$ we set
 \begin{equation}
 	\label{ple3eq2}
 	\tilde{w}^{(t)}_{i,j,s_1,s_2}=
 	\begin{cases}
 		w^{(t)}_{\frac{i-1}{n}+1,\frac{j-1}{n}+1,s_1,s_2}&,\text{ if }(i,j)\in J\\
 		0&,\text{ elsewhere}
 	\end{cases}
 \end{equation}
 and 
 \[
 \tilde{w}^{(t)}_{s_2}={w}^{(t)}_{s_2}.
 \]
 Since we want to show the assertion by an induction on the layers of the convolutional blocks we define $\tilde{o}^{(0)}=f$, ${o}^{(0)}=f_{sub}^{(n)}\circ f$ and for $t=1,\dots,z$ we define the functions 
 \[
 o^{(t)}=o_{(k'(t),k),M,\bw_t}\circ o^{(t-1)}\mbox{ and }\tilde{o}^{(t)}=\tilde{o}_{(k'(t),k),(M-1)\cdot n+1,\tilde{\bw}_t}\circ \tilde{o}^{(t-1)}.
 \]
 We then show by induction on $t$ that we have
 \[
 \big(\tilde{o}^{(t)}(\bx)\big)_{((i-1)\cdot n+1,(j-1)\cdot n+1),s}= \big({o}^{(t)}(\bx)\big)_{(i,j),s}
 \]
 for all $(i,j)\in\tilde{I}$, $\bx\in\R^{\{1,\dots,d_1\}\times\{1,\dots,d_2\}}$ and $t\in\{0,\dots,z\}$. For $t=0$ the assertion follows directly from the definition of the subsampling layer $f_{sub}^{(n)}$. Now assume the assertion is true for some $t\in\{0,\dots,z-1\}$. For $(i,j)\in\tilde{I}$, $s\in\{1,\dots,k\}$ and $\bx\in\R^{\{1,\dots,d_1\}\times\{1,\dots,d_2\}}$ we get by the induction hypothesis and the definition of the weights \eqref{ple3eq2}
 \begin{align*}
 	&\big(\tilde{o}^{(t+1)}(\bx)\big)_{((i-1)\cdot n+1,(j-1)\cdot n+1),s}\\
 	&=\left(\big(\tilde{o}_{(k'(t+1),k),(M-1)\cdot n+1,\bw_{t+1}}\circ\tilde{o}^{(t)}\big)(\bx)\right)_{((i-1)\cdot n+1,(j-1)\cdot n+1),s}\\
 	&=
 	\sigma \left(
 	\sum_{s_1=1}^{k'(t+1)}
 	\sum_{
 		\substack{
 			t_1,t_2 \in J
 			\\((i-1)\cdot n+t_1,(j-1)\cdot n+t_2)\in I}
 	}
 	\tilde{w}^{(t+1)}_{t_1,t_2,s_1,s_2}
 	\cdot
 	\big(\tilde{o}^{(t)}(\bx)\big)_{((i-1)\cdot n+t_1,(j-1)\cdot n+t_2),s_1}
 	+
 	\tilde{w}_{s_2}^{(t+1)}
 	\right)\\
 	&=
 	\sigma \Bigg(
 	\sum_{s_1=1}^{k'(t+1)}
 	\sum_{
 		\substack{
 			t_1,t_2 \in\{1,\dots,M\}
 			\\((i+t_1-2)\cdot n+1,(j+t_2-2)\cdot n+1)\in I}
 	}\\
 	&\hspace{5cm}
 	{w}^{(t+1)}_{t_1,t_2,s_1,s_2}
 	\cdot
 	\big(\tilde{o}^{(t)}(\bx)\big)_{((i+t_1-2)\cdot n+1,(j+t_2-2)\cdot n+1),s_1}
 	+
 	w_{s_2}^{(t+1)}
 	\Bigg)\\
 	&\stackrel{}{=}
 	\sigma \left(
 	\sum_{s_1=1}^{k'(t+1)}
 	\sum_{
 		\substack{
 			t_1,t_2 \in\{1,\dots,M\}
 			\\(i+t_1-1,i+t_2-1)\in\tilde{I}}
 	}
 	{w}^{(t+1)}_{t_1,t_2,s_1,s_2}
 	\cdot
 	\big({o}^{(t)}(\bx)\big)_{(i+t_1-1,i+t_2-1),s_1}
 	+
 	w_{s_2}^{(t+1)}
 	\right)\\
 	&=\big({o}^{(t+1)}(\bx)\big)_{(i,j),s},
 \end{align*}
 where the fourth equality follows from the induction hypothesis together with the fact that
 \begin{equation*}
 	\begin{split}
 		&((i+t_1-2)\cdot n+1,(j+t_2-2)\cdot n+1)\in I
 		\iff
 		(i+t_1-1,i+t_2-1)\in\tilde{I}
 	\end{split}
 \end{equation*}
 implied by Lemma 4 a) below.
 The assertion then follows because of the definition of the subsampling layer $f_{sub}^{(n)}.$
 \hfill $\Box$
 
 \begin{lemma}
 	\label{le4}
 	Let $a,b,c\in\N$.\\
 	{\bf a)} It holds that
 	\[
 	(a-1)\cdot b+1\leq c
 	\iff
 	a\leq\left\lceil\frac{c}{b}\right\rceil.
 	\]
 	{\bf b)} It holds that
 	\[
 	\left\lceil\frac{c}{a\cdot b}\right\rceil=\left\lceil\frac{\lceil c/a\rceil}{b}\right\rceil.
 	\]
 \end{lemma}
 {\bf Proof.} {\bf a)} We have
 \begin{equation*}
 	(a-1)\cdot b+1\leq c
 	\iff
 	a\leq\frac{c+(b-1)}{b}\stackrel{a\in\N}{\iff} a\leq\left\lfloor\frac{c+(b-1)}{b}\right\rfloor
 \end{equation*}
 and
 \begin{align*}
 	\left\lfloor\frac{c+(b-1)}{b}\right\rfloor&=\max\left\{k\in\Z : k\leq\frac{c+(b-1)}{b}\right\}\\
 	&=\min\left\{k\in\Z : k+1>\frac{c+(b-1)}{b}\right\}\\
 	&=\min\left\{k\in\Z : k+1\geq\frac{c+(b-1)}{b}+\frac{1}{b}\right\}\\
 	&=\left\lceil\frac{c}{b}\right\rceil.
 \end{align*}
 {\bf b)}
 We set $x\coloneqq\lceil c/a\rceil-c/a$ and calculate
 \begin{align*}
 	\left\lceil\frac{c}{a\cdot b}\right\rceil
 	&~~=\min\left\{k\in\Z : k\geq\frac{c}{a\cdot b}\right\}\\
 	&~~=\min\left\{k\in\Z : x\geq\lceil c/a\rceil-k\cdot b\right\}\\
 	&\stackrel{x\in[0,1)}{=}\min\left\{k\in\Z : 0\geq\lceil c/a\rceil-k\cdot b\right\}\\
 	&~~=\left\lceil\frac{\lceil c/a\rceil}{b}\right\rceil.
 \end{align*}
 \hfill $\Box$ 
 
 In the following, we show a connection between the three classes of convolutional neural networks from Section 3.6.
 \begin{lemma}
 	\label{le5}
 	Let $d_1,d_2,L,z\in\N$, $\bk=(k_1,\dots,k_L)\in\N^L$ and $s_0,\dots,s_{L-1}\in\{2^0,\dots,2^{L-1}\}$ with
 	\begin{equation}
 		\label{le4eq1}
 		\prod_{i=0}^{k}{s_i}\leq 2^{k}
 	\end{equation}
 	and denote $\bs=(s_1,\dots,s_{L-1})$ and $\bar{\bk}=(2\cdot k_1+4,\dots,2\cdot k_L+4)$.
 	For $r\in\{1,\dots,L\}$ set
 	\[\quad M_{r}=\frac{2^{r-1}}{\prod_{i=0}^{r-1}s_i}+1,\quad \bar{M}_r=2^{r-1}+1\quad\text{and}\quad s=\prod_{i=1}^{L-1}s_i.\]
 	Furthermore, set $k_{max}=\max\{k_1,\dots,k_L\}$, $s_{max}=\max\{s_1\dots,s_{L-1}\}$ and
 	\[\bar{z}=3\cdot k_{max}\cdot\log_2(s_{max}).\]
 	Let $\F_{1}(\btheta_1)$, $\F_2(\btheta_2)$ and $\F_3(\btheta_3)$ be the function classes from Subsection 3.5 with parameters
 	$\btheta_1=(L,\bk,\bM,z,\bs,\tilde{\bd})$, $\btheta_2=(L,\bar{\bk},\bM,\bar{z},\bs,\tilde{\bd})$ and $\btheta_3=(L,\bar{\bk},\bar{\bM},\bar{z},s,\tilde{\bd})$ with arbitrary output bounds $\tilde{\bd}$.
 	Then we have
 	\[\F_1(\btheta_1)\subset\F_2(\btheta_2)\subset\F_3(\btheta_3).\]
 \end{lemma} 
 {\bf Proof.}
 First we show that $\F_1(\btheta_1)\subset\F_2(\btheta_2)$. Let $f\in\F_1(\btheta_1)$ defined by
 \[
 f=f_{out}^{(\tilde{d}_1,\tilde{d}_2)}\circ o^{(z)}_{(k_{L-1},k_{L}),M_L}\circ f_{max}^{(s_{L-1})}\circ o^{(z)}_{(k_{L-2},k_{L-1}),M_{L-1}}\circ\dots\circ f_{max}^{(s_1)}\circ o^{(z)}_{(1,k_{1}),M_1}.
 \]
 with weight vector $\bw=\big((\bw_{r,1},\dots,\bw_{r,z})_{r\in\{1,\dots,L\}},\bw_{out}\big)$ (as defined in Section 3.5).
 The idea is to modify the convolutional neural network $f$ such that we can use Lemma \ref{le1} at the end of each convolutional block to replace each local max-pooling layer by a convolutional block followed by a local subsampling layer.
 To use Lemma \ref{le1}, we add $k_r+4$ channels in each convolutional block $r\in\{1,\dots,L\}$ to each convolutional layer. 
 Already existing weights remain the same and the weights added in the $t$-th convolutional layer in the $r$-th convolutional block are chosen such that we have
 \[
 w_{i,j,s_1,s_2}^{(t)}=0\quad\text{if}\quad s_1>k_{r-1}\quad\text{and}\quad s_2\leq k_r
 \]
 for the corresponding weight vector %
 (defined as in Section 3.1) for all $r\in\{1,\dots,L\}$ and $t\in\{1,\dots,z\}$ (with $k_0=1$).
Then we get convolutional blocks 
\[{o}_{(1,2\cdot k_{1}+4),M_1}^{(z)},\dots,{o}_{(2\cdot k_{L-1}+4,2\cdot k_{L}+4),M_L}^{(z)}\]
with the following property:
 \begin{equation}
 	\label{ple5eq1}
 	\begin{split}
 		&\big({o}^{({z})}_{(2\cdot k_{r-1}+4,2\cdot k_{r}+4),M_{r}}\circ\dots\circ f_{max}^{(s_{2})}\circ{o}^{({z})}_{(2\cdot k_{1}+4,2\cdot k_{2}+4),M_2}\circ f_{max}^{(s_1)}\circ{o}^{({z})}_{(1,2\cdot k_{1}+4),M_1}(\bx)\big)_{(i,j),s}\\
 		&=\big({o}^{({z})}_{(k_{r-1},k_{r}),M_{r}}\circ\dots\circ f_{max}^{(s_2)}\circ{o}^{({z})}_{(k_{1},k_{2}),M_2}\circ f_{max}^{(s_1)}\circ{o}^{({z})}_{(1,k_{1}),M_1}(\bx)\big)_{(i,j),s}
 	\end{split} 
 \end{equation}
 for all $s\in\{1,\dots,k_r\}$, $r\in\{1,\dots,L\}$, $\bx\in[0,1]^{\{1,\dots,d_1\}\times\{1,\dots,d_2\}}$ and $(i,j)\in I_r$, where $I_r\subset\N^2$ denotes the corresponding index set defined by the codomain of the convolutional blocks of the network $f$.
 Because of inequality \eqref{le4eq1} we have 
 \[M_r=\frac{2^{r}}{\prod_{i=0}^{r}s_i}\cdot\frac{s_r}{2}+1\geq\frac{s_r}{2}+1\]
 for all $r\in\{1,\dots,L-1\}$.
 Therefore we can apply Lemma \ref{le1}, which yields convolutional blocks
 \[
 \tilde{o}^{(3\cdot k_1\cdot\log_2(s_1))}_{(2\cdot k_{1}+4,2\cdot k_{1}+4),M_{1}},\dots,\tilde{o}^{(3\cdot k_{L-1}\cdot\log_2(s_{L-1}))}_{(2\cdot k_{L-1}+4,2\cdot k_{L-1}+4),M_{L-1}},
 \]
 so that together with equation \eqref{ple5eq1} we get
 \begin{equation*}
 	\begin{split}
 		&\big({o}^{({z})}_{(2\cdot k_{L-1}+4,2\cdot k_{L}+4),M_{L}}\circ f_{sub}^{(s_{L-1})}\circ\tilde{o}^{(3\cdot k_{L-1}\cdot\log_2(s_{L-1}))}_{(2\cdot k_{L-1}+4,2\cdot k_{L-1}+4),M_{L-1}}\circ{o}^{({z})}_{(2\cdot k_{L-2}+4,2\cdot k_{L-1}+4),M_{L-1}}\circ\dots\\
 		&\hspace{6cm}
 		\circ f_{sub}^{(s_1)}\circ\tilde{o}^{(3\cdot k_1\cdot\log_2(s_1))}_{(2\cdot k_{1}+4,2\cdot k_{1}+4),M_1}\circ{o}^{({z})}_{(1,2\cdot k_{1}+4),M_1}(\bx)\big)_{(i,j),s}\\
 		&=\big(o^{(z)}_{(k_{L-1},k_{L}),M_{L}}\circ f_{max}^{(s_{L-1})}\circ o^{(z)}_{(k_{L-2},k_{L-1}),M_{L-1}}\circ\dots\circ f_{max}^{(s_1)}\circ o^{(z)}_{(1,k_{1}),M_1}(\bx)\big)_{(i,j),s}
 	\end{split}
 \end{equation*}
 for all $s\in\{1,\dots,k_L\}$, $\bx\in[0,1]^{\{1,\dots,d_1\}\times\{1,\dots,d_2\}}$ and $(i,j)\in I_L$. By the idea of propagating calculated values to the next layer (cf., equation \eqref{ple1eq1}) we can add appropriate many convolutional layers to each convolutional block such that there are convolutional blocks
 \[
 \bar{o}^{(\bar{z})}_{(1,\bar{k}_1),M_{1}},\dots,\bar{o}^{(\bar{z})}_{(\bar{k}_{L-1},\bar{k}_{L}),M_{L}}
 \]
 satisfying
 \begin{equation}
 	\label{ple4eq3}
 	\begin{split}
 		&\big(\bar{o}^{(\bar{z})}_{(\bar{k}_{L-1},\bar{k}_{L}),M_{L}}\circ f_{sub}^{(s_{L-1})}\circ\bar{o}^{(\bar{z})}_{(\bar{k}_{L-2},\bar{k}_{L-1}),M_{L-1}}\circ\dots
 		\circ f_{sub}^{(s_1)}\circ\bar{o}^{(\bar{z})}_{(1,\bar{k}_{1}),M_1}\big)_{(i,j),s}\\
 		&=\big(o^{(z)}_{(k_{L-1},k_{L}),M_{L}}\circ f_{max}^{(s_{L-1})}\circ o^{(z)}_{(k_{L-2},k_{L-1}),M_{L-1}}\circ\dots\circ f_{max}^{(s_1)}\circ o^{(z)}_{(1,k_{1}),M_1}(\bx)\big)_{(i,j),s}
 	\end{split}
 \end{equation}
 for all $s\in\{1,\dots,k_L\}$, $(i,j)\in I_L$ and $\bx\in[0,1]^{\{1,\dots,d_1\}\times\{1,\dots,d_2\}}$.
 Finally, we define the output layer
 $
 \bar{f}_{out}^{(\tilde{d}_1,\tilde{d}_2)}:\R^{I_L\times\{1,\dots,2\cdot k_L+4\}}\rightarrow\R
 $
 by defining the output weights 
 \[\tilde{\bw}_{out}=(\tilde{w}_s)_{s\in\{1,\dots,2\cdot k_L+4\}}\]
 of $\tilde{f}_{out}^{(\tilde{d}_1,\tilde{d}_2)}$ using the output weights $\bw_{out}=(w_s)_{s\in\{1,\dots,k_L\}}$ of $f_{out}^{(\tilde{d}_1,\tilde{d}_2)}$ as follows:
 \[
 \tilde{w}_s=
 \begin{cases}
 	w_s&\text{if }s\in\{1,\dots,k_L\},\\
 	0&\text{else}.
 \end{cases}
 \]
 We conclude by using equation \eqref{ple4eq3}
 \[
 f=\tilde{f}_{out}^{(\tilde{d}_1,\tilde{d}_2)}\circ\bar{o}^{(\bar{z})}_{(\bar{k}_{L-1},\bar{k}_{L}),M_{L}}\circ f_{sub}^{(s_{L-1})}\circ\bar{o}^{(\bar{z})}_{(\bar{k}_{L-2},\bar{k}_{L-1}),M_{L-1}}\circ\dots
 \circ f_{sub}^{(s_1)}\circ\bar{o}^{(\bar{z})}_{(1,\bar{k}_{1}),M_1}\in\F_2(\btheta_2).
 \]
 Now it remains to show that $\F_2(\btheta_2)\subset\F_3(\btheta_3)$.  
 Therefore let $f\in\F_2(\btheta_2)$ defined by
 \[
 f={f}_{out}^{(\tilde{d}_1,\tilde{d}_2)}\circ{o}^{(\bar{z})}_{(\bar{k}_{L-1},\bar{k}_{L}),M_{L}}\circ f_{sub}^{(s_{L-1})}\circ{o}^{(\bar{z})}_{(\bar{k}_{L-2},\bar{k}_{L-1}),M_{L-1}}\circ\dots
 \circ f_{sub}^{(s_1)}\circ{o}^{(\bar{z})}_{(1,\bar{k}_{1}),M_1}.
 \]
 Here the idea is, as long as an expression of the form $o^{(\bar{z})}_{(k',k),M}\circ f_{sub}^{(s)}$ exists for some $s,M\in\N$ in the representation of $f$, to replace it with an expression of the form
 \[
 f_{sub}^{(s)}\circ\tilde{o}^{(\bar{z})}_{(k',k),(M-1)\cdot s+1}
 \]
 by using Lemma \ref{le3}.
 If we start with the subsampling layer $f_{sub}^{(s_{L-1})}$ and move each subsampling layer to the end of all convolutional blocks in descending order $L-1,L-2,\dots,1$, we need to apply Lemma \ref{le3}
 \[
 \underbrace{1}_{\text{for }f_{sub}^{(s_{L-1})}}+\underbrace{2}_{\text{for }f_{sub}^{(s_{L-2})}}+\dots+\underbrace{L-1}_{\text{for }f_{sub}^{(s_{1})}}=\frac{L\cdot(L-1)}{2}
 \]
 times to get convolutional blocks $\tilde{o}^{(z)}_{(\bar{k}_1,\bar{k}_2),{M}_2^{(1)}},\dots,\tilde{o}^{(z)}_{(\bar{k}_{L-1},\bar{k}_{L}),{M}_L^{(L-1)}}$ such that
 \begin{align*}
 	f&=f_{out}^{(\tilde{d}_1,\tilde{d}_2)}\circ o^{(z)}_{(\bar{k}_{L-1},\bar{k}_{L}),M_L}\circ f_{sub}^{(s_{L-1})}\circ o^{(z)}_{(\bar{k}_{L-2},\bar{k}_{L-1}),M_{L-1}}\circ\dots\circ f_{sub}^{(s_1)}\circ o^{(z)}_{(1,\bar{k}_{1}),M_1}\\
 	&=f_{out}^{(\tilde{d}_1,\tilde{d}_2)}f_{sub}^{(s_{L-1})}\circ\dots\circ f_{sub}^{(s_1)}\circ \tilde{o}^{(z)}_{(\bar{k}_{L-1},\bar{k}_{L}),{M}_L^{(L-1)}}\circ\dots\circ\tilde{o}^{(z)}_{(\bar{k}_{1},\bar{k}_{2}),{M}_2^{(1)}}\circ{o}^{(z)}_{(1,\bar{k}_{1}),{M_1}},
 \end{align*}
 where the filter size $M_r^{(r-1)}$ is recursively given by
 \[
 M_r^{(t)}=\left(M_r^{(t-1)}-1\right)\cdot s_{r-t}+1
 \]
 for $t\in\{1,\dots,r-1\}$ and $M_r^{(0)}=M_r$ for $r\in\{2,\dots,L\}$. By an induction on $t$ it is easy to see that
 \[
 M_r^{(t)}=\frac{2^{r-1}}{\prod_{i=0}^{r-1-t}s_i}+1
 \]
 for $t\in\{0,\dots,r-1\}$, which implies that $M_r^{(r-1)}=2^{r-1}+1=\bar{M}_r$ for $r\in\{2,\dots,L\}$.
 It now remains to show 
 \[
 f_{sub}^{(s_1\cdot s_2\cdot\ldots\cdot s_{L-1})}=f_{sub}^{(s_1)}\circ f_{sub}^{(s_2)}\circ\ldots\circ f_{sub}^{(s_{L-1})},
 \]
 which follows from Lemma 4 b).
 \hfill $\Box$
 
 \section{An approximation result for convolutional neural networks}
 In this section, we prove the approximation result for the hierarchical max-pooling model with feature constraint and local max-pooling parameter by the convolutional neural networks with several local max-pooling layers (as in $\F_1(\btheta_1)$) which we used in the proof of Theorem \ref{th1}.
 In the following $d_1,d_2\in\N\setminus\{1\}$ denote the image dimensions and we assume that the function 
 \begin{equation}
 	\label{fm}
 	m:[0,1]^{\{1,\dots,d_1\}\times\{1,\dots,d_2\}}\rightarrow[0,1]
 \end{equation}
 satisfies a hierarchical max-pooling model of level $l\in\N$ with feature constraint $\bb=(b_1,\dots,b_{l-1})$ and local max-pooling parameter $\bn=(n_1,\dots,n_{l-1})$ (see Definition 3 from Section 2)
 where the feature maps are described by
 \[z_{k,s}:[0,1]^{\{1,\dots,d_1\}\times\{1,\dots,d_2\}}\rightarrow[0,1]^{\{1,\dots,\lceil d_1(k)/n_k\rceil\}\times\{1,\dots,\lceil d_2(k)/n_k\rceil\}}\]
 for $k=0,\dots,l$ and $s=1,\dots,b_k$ and
 \[
 {y}_{k,s}:[0,1]^{\{1,\dots,d_1\}\times\{1,\dots,d_2\}}\rightarrow[0,1]^{\{1,\dots,d_1(k)\}\times\{1,\dots,d_2(k)\}}
 \]
 and
 \begin{equation}
 	\label{pgs}
 	g_{k,s}:\R^4\rightarrow[0,1]
 \end{equation}
 for $k=1,\dots,l$ and $s=1,\dots,b_k$.
 The above dimensions $d_1(k)$ and $d_2(k)$ are defined as in Definition 3 by
 \[
 d_1(k)=	\left\lceil
 \frac{d_1(k-1)}{n_{k-1}}
 \right\rceil
 -\delta_{k-1}
 \mbox{ and }
 d_2(k)=	\left\lceil
 \frac{d_1(k-1)}{n_{k-1}}
 \right\rceil
 -\delta_{k-1},
 \]
 for $k=1,\dots,l$ with $d_1(0)=d_1$, $d_2(0)=d_1$, and $\delta_k={2^{k}}/{\prod_{i=0}^{k}n_i}$.
 
 Our aim is to show that a convolutional neural network from class $\F_1(\btheta_1)$ can mimic a function
 \begin{equation*}
 	\bar{m}:[0,1]^{\{1,\dots,d_1\}\times\{1,\dots,d_2\}}\rightarrow[0,1]
 \end{equation*}
 of the following form, which then approximates the function $m$. 
 
 Define $\bar{m}$ by
 \begin{equation}
 	\label{defmq}
 	\bar{m}(x)=
 	\max_{
 		(i,j)\in\{1,\dots,d_1(l)\}\times\{1,\dots,d_2(l)\}}
 	\big(\bar{z}(\bx)\big)_{(i,j)},
 \end{equation}
 where $\bar{z}$ satisfy
 \[
 \bar{z}=\bar{z}_{l,1}
 \]
 for some
 \[
 \bar{z}_{k,s}:[0,1]^{\{1,\dots,d_1\}\times\{1,\dots,d_2\}}\rightarrow\R^{\{1,\dots,\lceil d_1(k)/n_k\rceil\}\times\{1,\dots,\lceil d_2(k)/n_k\rceil\}}
 \]
 recursively defined by:
 \begin{enumerate}
 	\item 
 	 Let $\bar{g}_{k,s}:\R^4\rightarrow\R$   
 	be functions for $k\in\{1, \dots, l\}$ and $s\in\{1, \dots,b_k\}$ and
set
 	\[
 	\bar{z}_{0,1}(\bx)=\bx.
 	\]
 	\item 	
 	We define recursively functions
 	\[
 	\bar{y}_{k,s}:[0,1]^{\{1,\dots,d_1\}\times\{1,\dots,d_2\}}\rightarrow\R^{\{1,\dots,d_1(k)\}\times\{1,\dots,d_2(k)\}}
 	\]
 	for $k=1,\dots,l$ by
 	\begin{equation}
 		\label{appry}
 		\begin{split}
 			&\big(\bar{y}_{k,s}(\bx)\big)_{(i,j)}=\sigma\Big(\bar{g}_{k,s}\Big(
 			\big(\bar{z}_{k-1,r_1(k,s)}(\bx)\big)_{(i,j)},
 			\big(\bar{z}_{k-1,r_2(k,s)}(\bx)\big)_{(i+\delta_{k-1},j)},
 			\\&\hspace{3.5cm}
 			\big(\bar{z}_{k-1,r_3(k,s)}(\bx)\big)_{(i,j+\delta_{k-1})},\big(\bar{z}_{k-1,r_4(k,s)}(\bx)\big)_{(i+\delta_{k-1},j+\delta_{k-1})}\Big)\Big)
 		\end{split}
 	\end{equation}
 	for $k=1,\dots,l$, $s=1,\dots,b_k$, $(i,j)\in\{1,\dots,d_1(k)\}\times\{1,\dots,d_2(k)\}$, 
 	\[r_1(k,s), r_2(k,s), r_3(k,s), r_4(k,s)\in\{1,\dots,b_{k-1}\}.\]
 	\item
 	Next we define
 	\begin{equation}
 		\label{apprz}
 		\big(\bar{z}_{k,s}(\bx)\big)_{(i,j)}=
 		\max_{(i_2,j_2)\in N^{(k)}_{(i,j)}}
 		\big(\bar{y}_{k,s}(\bx)\big)_{(i_2,j_2)}.
 	\end{equation}
 	for $k=1,\dots,l$, $s=1,\dots,b_k$ and $(i,j)\in\{1,\dots,\lceil d_1(k)/n_k\rceil\}\times\{1,\dots,\lceil d_2(k)/n_k\rceil\}$,
 \end{enumerate}    
 where the neighborhoods 
 $N_{(i,j)}^{(k)}$ are defined by equation \eqref{NeighbH}. In equation \eqref{appry}, the ReLU activation function
 $
 \sigma(x)= \max\{x,0\}
 $
 is applied on the right-hand side, as this will allow us to represent the approximation $\bar{m}$ of $m$ by a convolutional neural network, in the case where the functions $\bar{g}_{k,s}$ are standard feedforward neural networks (cf., Lemma \ref{le7}). On the other hand, applying the ReLU activation function does not affect our approximation result, since the functions $g_{k,s}$ in the definition of $m$ are $[0,1]$-valued.
 We start with the following result, which provides a slight extension of Lemma 4 in \cite{KoKrWa2020}.
 \begin{lemma}
 	\label{le6}
 	Assume that for all $k\in\{1,\dots,l\}$ and $s\in\{1,\dots,b_k\}$ the restriction
 	$g_{k,s}|_{[-2,2]^4}: [-2,2]^4 \rightarrow [0,1]$ of the function \eqref{pgs} is Lipschitz continuous (with respect to the Euclidean distance)
 	with Lipschitz constant $C>0$.
 	Furthermore, assume that for all $k\in\{1,\dots,l\}$ and $s\in\{1,\dots,b_k\}$
 	\begin{equation}
 		\left\|\bar{g}_{k,s}\right\|_{[-2,2]^4,\infty}\leq2.
 		\label{le4eq3}
 	\end{equation} Then for any
 	$\bx \in [0,1]^{\{1, \dots, d_1\} \times \{1, \dots, d_2\}}$ it holds:
 	\[
 	|m(\bx)-\bar{m}(\bx)| \leq
 	(2C+1)^{l-1}
 	\cdot \max_{k \in \{1, \dots,l\}, s \in \{ 1, \dots, b_k\}}
 	\|                      g_{k,s}
 	-
 	\bar{g}_{k,s}
 	\|_{[-2,2]^4,\infty}
 	.
 	\]
 \end{lemma}
  \noindent 
 {\bf Proof.}
 If
 $a_1$, $b_1$, \dots, $a_n$, $b_n \in \R$, then
 \begin{equation}
 	\label{ple4eq1}
 	| \max_{i=1,\dots,n} a_i - \max_{i=1, \dots, n} b_i| \leq
 	\max_{i=1, \dots, n} |a_i-b_i|.
 \end{equation}
 Indeed, in case $ a_1=\max_{i=1,\dots,n} a_i \geq  \max_{i=1, \dots, n} b_i$
 (which we can assume w.l.o.g.) we have
 \begin{eqnarray*}
 	&&
 	| \max_{i=1,\dots,n} a_i - \max_{i=1, \dots, n} b_i|
 	=
 	a_1 - \max_{i=1, \dots, n} b_i
 	\leq
 	a_1-b_1 \leq
 	\max_{i=1, \dots, n} |a_i-b_i|.
 \end{eqnarray*}
 Consequently it suffices to show
 \begin{equation*}
 	\begin{split}
 		&
 		\max_{
 			(i,j)\in\{1, \dots, d_1\} \times \{1, \dots, d_2\}
 		}
 		\left|
 		\big(z(\bx)\big)_{(i,j)}
 		-
 		\big(\bar{z}(\bx)\big)_{(i,j)}
 		\right|
 		\\
 		&
 		\leq
 		(2C+1)^{l-1}
 		\cdot \max_{k \in \{1, \dots,l\}, s \in \{ 1, \dots,b_k\}}
 		\|                      g_{k,s}
 		-
 		\bar{g}_{k,s}
 		\|_{[-2,2]^4,\infty}.
 	\end{split}
 \end{equation*}
 This in turn follows from
 \begin{equation}
 	\label{ple4eq2}
 	|\big(z_{k,s}(\bx)\big)_{(i,j)}-\big(\bar{z}_{k,s}(\bx)\big)_{(i,j)}| \leq 
 	(2C+1)^{k-1}\cdot 
 	\max_{i \in \{1, \dots,k\}, s \in \{1, \dots, b_k\}}
 	\|                      g_{i,s}
 	-
 	\bar{g}_{i,s}
 	\|_{[-2,2]^4,\infty}
 \end{equation}
 for all $k \in \{1, \dots, l\}$, all $s \in \{1, \dots, b_k\}$, all $(i,j)\in\{1,\dots,d_1(k)\}\times\{1,\dots,d_2(k)\}$
 and all $\bx \in [0,1]^{\{1, \dots, d_1\} \times \{1, \dots, d_2\}}$,
 which we show in the sequel by induction on $k$.
 
 Because of equation \eqref{ple4eq1} and since $g_{1,s}$ is $[0,1]$-valued for all $s\in\{1,\dots,b_1\}$ we get for $k=1$, $s \in \{1, \dots, b_1\}$ and $(i,j)\in\{1,\dots,\lceil d_1(1)/n_1\rceil\}\times\{1,\dots,\lceil d_2(1)/n_1\rceil\}$ that
 \begin{eqnarray*}
 	&&\left|
 	\big(z_{1,s}(\bx)\big)_{(i,j)}
 	-
 	\big(\bar{z}_{1,r}(\bx)\big)_{(i,j)}
 	\right|\\
 	&  = &
 	\Big|
 	\max_{(i_2,j_2)\in N^{(1)}_{(i,j)}}g_{1,s}(x_{i,j},x_{i+1,j},x_{i,j+1},x_{i+1,j+1})\\
 &~&\hspace{1cm}	-
 	\max_{(i_2,j_2)\in N^{(1)}_{(i,j)}}
 	\sigma\big(\bar{g}_{1,s}(x_{i,j},x_{i+1,j},x_{i,j+1},x_{i+1,j+1})\big)
 	\Big|\\
 	&  \leq &
 	\max_{(i_2,j_2)\in N^{(1)}_{(i,j)}}
 	\left|
 	g_{1,s}(x_{i,j},x_{i+1,j},x_{i,j+1},x_{i+1,j+1})
 	-
 	\bar{g}_{1,s}(x_{i,j},x_{i+1,j},x_{i,j+1},x_{i+1,j+1})
 	\right|\\
 	&
 	\leq&
 	\left\| g_{1,s} - \bar{g}_{1,s}\right\|_{[0,1]^4,\infty} .
 \end{eqnarray*}
 Assume now that (\ref{ple4eq2}) holds for some $k \in \{1, \dots, l-1\}$. The definition of $\bar{z}_{k,s}$ and inequality \eqref{le4eq3} imply that
 \[\left|\big(\bar{z}_{k,s}(\bx)\big)_{(i,j)}\right|\leq2\]
 for all $\bx\in[0,1]^{\{1,\dots,d_1\}\times\{1,\dots,d_2\}}$, $s\in\{1,\dots,b_k\}$ and $(i,j)\in\{1,\dots,d_1(k)\}\times\{1,\dots,d_2(k)\}$.
 Then, because of the triangle inequality, the Lipschitz assumption on $g$, inequality \eqref{ple4eq1} and inequality \eqref{ple4eq2} and since $g_{k+1,s}$ is $[0,1]$-valued we get
 \begin{align*}
 	&~\left|\big(z_{k+1,s}(\bx)\big)_{(i,j)}-\big(\bar{z}_{k+1,s}(\bx)\big)_{(i,j)}\right|\\
 	&~=\left|\max_{(i_2,j_2)\in N^{(k+1)}_{(i,j)}}
 	\big(y_{k+1,s}(\bx)\big)_{i_2,j_2}
 	-
 	\max_{(i_2,j_2)\in N^{(k+1)}_{(i,j)}}
 	\big(\bar{y}_{k+1,s}(\bx)\big)_{i_2,j_2}
 	\right|\\
 	&\stackrel{\eqref{ple4eq1}}{\leq}\max_{(i_2,j_2)\in N^{(k+1)}_{(i,j)}}
 	\left|
 	\big(y_{k+1,s}(\bx)\big)_{i_2,j_2}
 	-
 	\big(\bar{y}_{k+1,s}(\bx)\big)_{i_2,j_2}
 	\right|\\
 	&~~{\leq}\max_{(i_2,j_2)\in N^{(k+1)}_{(i,j)}}
 	\Big|
 	g_{k+1,s}\Big(
 	\big(z_{k,r_1(k+1,s)}(\bx)\big)_{i_2,j_2},
 	\big(z_{k,r_2(k+1,s)}(\bx)\big)_{i_2+\delta_k,j_2},
 	\\&\hspace{3cm}
 	\big(z_{k,r_3(k+1,s)}(\bx)\big)_{i_2,j_2+\delta_k},\big(z_{k,r_4(k+1,s)}(\bx)\big)_{i_2+\delta_k,j_2+\delta_k}\Big)
 	\\
 	&
 	\hspace{2.5cm}
 	-
 	\bar{g}_{k+1,s}\Big(
 	\big(\bar{z}_{k,r_1(k+1,s)}(\bx)\big)_{i_2,j_2},
 	\big(\bar{z}_{k,r_2(k+1,s)}(\bx)\big)_{i_2+\delta_k,j_2},
 	\\&\hspace{4cm}
 	\big(\bar{z}_{k,r_3(k+1,s)}(\bx)\big)_{i_2,j_2+\delta_k},\big(\bar{z}_{k,r_4(k+1,s)}(\bx)\big)_{i_2+\delta_k,j_2+\delta_k}\Big) \Big|
 	\\
 	&~~\leq\max_{(i_2,j_2)\in N^{(k+1)}_{(i,j)}}
 	\Big|
 	g_{k+1,s}\Big(
 	\big(z_{k,r_1(k+1,s)}(\bx)\big)_{i_2,j_2},
 	\big(z_{k,r_2(k+1,s)}(\bx)\big)_{i_2+\delta_k,j_2},
 	\\&\hspace{3cm}
 	\big(z_{k,r_3(k+1,s)}(\bx)\big)_{i_2,j_2+\delta_k},\big(z_{k,r_4(k+1,s)}(\bx)\big)_{i_2+\delta_k,j_2+\delta_k}\Big)
 	\\
 	&
 	\hspace{2.5cm}
 	-
 	g_{k+1,s}\Big(
 	\big(\bar{z}_{k,r_1(k+1,s)}(\bx)\big)_{i_2,j_2},
 	\big(\bar{z}_{k,r_2(k+1,s)}(\bx)\big)_{i_2+\delta_k,j_2},
 	\\&\hspace{4cm}
 	\big(\bar{z}_{k,r_3(k+1,s)}(\bx)\big)_{i_2,j_2+\delta_k},\big(\bar{z}_{k,r_4(k+1,s)}(\bx)\big)_{i_2+\delta_k,j_2+\delta_k}\Big)\Big|\\
 	&\hspace{2cm}
 	+ \Big|
 	g_{k+1,s}\Big(
 	\big(\bar{z}_{k,r_1(k+1,s)}(\bx)\big)_{i_2,j_2},
 	\big(\bar{z}_{k,r_2(k+1,s)}(\bx)\big)_{i_2+\delta_k,j_2},
 	\\&\hspace{3cm}
 	\big(\bar{z}_{k,r_3(k+1,s)}(\bx)\big)_{i_2,j_2+\delta_k},\big(\bar{z}_{k,r_4(k+1,s)}(\bx)\big)_{i_2+\delta_k,j_2+\delta_k}\Big)
 	\\
 	&
 	\hspace{2.5cm}
 	-
 	\bar{g}_{k+1,s}\Big(
 	\big(\bar{z}_{k,r_1(k+1,s)}(\bx)\big)_{i_2,j_2},
 	\big(\bar{z}_{k,r_2(k+1,s)}(\bx)\big)_{i_2+\delta_k,j_2},
 	\\&\hspace{4cm}
 	\big(\bar{z}_{k,r_3(k+1,s)}(\bx)\big)_{i_2,j_2+\delta_k},\big(\bar{z}_{k,r_4(k+1,s)}(\bx)\big)_{i_2+\delta_k,j_2+\delta_k}\Big) \Big|
 	\\
 	&~~
 	\leq\max_{(i_2,j_2)\in N^{(k+1)}_{(i,j)}}
 	C \cdot
 	\Bigg(
 	\Big|
 	\big(z_{k,r_1(k+1,s)}(\bx)\big)_{i_2,j_2}
 	-
 	\big(\bar{z}_{k,r_1(k+1,s)}(\bx)\big)_{i_2,j_2}
 	\Big|^2
 	\\
 	&
 	\quad
 	+
 	\Big|
 	\big(z_{k,r_2(k+1,s)}(\bx)\big)_{i_2+\delta_k,j_2}
 	-
 	\big(\bar{z}_{k,r_2(k+1,s)}(\bx)\big)_{i_2+\delta_k,j_2}
 	\Big|^2
 	\\
 	&
 	\quad
 	+
 	\Big| 
 	\big(z_{k,r_3(k+1,s)}(\bx)\big)_{i_2,j_2+\delta_k}
 	-
 	\big(\bar{z}_{k,r_3(k+1,s)}(\bx)\big)_{i_2,j_2+\delta_k}
 	\Big|^2
 	\\
 	&
 	\quad
 	+
 	\Big|   
 	\big(z_{k,r_4(k+1,s)}(\bx)\big)_{i_2+\delta_k,j_2+\delta_k}
 	-
 	\big(\bar{z}_{k,r_4(k+1,s)}(\bx)\big)_{i_2+\delta_k,j_2+\delta_k}
 	\Big|^2
 	\Big)^{1/2}
 	\\
 	&
 	\quad
 	+ \|
 	g_{k+1,s}
 	-
 	\bar{g}_{k+1,s}
 	\|_{[-2,2]^4,\infty}
 	\\
 	&
 	\stackrel{\eqref{ple4eq2}}{\leq}
 	(2 \cdot C) \cdot 
 	(2C+1)^{k-1}
 	\cdot \max_{i \in \{1,
 		\dots,k\}, s \in \{1, \dots,b_i \}}
 	\|                      g_{i,s}
 	-
 	\bar{g}_{i,s}
 	\|_{[-2,2]^4,\infty}
 	\\
 	&
 	\quad
 	+
 	\|
 	g_{k+1,s}
 	-
 	\bar{g}_{k+1,s}
 	\|_{[-2,2]^4,\infty}
 	\\
 	&~~
 	\leq
 	(2C+1)^{k} \cdot 
 	\max_{i \in \{1, \dots,k+1\}, s \in \{1, \dots, b_i\}}
 	\|                      g_{i,s}
 	-
 	\bar{g}_{i,s}
 	\|_{[-2,2]^4,\infty}
 \end{align*}
 for all  $s \in \{1, \dots, b_{k+1}\}$, $(i,j)\in\{1,\dots,d_1(k+1)\}\times\{1,\dots,d_2(k+1)\}$
 and $\bx\in[0,1]^{\{1,\dots,d_1\}\times\{1,\dots,d_2\}}$.
 \hfill $\Box$
 
 Our main approximation result is the following lemma which enables
 us to construct a convolutional neural network from feedforward
 neural networks approximating the functions in the definition
 of the hierarchical max-pooling model with feature constraint and local max-pooling parameter.
 
 \begin{lemma}
 	\label{le7}
 	Assume that the image dimensions $d_1$ and $d_2$ satisfy
 	\begin{equation*}
 		d_1=2^l\cdot m_1-1\quad d_2=2^l\cdot m_2-1
 	\end{equation*}
 	for some $m_1,m_2\in\N\setminus\{1\}$ and that the functions $\bar{g}_{r,s}:\R^4\rightarrow\R$ in definition \eqref{defmq} of $\bar{m}$ are standard feedforward neural networks with $L_{net}\in\N$ hidden layers and $r_{net}\in\N$ neurons per hidden layer
 	for $r \in \{1, \dots,l\}$
 	and $s \in \{1, \dots, b_r\}$.
 	Furthermore, set $b_{max}=\max\{b_1,\dots,b_{l}\}$, choose the parameters
 	\[
 	l_{net}=l,\quad\tilde{\bd}=(\tilde{d}_1,\tilde{d}_2)=(d_1(l),d_2(l))\quad\mbox{and}\quad z=b_{max}\cdot(L_{net}+1),
 	\]
 	and for $r\in\{1,\dots,l_{net}\}$ set
 	\[
 	k_r=2\cdot b_{max} + r_{net},\quad M_{r}=\delta_{r-1}+1,\quad s_r=n_r.
 	\]
 	Then there exists some $m_{net}\in \F_{1}\left(\big(l_{net},\bk,\bM,z,\bs,\tilde{\bd}\big)\right)$ such that
 	\[
 	\bar{m}(\bx) = m_{net}(\bx)
 	\]
 	holds for all $\bx \in [0,1]^{\{1, \dots, d_1\} \times \{1, \dots, d_2\}}$.
 \end{lemma}
 
 \noindent
 {\bf Proof.}
 The class of convolutional neural networks $\F_{1}\left(\big(l_{net},\bk,\bM,z,\bs,\tilde{\bd}\big)\right)$ with parameters defined in Lemma \ref{le7} consists of functions of the form
 \begin{equation}
 	\label{ple6eq1}
 	f_{out}^{(\tilde{d}_1,\tilde{d}_2)}\circ o^{(z)}_{(k_{l-1},k_{l}),M_l}\circ f_{max}^{(n_{l-1})}\circ o^{(z)}_{(k_{l-2},k_{l-1}),M_{l-1}}\circ\dots\circ f_{max}^{(n_1)}\circ o^{(z)}_{(1,k_{1}),M_1}.
 \end{equation}
 The idea of the proof is to successively calculate the values 
 \[
 \big((\bar{y}_{r,s}(\bx))_{(i,j)}\big)_{(i,j)\in\{1,\dots,d_1(r)\}\times\{1,\dots,d_2(r)\}}
 \]
 from equation \eqref{appry} in distinct channels of the convolutional block $o_{(k_{r-1},k_{r}),M_r}^{(z)}$ corresponding to distinct values of $s\in\{1,\dots,b_r\}$ by applying Lemma 2 for each $s\in\{1,\dots,b_r\}$.
 Once computed, values are then propagated to the next layer using the weights from equation \eqref{ple1eq1}.
 The values
 \[
 \big((\bar{z}_{r,s}(\bx))_{(i,j)}\big)_{(i,j)\in\{1,\dots,\lceil d_1(r)/n_r\rceil\}\times\{1,\dots,\lceil d_2(r)/n_r\rceil)\}}
 \]
 from equation \eqref{apprz} can then be calculated by the local max-pooling layer $f_{max}^{(n_r)}$.
 
 Since we will show the result by induction on the $l$ convolutional blocks, we define a recursive representation of \eqref{ple6eq1}.
 We define $f^{(0)}:\R^{\{1,\dots,d_1\}\times\{1,\dots,d_2\}}\rightarrow\R^{\{1,\dots,d_1\}\times\{1,\dots,d_2\}\times\{1\}}$ by
 \[
 \big(f^{(0)}(\bx)\big)_{(i,j),1}=x_{i,j}
 \]
 for $(i,j)\in\{1,\dots,d_1\}\times\{1,\dots,d_2\}$ and for $r=1,\dots,l$ we recursively define the functions
 \[
 g^{(r)}=o_{(k_{r-1},k_{r}),M_r}^{(z)}\circ f^{(r-1)},
 \]
 where we set $k_0=1$ and define
 \[
 f^{(r)}=f_{max}^{(n_r)}\circ g^{(r)}.
 \]
 We show that we can choose the weights of $f^{(l)}$ such that the following property holds for $r\in\{0,\dots,l\}$ by induction on $r$:
 \begin{itemize}
 	\item [($\ast$)]
 	For all $s\in\{1,\dots,b_r\}$, $(i,j)\in\{1,\dots,\lceil d_1(r)/n_r\rceil\}\times\{1,\dots,\lceil d_2(r)/n_r\rceil\}$ and $\bx\in[0,1]^{\{1,\dots,d_1\}\times\{1,\dots,d_2\}}$ it holds that
 	\begin{equation*}
 		\label{propl6}
 		\Big(f^{(r)}(\bx)\Big)_{(i,j),s}=\big(\bar{z}_{r,s}(\bx)\big)_{(i,j)}.
 	\end{equation*}
 \end{itemize}
 Because we have $b_0=n_0=1$ property ($\ast$) is true  for $r=0$ by definition.
 Now assume property ($\ast$) is true for $r\in\{0,\dots,l-1\}$.
 We show that property ($\ast$) holds for $r+1$ in two steps.
 
 {\it In the first step} we show that there is a convolutional block $o_{(k_{r},k_{r+1}),M_{r+1}}^{(z)}$ such that
 \begin{equation}
 	\label{ple5eq3}
 	\begin{split}
 		&\Big(g^{(r+1)}(\bx)\Big)_{(i,j),s}\\
 		&=\sigma\Big(\bar{g}_{r+1,s}\Big(
 		\big(\bar{z}_{r,r_1(r+1,s)}(\bx)\big)_{(i,j)},
 		\big(\bar{z}_{r,r_2(r+1,s)}(\bx)\big)_{(i+\delta_{r},j)},
 		\\&\hspace{3.5cm}
 		\big(\bar{z}_{r,r_3(r+1,s)}(\bx)\big)_{(i,j+\delta_{r})},\big(\bar{z}_{r,r_4(r+1,s)}(\bx)\big)_{(i+\delta_{r},j+\delta_{r})}\Big)\Big)
 	\end{split}
 \end{equation}
 for all $\bx\in[0,1]^{\{1,\dots,d_1\}\times\{1,\dots,d_2\}}$, $s\in\{1,\dots,b_{r+1}\}$ and 
 $(i,j)\in\{1,\dots,\lceil d_1(r)/n_r\rceil-\delta_r\}\times\{1,\dots,\lceil d_2(r)/n_r\rceil-\delta_r\}.$
 A convolutional block $o_{(k_{r},k_{r+1}),M_{r+1}}^{(z)}$ is of the form
 \[
 o_{(k_{r},k_{r+1}),M_{r+1}}^{(z)}=o_{(k_{r+1},k_{r+1}),M_{r+1},\bw_z}\circ\dots\circ o_{(k_{r},k_{r+1}),M_{r+1},\bw_1}
 \]
 for weight vectors $\bw_1,\dots,\bw_z$ defined as in Section 3.1. Since we want to choose the weight vectors $\bw_1,\dots,\bw_z$ layer by layer, we
 set
 \[
 o^{(t)}=o_{(k_{r+1},k_{r+1}),M_{r+1},\bw_t}\circ\dots\circ o_{(k_{r},k_{r+1}),M_{r+1},\bw_1}
 \]
 for $t=1,\dots,z$.
 
 We successively use Lemma \ref{le2} for the computation of each network 
 \begin{equation}
 	\label{ple7eq3}
 	\begin{split}
 		&\sigma\Bigg(\bar{g}_{r+1,s}\Big(
 		\big(\bar{z}_{r,r_1(r+1,s)}(\bx)\big)_{(i,j)},
 		\big(\bar{z}_{r,r_2(r+1,s)}(\bx)\big)_{(i+\delta_{r},j)},
 		\\&\hspace{3.5cm}
 		\big(\bar{z}_{r,r_3(r+1,s)}(\bx)\big)_{(i,j+\delta_{r})},\big(\bar{z}_{r,r_4(r+1,s)}(\bx)\big)_{(i+\delta_{r},j+\delta_{r})}\Big)\Bigg)
 	\end{split}
 \end{equation} 
 for $s=1,\dots,b_{r+1}$ and store the computed values 
 in the corresponding channels 
 \[1,\dots, b_{r+1}\]
 by using equation \eqref{ple1eq1}.
 With the idea of propagating computed values to the next layer by equation \eqref{ple1eq1} and because of the induction hypothesis ($\ast$) we can choose the weights of our convolutional block in the channels 
 \[b_{max}+1,\dots,b_{max}+b_r\]
 such that
 for all $t\in\{1,\dots,b_{max}\cdot(L_{net}+1)\}$, $s\in\{1,\dots,b_r\}$ and $(i,j)\in\{1,\dots,\lceil d_1(r)/n_r\rceil\}\times\{1,\dots,\lceil d_2(r)/n_r\rceil\}$ it holds that
 \[\Big(\big(o^{(t)}\circ f^{(r)}\big)(\bx)\Big)_{(i,j),b_{max}+s}=\big(\bar{z}_{r,s}(\bx)\big)_{(i,j)}\]
 for all $\bx\in[0,1]^{\{1,\dots,d_1\}\times\{1,\dots,d_2\}}$.
 Now, by using Lemma \ref{le2} with parameters
 \[
 s_m=\IND_{\N\setminus\{1\}}(s)\cdot b_{max}+r_m(r+1,s)
 \]
 for $m\in\{1,\dots,4\}$ and $s_5=s$,
 we can calculate the values \eqref{ple7eq3}  in layers 
 \[(s-1)\cdot(L_{net}+1)+1,\dots,s\cdot(L_{net}+1)\]
 by choosing corresponding weights 
 in channels 
 \[s,2\cdot b_{max}+1,\dots,2\cdot b_{max}+r_{net}\]
 such that we have
 \begin{align*}
 	&\Big(\big(o^{(s\cdot(L_{net}+1))}\circ f^{(r)}\big)(\bx)\Big)_{(i,j),s}\\
 	&=\sigma\Bigg(\bar{g}_{r+1,s}\Big(
 	\big(\bar{z}_{r,r_1(r+1,s)}(\bx)\big)_{(i,j)},
 	\big(\bar{z}_{r,r_2(r+1,s)}(\bx)\big)_{(i+\delta_{r},j)},
 	\\&\hspace{3.5cm}
 	\big(\bar{z}_{r,r_3(r+1,s)}(\bx)\big)_{(i,j+\delta_{r})},\big(\bar{z}_{r,r_4(r+1,s)}(\bx)\big)_{(i+\delta_{r},j+\delta_{r})}\Big)\Bigg)
 \end{align*}
 for all $\bx\in[0,1]^{\{1,\dots,d_1\}\times\{1,\dots,d_2\}}$, $s\in\{1,\dots,b_{r+1}\}$ and $(i,j)\in\{1,\dots,\lceil d_1(r)/n_r\rceil-\delta_r\}\times\{1,\dots,\lceil d_2(r)/n_r\rceil-\delta_r\}$.
 Once a value has been saved in layer $s\cdot(L_{net}+1)$ for $s\in\{1,\dots,b_{r+1}\}$, it will be propagated to the next layer using equation \eqref{ple1eq1} such that we have
 \begin{align*}
 	&\Big(\big(o^{(b_{max}\cdot(L_{net}+1))}\circ f^{(r)}\big)(\bx)\Big)_{(i,j),s}\\
 	&=\sigma\Bigg(\bar{g}_{r+1,s}\Big(
 	\big(\bar{z}_{r,r_1(r+1,s)}(\bx)\big)_{(i,j)},
 	\big(\bar{z}_{r,r_2(r+1,s)}(\bx)\big)_{(i+\delta_{r},j)},
 	\\&\hspace{3.5cm}
 	\big(\bar{z}_{r,r_3(r+1,s)}(\bx)\big)_{(i,j+\delta_{r})},\big(\bar{z}_{r,r_4(r+1,s)}(\bx)\big)_{(i+\delta_{r},j+\delta_{r})}\Big)\Bigg)
 \end{align*}
 for all $\bx\in[0,1]^{\{1,\dots,d_1\}\times\{1,\dots,d_2\}}$, $s\in\{1,\dots,b_{r+1}\}$ and $(i,j)\in\{1,\dots,\lceil d_1(r)/n_r\rceil-\delta_r\}\times\{1,\dots,\lceil d_2(r)/n_r\rceil-\delta_r\}$, 
 which implies equation \eqref{ple5eq3}.
 
 {\it In the second step}, we conclude from equation \eqref{ple5eq3} that porperty ($\ast$) is satisfied for $r+1$.
 First note that $d_1(r+1)=\lceil d_1(r)/n_r\rceil-\delta_r$ and $d_2(r+1)=\lceil d_2(r)/n_r\rceil-\delta_r$.
 Since $n_{r+1}$ divides $d_1(r+1)$ and $d_2(r+1)$ by Lemma \ref{le8} below we have
 \begin{align*}
 	N^{(r+1)}_{(i,j)}&=\{(i-1)\cdot n_{r+1}+1,\dots,i\cdot n_{r+1}\}\times\{(j-1)\cdot n_{r+1}+1,\dots,j\cdot n_{r+1}\}
 \end{align*}
 for all $(i,j)\in\{1,\dots,\lceil d_1(r+1)/n_{r+1}\rceil\}\times\{1,\dots,\lceil d_2(r+1)/n_{r+1}\rceil\}$ and therefore we get together with equation \eqref{ple5eq3}
 \begin{align*}
 	\Big(f^{(r+1)}(\bx)\Big)_{(i,j),s}
 	&=\Big(\big(f_{max}^{(n_{r+1})}\circ g^{(r+1)}\big)(\bx)\Big)_{(i,j),s}\\
 	&=\max_{(i_2,j_2)\in N^{(r+1)}_{(i,j)}} \big(g^{(r+1)}(\bx)\big)_{(i_2,j_2),s}\\
 	&\stackrel{\eqref{ple5eq3}}{=}\max_{(i_2,j_2)\in N^{(r+1)}_{(i,j)}} \big(\bar{y}_{r+1,s}(\bx)\big)_{(i_2,j_2)}\\
 	&=\big(\bar{z}_{r+1,s}(\bx)\big)_{(i,j)}
 \end{align*}
 for all $\bx\in[0,1]^{\{1,\dots,d_1\}\times\{1,\dots,d_2\}}$, $s\in\{1,\dots,b_{r+1}\}$ and $(i,j)\in\{1,\dots,\lceil d_1(r+1)/n_{r+1}\rceil\}\times\{1,\dots,\lceil d_2(r+1)/n_{r+1}\rceil\}$, which implies property ($\ast$) holds for all $r\in\{0,\dots,l\}$.
 
 Next we choose the output weights 
 \[
 \bw_{out}=(w_{s})_{
 	s \in \{1, \dots, k_l\}
 },
 \]
 by setting $w_1=1$ and $w_{s_2}=0$ for $s_2\in\{2,\dots,k_l\}$. This implies that the output of our network is given by
 \begin{align*}
 	m_{net}(\bx)&=\Big(f_{out}^{(\tilde{d_1},\tilde{d_2})}\circ f^{(l)}\Big)(\bx)\\
 	&=\max\left\{\big(\bar{z}_{l,1}(\bx)\big)_{(i,j)} : (i,j)\in\{1,\dots,d_1(l)\}\times\{1,\dots,d_2(l)\}\right\}\\
 	&=\bar{m}(\bx).
 \end{align*}
 for all $\bx\in[0,1]^{\{1,\dots,d_1\}\times\{1,\dots,d_2\}}$.
 \hfill $\Box$
 \begin{lemma}
 	\label{le8}
 	Let $l,m\in\N$  with $m>1$ and let $n_0,n_1,\dots,n_{l}\in\{2^0,2^1,\dots,2^{l-1}\}$ with $n_0=n_l=1$. Set 
 	$d=2^l\cdot m-1$ 
 	and $\delta_k={2^{k}}/{\prod_{i=0}^{k}n_i}$
 	and assume that
 	\begin{equation}
 		\label{l8eq1}
 		\delta_k\geq 1
 	\end{equation} 
 	holds for $k=1,\dots,l$.
 	For $k=1,\dots,l$ we define recursively the dimensions
 	\[
 	d(k)=	\left\lceil
 	\frac{d(k-1)}{n_{k-1}}
 	\right\rceil
 	-\delta_{k-1},
 	\]
 	where we set $d(0)=d$.	
 	
 	Then for any $k\in\{0,\dots,l\}$ it holds that
 	\[
 	d(k)=\frac{2^{k}}{\prod_{i=0}^{k-1}n_i}\cdot(2^{l-k}\cdot m-1).
 	\]
 	In particular, $n_k$ then divides $d(k)$.
 \end{lemma}
 \noindent
 {\bf Proof.}
 We show the assertion by an induction on $k$. For $k=0$ the assertion is true because of the definition. Now suppose the assertion is true for some $k\in\{0,\dots,l-1\}$. Then we have 
 \begin{align*}
 	d(k+1)&=\left\lceil\frac{d(k)}{n_k}\right\rceil-\delta_{k}\\
 	&=\frac{2^{k}}{\prod_{i=0}^{k}n_i}\cdot(2^{l-k}\cdot m-1)-\frac{2^{k}}{\prod_{i=0}^{k}n_i}\\
 	&=\frac{2^{k}}{\prod_{i=0}^{k}n_i}\cdot(2^{l-k}\cdot m-2)\\
 	&=\frac{2^{k+1}}{\prod_{i=0}^{k}n_i}\cdot(2^{l-(k+1)}\cdot m-1).
 \end{align*}
 Assumption \eqref{l8eq1} then yields $\delta_{k+1}\in\N$ by the choice of $n_0,\dots,n_{k+1}$, which implies that $n_{k+1}$ divides $d(k+1)$.
 \hfill $\Box$

 \section{A bound on the covering number}
 Our next result enables us to bound the covering number
 of our function class $\F_3(\btheta)$ defined as in Section \ref{se3}.
 
 \begin{lemma}
 	\label{le9}
 	Let $\sigma(x)=\max\{x,0\}$ be the ReLU activation function,
 	define 
 	$\F_{3}(\btheta)$ 
 	as in Section \ref{se3} with parameter vector $\theta=(L,\bk,\bM,z,s,\tilde{\bd})$ and set
 	\[
 	k_{max}=\max\left\{k_1, \dots, k_{L}\right\},
 	\quad
 	M_{max}=\max\{ M_1, \dots, M_L\}.
 	\]
 	Assume $d_1\cdot d_2>1$ and $c_4 \cdot \log n \geq 2$.
 	Then we have for any
 	$\epsilon \in (0,1)$:
 	\begin{eqnarray*}
 		&&
 		\sup_{\bx_1^n \in (\R^{ \{ 1,
 				\dots, d_1\} \times \{1, \dots, d_2\}})^n} \log\left(
 		\mathcal{N}_1 \left(\epsilon,T_{c_{4} \cdot \log n}  \F_3(\btheta), \bx_1^n\right) \right)
 		\\
 		&&
 		\leq
 		c_{11} \cdot z^2 \cdot \log(z \cdot d_1 \cdot d_2) \cdot
 		\log \left(
 		\frac{c_4\cdot\log n}{\epsilon}
 		\right)
 	\end{eqnarray*}
 	for some constant $c_{11} >0$ which depends only on $L$, $k_{max}$ and $M_{max}$.
 \end{lemma}
With the aim of proving Lemma \ref{le9}, we first have to study the VC dimension of our function class $\F_3\left(\btheta\right)$. For a class of subsets of $\R^d$, the VC dimension is defined as follows: 
\begin{definition}
	Let $\A$ be a class of subsets of $\R^d$ with $\A\neq\emptyset$ and $m\in\mathbb N$.
	\begin{enumerate}
		\item For $\bx_1,...,\bx_m\in\mathbb R^d$ we define
		\[s(\mathcal A,\left\{\bx_1,...,\bx_m\right\})\coloneqq|\left\{A\cap\{\bx_1,...,\bx_m\}~:~A\in\mathcal A\right\}|.\]
		\item Then the $m$th \textbf{shatter coefficient} $S(\mathcal A,m)$ of $\mathcal A$ is defined by
		\[S(\mathcal A,m)\coloneqq\max_{\{\bx_1,...,\bx_m\}\subset\mathbb R^d}s(\mathcal A,\{\bx_1,...,\bx_m\}).\]
		\item The \textbf{VC dimension} (Vapnik-Chervonenkis-Dimension) $V_{\mathcal A}$ of $\mathcal A$ is defined as
		\[V_{\mathcal A}\coloneqq\sup\{m\in\mathbb N~:~S(\mathcal A,m)=2^m\}.\]
	\end{enumerate}
\end{definition}
For a class of real-valued functions, we define the VC dimension as follows:
\begin{definition}
	Let $\mathcal H$ denote a class of functions from $\R^d$ to $\{0,1\}$ and let $\F$ be a class of real-valued functions.
	\begin{enumerate}
		\item For any non-negative integer $m$, we define the \textbf{growth function} of $H$ as
		\[\Pi_{\mathcal H}(m)\coloneqq\max_{\bx_1,\dots,\bx_m\in\R^d}|\{(h(\bx_1),\dots,h(\bx_m)) : h\in H\}|.\]
		\item The \textbf{VC dimension} (Vapnik-Chervonenkis-Dimension) of $\mathcal H$ we define as
		\[\VC(\mathcal H)\coloneqq\sup\{m\in\N : \Pi_{\mathcal H}(m)=2^m\}.\]
		\item For $f\in\F$ we denote $\sgn(f)\coloneqq\IND_{\{f\geq0\}}$ and $\sgn(\F)\coloneqq\{\sgn(f) : f\in\F\}$. Then the \textbf{VC dimension} of $\F$ is defined as 
		\[\VC(\F)\coloneqq\VC(\sgn(\F)).\]
	\end{enumerate}
\end{definition}
A well-known connection between both definitions is given by the following lemma.
\begin{lemma}
	\label{le13}
	Suppose $\F$ is a class of real-valued functions on $\R^d$.
	Furthermore, we define
	\[\F^+\coloneqq\{\{(\bx,y)\in\R^d\times\R : f(\bx)\geq y\} : f\in\F\}\]
	and define the class $\mathcal H$ of real-valued functions on $\R^d\times\R$ by
	\[\mathcal H\coloneqq\{h((\bx,y))=f(\bx)-y : f\in\F\}.\]
	Then, it holds that
	\[V_{\F^+}=\VC(\mathcal H).\]
\end{lemma}
\noindent
{\bf Proof.}
See, e.g., Lemma 8 in Kohler, Krzy\.zak and Walter (2020).
\hfill $\Box$ 

~\\
\noindent
In order to bound the VC dimension of our function class, we need the following two auxiliary results. The first one is also known as weighted AM-GM inequality.
\begin{lemma}
	\label{le14}
	Suppose $x_1,\dots,x_n>0$ and $w_1,\dots,w_n>0$. We denote $w\coloneqq\sum_{i=1}^{n}w_i$. Then, it holds that 
	\begin{equation}
		\label{eq:amgm}
		\prod_{i=1}^{n}\left(\frac{x_i}{w_i}\right)^{w_i}\leq\left(\frac{\sum_{i=1}^{n}x_i}{w}\right)^{w}.
	\end{equation}
\end{lemma}
\noindent
{\bf Proof.}
See, e.g., Lemma 9 in \cite{KoKrWa2020}.
\hfill $\Box$
\begin{lemma}
	\label{le15}
	Suppose $W,m\in\N$ with $W\leq m$ and let $f_1,...,f_m$ be polynomials of degree at most $D$ in $W$ variables. Define
	\[
	K\coloneqq|\{\left(\sgn(f_1(\ba)),\dots,\sgn(f_m(\ba))\right) : \ba\in\R^{W}\}|.
	\]
	Then we have 
	\[
	K\leq2\cdot\left(\frac{2\cdot e\cdot m\cdot D}{W}\right)^{W}.
	\]
\end{lemma}
\noindent
{\bf Proof.} See Theorem 8.3 in \cite{Anthony1999}.
\hfill $\Box$

~\\
The next two lemmas provide a modification of Theorem 6 in \cite{Bartlett2019}.
\begin{lemma}
	\label{le16}
	Let $k',m\in\N$ and $D,W\in\N_0$ with $W\leq m$ and let $I=\{1,\dots,i_1\}\times\{1,\dots,i_2\}$ be an index set with $i_1,i_2\in\N$. Furthermore, let $\mathcal S$ be a finite partition of $\R^W$ (where $\R^0\coloneqq\{0\}$) and let
	\[
	f_1,\dots,f_m:\R^{W}\rightarrow\R^{I\times\{1,\dots,k'\}}
	\]
	be functions such that the following property holds:
	\begin{itemize}[label={}]
		\item For each element $S\in\mathcal S$, each $(i,j) \in I$, each $s\in\{1,\dots,k'\}$, each $t\in\{1,\dots,m\}$  when $\ba$ varies in $S$,
		\[\big(f_t(\ba)\big)_{(i,j),s}\]
		is a fixed polynomial function in the $W$ variables $\ba\in\R^{W}$, of total degree no more than $D$.
	\end{itemize}
	
	\noindent
	{\bf a)}
	Let $o_{(k',k),M,\bw}:\R^{I\times\{1,\dots,k'\}}\rightarrow\R^{I\times\{1,\dots,k\}}$ be a convolutional layer as defined in Subsection 3.1 with $k\in\N$ channels and a filter size of $M\in\N$. Then the weight vector
	\[
	\bw= \left(
	\left(
	w_{i,j,s_1,s_2}
	\right)_{
		1 \leq i,j \leq M, s_1 \in \{1, \dots, k'\}, s_2 \in \{1, \dots, k\}
	}
	,
	\left(
	w_{s_2}
	\right)_{
		s_2 \in \{1, \dots, k\}.
	}         
	\right)
	\]
	consists of $|\bw|=M^2\cdot k'\cdot k+k$ real-valued weights.
	For any $t\in\{1,\dots,m\}$ let $g_t:\R^{W+|\bw|}\rightarrow\R^{I\times\{1,\dots,k\}}$ be the function defined by
	\[
	g_t(\ba,\bw)=\big(o_{(k',k),M,\bw}\circ f_t\big)(\ba).
	\]
	Then there exist a partition $\mathcal S'$ of $\R^{W+|\bw|}$
	with the following two properties:
	\begin{enumerate}
		\item \begin{equation}
			\label{prop1}
			|\mathcal S'|\leq|\mathcal S|\cdot2\left(\frac{2\cdot e\cdot (m\cdot i_1\cdot i_2\cdot k)\cdot (D+1)}{W+|\bw|}\right)^{W+|\bw|},
		\end{equation}
		\item For each element $S\in\mathcal S'$, each $(i,j) \in I$, each $s\in\{1,\dots,k\}$, each $t\in\{1,\dots,m\}$ when $(\ba,\bw)$ varies in $S$,
		\[\big(g_{t}(\ba,\bw)\big)_{(i,j),s}\]
		is a fixed polynomial function in the $W+|\bw|$ variables $(\ba,\bw)\in\R^{W+|\bw|}$, of total degree no more than $D+1$.
	\end{enumerate}
	\noindent
	{\bf b)}
	Let
	\[ 
	f_{sub}^{(n)}:\R^{I\times\{1,\dots,k'\}}\rightarrow\R^{\left\{1,\dots,\left\lceil\frac{i_1}{n}\right\rceil\right\}\times\left\{1,\dots,\left\lceil\frac{i_2}{n}\right\rceil\right\}\times \{1,\dots,k'\}}
	\]
	be a subsampling layer defined as in Subsection 3.3 with parameter $n\in\N$. 
	For any $t\in\{1,\dots,m\}$ let 
	\[
	g_t:\R^{W}\rightarrow\R^{\left\{1,\dots,\left\lceil\frac{i_1}{n}\right\rfloor\right\}\times\left\{1,\dots,\left\lceil\frac{i_2}{n}\right\rceil\right\}\times \{1,\dots,k'\}}
	\]
	be the function defined by
	\[
	g_t(\ba)=\big(f_{sub}^{(n)}\circ f_t\big)(\ba).
	\]
	Then the following property holds:
	\begin{itemize}[label={}]
		\item For each element $S\in\mathcal S$, each $(i,j) \in I$, each $s\in\{1,\dots,k\}$, each $t\in\{1,\dots,m\}$ when $\ba$ varies in $S$,
		\[\big(g_{t}(\ba)\big)_{(i,j),s}\]
		is a fixed polynomial function in the $W$ variables $\ba\in\R^{W}$, of total degree no more than $D$.
	\end{itemize}
	
	\noindent
	{\bf c)}
	Let
	\[ 
	f_{\bw_{out}}^{(\tilde{d}_1,\tilde{d}_2)}:\R^{I\times\{1,\dots,k'\}}\rightarrow\R
	\]
	be a output layer defined as in Subsection 3.5 with output bounds $\tilde{d}_1,\tilde{d}_2\in I$. The output layer depends on a weight vector
	\[
	\bw_{out}=(w_{s})_{s\in\{1,\dots,k'\}},
	\]
	which consists of $|\bw_{out}|=k'$ real-valued weights.
	For any $t\in\{1,\dots,m\}$ let $g_t:\R^{W+|\bw_{out}|}\rightarrow\R$ be the function defined by
	\[
	g_t(\ba,\bw_{out})=\big(f_{\bw_{out}}^{(\tilde{d}_1,\tilde{d}_2)}\circ f_t\big)(\ba).
	\]
	Then there exist a partition $\mathcal S'$ of $\R^{W+|\bw_{out}|}$
	with the following two properties:
	\begin{enumerate}
		\item \begin{equation}
			\label{prop3}
			|\mathcal S'|\leq|\mathcal S|\cdot2\left(\frac{2\cdot e\cdot(i_1^2\cdot i_2^2\cdot m)\cdot(D+1)}{W+|\bw_{out}|}\right)^{W+|\bw_{out}|},
		\end{equation}
		\item For each element $S\in\mathcal S'$, each $t\in\{1,\dots,m\}$ when $(\ba,\bw_{out})$ varies in $S$,
		\[g_{t}(\ba,\bw_{out})\]
		is a fixed polynomial function in the $W+|\bw_{out}|$ variables $(\ba,\bw_{out})\in\R^{W+|\bw_{out}|}$, of total degree no more than $D+1$.
	\end{enumerate}
\end{lemma}
{\bf Proof.}
{\bf a)}
For $S\in\mathcal S$, $t\in\{1,\dots,m\}$, $(i,j)\in I$ and $s_2\in\{1,\dots,k'\}$ let $p_{S,t,(i,j),s_2}(\ba)$ denote the function $\big(f_{t}(\ba)\big)_{(i,j),s_2}$ when $\ba\in S$. By assumption
$p_{S,t,(i,j),s_2}(\ba)$ is a polynomial with degree no more than $D$ in the $W$ variables of $\ba$ for any $t\in\{1,\dots,m\}$, $(i,j)\in I$ and $s_2\in\{1,\dots,k'\}$.
Hence for
any  $t\in\{1,\dots,m\}$, $(i,j)\in I$ and $s_2\in\{1,\dots,k\}$
\[
\sum_{s_1=1}^{k'}
\sum_{
	\substack{
		t_1,t_2 \in \{1, \dots, M\}
		\\(i+t_1-1,j+t_2-1)\in I}
}
w_{t_1,t_2,s_1,s_2}
\cdot
p_{S,t,(i+t_1-1,j+t_2-1),s_2}(\ba)
+
w_{s_2}
\] 
is a polynomial in the $W+|\bw|$ variables $(\ba,\bw)$ with total degree no more than $D+1$.
Then, by Lemma \ref{le15}, the collection of polynomials 
\begin{align*}
	&\left\{\sum_{s_1=1}^{k'}
	\sum_{
		\substack{
			t_1,t_2 \in \{1, \dots, M\}
			\\(i+t_1-1,j+t_2-1)\in I}
	}
	w_{t_1,t_2,s_1,s_2}
	\cdot
	p_{S,t,(i+t_1-1,j+t_2-1),s_2}(\ba)
	+
	w_{s_2} :\right.
	\\
	&\hspace{4.5cm}
	\left.  t\in\{1,\dots,m\}, (i,j)\in I,s_2\in\{1,\dots,k\}\vphantom{\sum_{\substack{t_1,t_2 \in \{1, \dots, M_r\}\\(i+t_1-1,j+t_2-1)\in D}}}\right\}
\end{align*}	
attains at most
\[\Pi\coloneqq2\left(\frac{2\cdot e\cdot(m\cdot i_1\cdot i_2\cdot k)\cdot(D+1)}{W+|\bw|}\right)^{W+|\bw|}\]
distinct sign patterns when $(\ba,\bw)\in\R^{W+|\bw|}$. 
Therefore, we can partition $S\times\R^{|\bw|}\subset\R^{W+|\bw|}$ into $\Pi$ subregions, such that all the polynomials don't change their signs within each subregion. Doing this for all regions $S\in\mathcal S$ we get our required partition $\mathcal S'$ by assembling all of these subregions. In particular, property 1 (inequality \eqref{prop1}) is then satisfied.
Fix some $S'\in\mathcal S'$. Notice that, when $(\ba,\bw)$ varies in $S'$, all the polynomials
\begin{align*}
	&\left\{\sum_{s_1=1}^{k'}
	\sum_{
		\substack{
			t_1,t_2 \in \{1, \dots, M\}
			\\(i+t_1-1,j+t_2-1)\in I}
	}
	w_{t_1,t_2,s_1,s_2}
	\cdot
	p_{S,t,(i+t_1-1,j+t_2-1),s_2}(\ba)
	+
	w_{s_2} :\right.
	\\
	&\hspace{4.5cm}
	\left.  t\in\{1,\dots,m\}, (i,j)\in I,s_2\in\{1,\dots,k\}\vphantom{\sum_{\substack{t_1,t_2 \in \{1, \dots, M_r\}\\(i+t_1-1,j+t_2-1)\in D}}}\right\}
\end{align*}
don't change their signs, hence when $(\ba,\bw)$ varies in $S'$
\begin{align*}
	&\big(g_{t}(\ba,\bw)\big)_{(i,j),s_2}\\
	&=\sigma\left(
	\sum_{s_1=1}^{k'}
	\sum_{
		\substack{
			t_1,t_2 \in \{1, \dots, M\}
			\\(i+t_1-1,j+t_2-1)\in I}
	}
	w_{t_1,t_2,s_1,s_2}
	\cdot
	f_{t,(i+t_1-1,j+t_2-1),s_2}(\ba)
	+
	w_{s_2} 
	\right)\\
	&=\max \left\{
	\sum_{s_1=1}^{k'}
	\sum_{
		\substack{
			t_1,t_2 \in \{1, \dots, M\}
			\\(i+t_1-1,j+t_2-1)\in I}
	}
	w_{t_1,t_2,s_1,s_2}
	\cdot
	f_{t,(i+t_1-1,j+t_2-1),s_2}(\ba)
	+
	w_{s_2}
	,0\right\}
\end{align*}
is either a polynomial of degree no more than $D+1$ in the $W+|\bw|$ variables of $(\ba,\bw)$ or a constant polynomial with value $0$ for all $(i,j)\in I$,  $s_2\in\{1,\dots,k\}$ and $t\in\{1,\dots,m\}$. Hence, property 2 is also satisfied and we are able to construct our desired partition $\mathcal S'$.

\noindent
{\bf b)}
The assertion is trivial.

\noindent
{\bf c)}
For $S\in\mathcal S$, $t\in\{1,\dots,m\}$, $(i,j)\in I$ and $s\in\{1,\dots,k'\}$ let $p_{S,t,(i,j),s}(\ba)$ denote the function $\big(f_{t}(\ba)\big)_{(i,j),s}$ when $\ba\in S$. By assumption
$p_{S,t,(i,j),s}(\ba)$ is a polynomial with degree no more than $D$ in the $W$ variables of $\ba$ for any $t\in\{1,\dots,m\}$, $(i,j)\in I$ and $s\in\{1,\dots,k'\}$.
Hence for
any  $t\in\{1,\dots,m\}$ and $(i,j)\in I$ 
\[
\sum_{s=1}^{k'}w_{s_2}\cdot p_{S,t,(i,j),s}(\ba)
\]
is a polynomial in the $W+|\bw_{out}|$ variables $(\ba,\bw_{out})$ with total degree no more than $D+1$.
Then by Lemma \ref{le15}, the collection of polynomials
\begin{align*}
	&\Bigg\{\sum_{s=1}^{k'}w_{s_2}\cdot p_{S,t,(i_1,j_1),s}(\ba)-\sum_{s=1}^{k'}w_{s_2}\cdot p_{S,t,(i_2,j_2),s}(\ba) : \\
	&\quad(i_1,j_1),(i_2,j_2)\in I, (i_1,j_1)\neq(i_2,j_2), t\in\{1,\dots,m\}
	\Bigg\}
\end{align*}
attains at most
\[\Pi\coloneqq2\left(\frac{2\cdot e\cdot(i_1^2\cdot i_2^2\cdot m)\cdot(D+1)}{W+|\bw_{out}|}\right)^{W+|\bw_{out}|}\]
distinct sign patterns when $(\ba,\bw_{out})\in\R^{W+|\bw_{out}|}$.
Therefore, we can partition $S\times\R^{|\bw_{out}|}\subset\R^{W+|\bw_{out}|}$ into $\Pi$ subregions, such that all the polynomials don't change their signs within each subregion. Doing this for all regions $S\in\mathcal S$ we get our required partition $\mathcal S'$ by assembling all of these subregions. In particular, property 1 (inequality \eqref{prop3}) is then satisfied.

Fix some $S'\in\mathcal S'$. Notice that, when $\ba$ varies in $S'$, all the polynomials 
\begin{align*}
	&\Bigg\{\sum_{s=1}^{k'}w_{s}\cdot p_{S,t,(i_1,j_1),s}(\ba)-\sum_{s=1}^{k'}w_{s}\cdot p_{S,t,(i_2,j_2),s}(\ba) : \\
	&\quad(i_1,j_1),(i_2,j_2)\in I, (i_1,j_1)\neq(i_2,j_2), t\in\{1,\dots,m\}
	\Bigg\}
\end{align*}
don't change their signs. Hence, there is a permutation $\pi_{t}$ of the set 
$
\{1,\dots,\tilde{d}_1\}\times\{1,\dots,\tilde{d}_2\}
$
for any $t\in\{1,\dots,m\}$ such that
\[\sum_{s=1}^{k'}w_{s}\cdot\big(f_{t}(\ba)\big)_{\pi_{t}((1,1)),s}\geq\dots\geq\sum_{s=1}^{k'}w_{s}\cdot\big(f_{t}(\ba)\big)_{\pi_{t}((\tilde{d}_1,\tilde{d}_2)),s}\]
for $(\ba,\bw_{out})\in S'$ and any $t\in\{1,\dots,m\}$. Therefore, it holds that
\begin{align*}
	g_{t}(\ba,\bw_{out})&=\max\left\{\sum_{s=1}^{k'}w_{s}\cdot\big(f_{t}(\ba)\big)_{(1,1),s},\dots,\sum_{s=1}^{k'}w_{s}\cdot\big(f_{t}(\ba)\big)_{(\tilde{d}_1,\tilde{d}_2),s}\right\}\\
	&=\sum_{s=1}^{k'}w_{s}\cdot\big(f_{t}(\ba)\big)_{\pi_{t}((1,1)),s},
\end{align*}
for $(\ba,\bw_{out})\in S'$. Since $\sum_{s=1}^{k'}w_{s}\cdot\big(f_{t}(\ba)\big)_{\pi_{t}((1,1)),s}$ is a polynomial within $S'$, also $g_{t}(\ba,\bw_{out})$ is a polynomial within $S'$ with degree no more than $D+1$ and in the $W+|\bw_{out}|$ variables of $(\ba,\bw_{out})\in\R^{W+|\bw_{out}|}$ for any $t\in\{1,\dots,m\}$.
\hfill $\Box$
\begin{lemma}
	\label{le17}
	Let $\sigma(x)=\max\{x,0\}$ be the ReLU activation function,
	define 
	$\F\coloneqq\F_{3}\left(\btheta\right)$
	as in Section \ref{se2} with parameter vector $\btheta=(L,\bk,\bM,z,s,\tilde{\bd})$ and set
	\[
	k_{max}=\max\left\{k_1, \dots, k_{L}\right\},
	\quad
	M_{max}=\max\{ M_1, \dots, M_L\}.
	\]
	Assume $d_1\cdot d_2>1$. Then, we have 
	\[
	V_{\F^+}\leq c_{12}\cdot z^2\cdot \log_2(z\cdot d_1\cdot d_2)
	\]
	for some constant $c_{12}>0$ which depends only on
	$L$, $k_{max}$ and $M_{max}$.
\end{lemma}

\noindent
{\bf Proof.}
We want to use Lemma \ref{le13} to bound $\mathcal V_{\F^+}$ by $\VC(\mathcal H)$, where $\mathcal H$ is the class of real-valued functions on $[0,1]^{\{1,\dots,d_1\}\times\{1,\dots,d_2\}}\times\R$  defined by
\[\mathcal H\coloneqq\{h((\bx,y))=f(\bx)-y : f\in\F\}.\]
To get an upper bound for the VC-dimension of $\mathcal H$, we will bound the growth function $\Pi_{\sgn(\mathcal H)}(m)$ for $m\in\N$. 
To bound the growth function $\Pi_{\sgn(\mathcal H)}(m)$, we fix aritrary input values \[(\bx_1,y_1),\dots,(\bx_m,y_m)\in [0,1]^{\{1,\dots, d_1\} \times \{1, \dots,
	d_2\}}\times\R\]
and obtain an upper bound for the growth function $\Pi_{\sgn(\mathcal H)}(m)$ by deriving an upper bound for 
\begin{equation}
	\label{DefK}
	K\coloneqq|\{(\sgn(h((\bx_1,y_1))),\dots,\sgn(h((\bx_m,y_m)))) : h\in\mathcal H\}|.
\end{equation}
Let $h\in\mathcal H$, then $h((\bx,y))=f(\bx)-y$ depends on a weight vector $\bw\in\R^W$ for some $W\in\N$, which denotes the number of weights. The weight vector $\bw$ is composed of the weights of the individual layers of the convolutional neural network $f$. Let $f_{\bw}\in\F$ denote the convolutional neural network with weight vector $\bw\in\R^W$.
For $k\in\{1,\dots,m\}$ we define functions $h_k:\R^{W}\rightarrow\R$ by
\[
h_k(\bw)=f_{\bw}(\bx_k)-y_k.
\]
Then formula \eqref{DefK} can be written as
\[K=|\{(\sgn(h_1(\bw)),\dots,\sgn(h_m(\bw))) : \bw\in\R^{W}\}|.\]
For any finite partition $\mathcal S$ of $\R^W$ it holds that
\begin{eqnarray}
	K\leq\sum_{S\in\mathcal S}|\{(\sgn(h_1(\bw)),\dots,\sgn(h_m(\bw)) : \bw\in S\}|.
	\label{sum}
\end{eqnarray}
In the sequel we will construct a partition $\mathcal S$ of $\R^W$ such that within each region $S\in\mathcal S$, the functions $h_k(\cdot)$		
are all fixed polynomials of bounded degree for $k\in\{1,\dots,m\}$,
so that each summand of equation \eqref{sum} can be bounded via Lemma \ref{le15}. 
We construct the partition $\mathcal S$ iteratively layer by layer, by applying Lemma \ref{le16} several times.
We start by counting the weights of the individual layers. 
We have
\begin{equation}
	\label{fak}
	\begin{split}
		f_{\bw}(\bx)
		&=f_{out}^{(\tilde{d}_1,\tilde{d}_2)}\circ f_{sub}^{(s)}\circ o^{(z)}_{(k_{L-1},k_{L}),M_L}\circ\dots\circ o^{(z)}_{(1,k_1),M_1}(\bx).%
	\end{split}
\end{equation}
with
\[
o_{(k_{j-1},k_j),M_j}^{(z)}(\bx)
=\big(
o_{(k_{j},k_j),M_j, \bw_{j,z}}\circ o_{(k_{j},k_j),M_j, \bw_{j,z-1}}\circ\dots\circ o_{(k_{j-1},k_j),M_j,\bw_{j,1}}
\big)(\bx)
\]
For $j\in\{1,\dots,L\}$ and $i\in\{1,\dots,z\}$ we set
\begin{equation}
	k_{j,i}'=
	\begin{cases}
		k_{j-1}&,\mbox{ if }i=1\\
		k_{j}&,\mbox{ else}
	\end{cases}
\end{equation}
(where $k_0=1$) and get
\[
|\bw_{j,i}|=M_j^2\cdot k_{j,i}'\cdot k_j+k_j
\]
for the number of weights in the $i$-th convolutional layer in the $j$-th convolutional block
for $j\in\{1,\dots,L\}$ and $i\in\{1,\dots,z\}$.
The number of output weights is given by
\[
|\bw_{out}|=k_L.
\]
and the total number of weights is then given by
\begin{equation}
	\label{eqW}
	\begin{split}
		&W=\sum_{j=1}^{L}\sum_{i=1}^{z}|\bw_{j,i}|+|\bw_{out}|\\
		&\leq L\cdot z\cdot(M_{max}^2\cdot k_{max}^2+k_{max})+k_{max}\\
		&\leq L\cdot z\cdot M_{max}^2\cdot k_{max}^2+(L\cdot z+1)\cdot k_{max}\\
		&\leq 2\cdot (L\cdot z+1)\cdot M_{max}^2\cdot k_{max}^2.
	\end{split}
\end{equation}
The number of weights used up to the $i$-th convolutional layer in the $j$-th convolutional block we denote by
\[W_{(j-1)\cdot z+i}=\sum_{j'=1}^{j-1}\sum_{i'=1}^{z}|\bw_{j',i'}|+\sum_{i'=1}^{i}|\bw_{j,i'}|\]
for $j\in\{1,\dots,L\}$ and $i\in\{1,\dots,z\}$.
In order to make use of Lemma \ref{le16} we assume in the following that $m$ is a positive integer with
\begin{equation}
	\label{eqV1}
	m\geq W
\end{equation}
and define $f_1,\dots,f_m:\{0\}\rightarrow\R^{\{1,\dots,d_1\}\times\{1,\dots,d_2\}\times\{1\}}$ by 
\[\big(f_k(0)\big)_{(i,j),1}=(\bx_k)_{i,j}\]
for $k\in\{1,\dots,m\}$ and set $\mathcal S=\{0\}$. Now we are able to apply Lemma \ref{le16} iteratively layer by layer to obtain a partition $\mathcal S'$ of $\R^W$ satisfying
\begin{equation}
	\label{ple14eq1}
	\begin{split}
		|\mathcal S'|&\leq\prod_{j=1}^{L}\prod_{i=1}^{z}2\cdot\left(\frac{2\cdot e\cdot(m\cdot d_1\cdot d_2\cdot k_j)\cdot((j-1)\cdot z+i)}{W_{(j-1)\cdot z+i}}\right)^{W_{(j-1)\cdot z+i}}\\
		&\hspace{2cm}
		\cdot2\left(\frac{2\cdot e\cdot m\cdot d_1^2\cdot d_2^2\cdot(z\cdot L+1)}{W}\right)^{W}
	\end{split}
\end{equation}
such that within each region $S\in\mathcal S'$, the functions $h_k:\R^W\rightarrow\R$		
are all fixed polynomials of degree no more than $z\cdot L+1$ for $k\in\{1,\dots,m\}$.
By condition \eqref{eqV1} and another application of Lemma \ref{le15} it holds for any $S'\in\mathcal S$ that
\begin{align*}
	&|\{(\sgn(h_1(\bw)),\dots,\sgn(h_m(\bw))) : \bw\in S'\}|\\
	&\leq2\cdot\left(\frac{2\cdot e\cdot m\cdot(z\cdot L+1)}{W}\right)^{W}.
\end{align*}
Now we are able to bound $K$ via equation \eqref{sum} and because $K$ is an upper bound for the growth function we get
\begin{equation}
	\label{In1}
	\begin{split}
		&\Pi_{\sgn(\mathcal H)}(m)\\
		&\leq|\mathcal S'|\cdot2\cdot\left(\frac{2\cdot e\cdot m\cdot\big(z\cdot L+1\big)}{W}\right)^{W}\\
		&\stackrel{\eqref{ple14eq1}}{\leq}2^{z\cdot L+2}\cdot\prod_{r=1}^{z\cdot L+2}\left(\frac{m\cdot x_r}{W_r}\right)^{W_r}\\
		&\stackrel{\eqref{eq:amgm}}{\leq}2^{z\cdot L+2}\left(\frac{m\cdot\sum_{r=1}^{z\cdot L+2}x_r}{\sum_{r=1}^{z\cdot L+2}W_r}\right)^{\sum_{r=1}^{z\cdot L+2}W_r}
	\end{split}
\end{equation}
with $W_{z\cdot L+2}=W_{z\cdot L+1}=W$, 
\[x_{z\cdot L+1}=2\cdot e\cdot d_1^2\cdot d_2^2\cdot(z\cdot L+1),\quad x_{z\cdot L+2}=2\cdot e\cdot (z\cdot L+1)\]
and
\[x_{(j-1)\cdot z+i}=	2\cdot e\cdot d_1\cdot d_2\cdot k_j\cdot((j-1)\cdot z+i)\]
for $j\in\{1,\dots,L\}$ and $i\in\{1,\dots,z\}$.
In the fourth line of \eqref{In1} we used inequality \eqref{eq:amgm} of Lemma \ref{le14}. 
Without loss of generality, we can assume that 
$\VC(\mathcal H)\geq\sum_{r=1}^{z\cdot L+2}W_r$
because in the case $\VC(\mathcal H)<\sum_{r=1}^{z\cdot L+2}W_r$ we have
\begin{align*}
	\VC(\mathcal H)&~<(z\cdot L+2)\cdot W\\
	&\stackrel{\eqref{eqW}}{\leq}2\cdot (z\cdot L+2)^2\cdot M_{max}^2\cdot k_{max}^2 \\
	&~\leq c_{12}\cdot z^2
\end{align*}
for some constant $c_{12}>0$ which only depends on $L$, $M_{max}$ and $k_{max}$ and get the assertion by Lemma \ref{le13}.
Hence we get by the definition of the VC--dimension and inequality \eqref{In1} (which only holds for $m\geq W$)
\[2^{\VC(\mathcal H)}=\Pi_{\sgn(\mathcal H)}(\VC(\mathcal H))
\leq2^{z\cdot L+2}\left(\frac{\VC(\mathcal H)\cdot\sum_{r=1}^{z\cdot L+2}x_r}{\sum_{r=1}^{z\cdot L+2}W_r}\right)^{\sum_{r=1}^{z\cdot L+2}W_r}
.\]
Since 
\[\sum_{r=1}^{z\cdot L+2}x_r> x_{z\cdot L+1}+x_{z\cdot L+2}\geq8\cdot e>16\]
and
\begin{align*}
	\sum_{r=1}^{z\cdot L+2}x_r&\leq(z\cdot L+2)^2\cdot2\cdot e\cdot d_1^2\cdot 2_2^2\cdot k_{max}\\
	&\leq (3\cdot z\cdot L\cdot d_1\cdot d_2)^2\cdot 2\cdot e\cdot k_{max}\\
	&\leq(7\cdot z\cdot L\cdot d_1\cdot d_2\cdot k_{max})^2
\end{align*}
Lemma \ref{le18} below (with parameters $R=\sum_{r=1}^{z\cdot L+2}x_r$, $m=\VC(\mathcal H)$, $w=\sum_{r=1}^{z\cdot L+2}W_r$ and $l=z\cdot L+2$) implies that
\begin{align*}
	\VC(\mathcal H)
	&~~~\leq z\cdot L+2+\left(\sum_{r=1}^{z\cdot L+2} W_r\right)\cdot\log_2\left(2\cdot\sum_{r=1}^{z\cdot L+2}x_r\cdot\log_2\left(\sum_{r=1}^{z\cdot L+2}x_r\right)\right)\\
	&~~~\leq z\cdot L+2+(z\cdot L+2)\cdot W\cdot \log_2\left(4\cdot(7\cdot z\cdot L\cdot d_1\cdot d_2\cdot k_{max})^3\right)\\
	&\stackrel{4\cdot7^3\leq12^3}{\leq}2\cdot(z\cdot L+2)\cdot W\cdot \log_2\left((12\cdot z\cdot L\cdot d_1\cdot d_2\cdot k_{max})^3\right)\\
	&~~\stackrel{\eqref{eqW}}{\leq}12\cdot (z\cdot L+2)^2\cdot M_{max}^2\cdot k_{max}^2\cdot\log_2\left(12\cdot z\cdot L\cdot d_1\cdot d_2\cdot k_{max}\right)\\
	&~~~\leq c_{12}\cdot z^2\cdot \log_2(z\cdot d_1\cdot d_2),
\end{align*}
for some constant $c_{12}>0$ which only depends on $L$, $k_{max}$ and $M_{max}$. In the third row we used equation \eqref{eqW} for the total number of weights $W$. Now we make use of Lemma \ref{le13} and finally get
\[V_{\F^+}\leq c_{12}\cdot z^2\cdot \log_2(z\cdot d_1\cdot d_2).\]	
\hfill $\Box$
\begin{lemma}
	\label{le18}
	Suppose that $2^m\leq2^l\cdot(m\cdot R/w)^w$ for some $R\geq16$ and $m\geq w\geq l\geq0$. Then,
	\[
	m\leq l+w\cdot\log_2(2\cdot R\cdot\log_2(R)).
	\]
\end{lemma}
\noindent
{\bf Proof.}
See Lemma 16 in Bartlett et al. (2019).
\hfill $\Box$
~\\~\\
\noindent
{\bf Proof of Lemma \ref{le9}.} 
Using Lemma \ref{le17} and
\[
V_{T_{c_{4} \cdot \log n} \F_3(\btheta)^+}
\leq
V_{\F_3(\btheta)^+},
\]
we can conclude from this together with Lemma 9.2 and Theorem 9.4
in Gy\"orfi et al. (2002)
\begin{eqnarray*}
	&&
	\mathcal{N}_1 \left(\epsilon,   T_{c_{4} \cdot \log n} \F_3(\btheta),
	\bx_1^n\right)
	\\
	&&
	\leq
	3 \cdot \left(
	\frac{4 e \cdot c_{4} \cdot \log n}{\epsilon}
	\cdot
	\log
	\frac{6 e \cdot c_{4} \cdot \log n}{\epsilon}
	\right)^{V_{T_{c_{4} \cdot \log n} \F_3^+}}
	\\
	&&
	\leq
	3 \cdot \left(
	\frac{6 e \cdot c_{4} \cdot \log n}{\epsilon}
	\right)^{
		2 \cdot
		c_{12} \cdot z^2 \cdot \log (z \cdot d_1 \cdot d_2)
	}
	.
\end{eqnarray*}
This completes 
the proof of Lemma \ref{le9}.
\hfill $\Box$
 \section{Auxiliary results}
 In the following section, we present some results from the literature which we have used in the proof of Theorem \ref{th1}.
 Our first auxiliary result relates
 the misclassification error of our plug-in estimate
 to the $L_2$ error of the corresponding least squares estimates.
 
 \begin{lemma}
 	\label{le10}
 	Let $j \in \{1,2,3\}$ and
 	define $(X,Y)$, $(X_1,Y_1)$, \dots, $(X_n,Y_n)$, and $\D_n$,
 	$\eta$, $f^*$ and $f_n^{(j)}$ as in Section \ref{se1}.
 	Then
 	\begin{eqnarray*}
 		\PROB\{f_n^{(j)} (X) \neq Y|\D_n\}
 		-
 		\PROB\{ f^*(X) \neq Y\}
 		&\leq&
 		2 \cdot
 		\int |\eta_n^{(j)}(x)-\eta(x)| \, \PROB_X(dx)
 		\\
 		&\leq&
 		2 \cdot
 		\sqrt{
 			\int |\eta_n^{(j)}(x)-\eta(x)|^2  \PROB_X(dx)
 		}
 	\end{eqnarray*}
 	holds.
 \end{lemma}
 
 \noindent
 {\bf Proof.}
 See Theorem 1.1 in \cite{Gyoerfi2002}.
 \hfill $\Box$
 
 Our next result
 bound the error of the least squares estimate
 via empirical process theory.

 \begin{lemma}
 	\label{le11}
 	Let
 	$(X,Y)$, $(X_1,Y_1)$, \dots, $(X_n,Y_n)$
 	be independent and identically distributed $\R^d \times \R$-valued
 	random variables.
 	Assume that the distribution of $(X,Y)$ satisfies
 	\begin{align*}
 		\E\{\exp(c_{13} \cdot Y^2)\} < \infty
 	\end{align*}
 	for some constant $c_{13} > 0$ and that the regression function
 	$m(\cdot)=\EXP\{ Y |X=\cdot \}$
 	is bounded in absolute value. Let $\tilde{m}_n$ be the least squares estimate
 	\begin{align*}
 		\tilde{m}_n(\cdot) = \arg \min_{f \in \mathcal{F}_n} \frac{1}{n} \sum_{i=1}^n |Y_i - f(X_i)|^2
 	\end{align*}
 	based on some function space $\mathcal{F}_n$
 	consisting of functions $f:\R^d \rightarrow \R$
 	and set $m_n = T_{c_{4} \cdot \log(n)} \tilde{m}_n$ for some constant
 	$c_{4} > 0$.
 	Then $m_n$ satisfies
 	\begin{align*}
 		& \mathbf E \int |m_n(x) - m(x)|^2 {\PROB}_X (dx)\notag\\
 		&\leq \frac{c_{14} \cdot (\log(n))^2 \cdot \sup_{x_1^n \in (\R^d)^n} \left(\log\left(
 			\mathcal{N}_1 \left(\frac{1}{n\cdot c_{4} \log(n)},  T_{c_{4} \log(n)} \mathcal{F}_n, x_1^n\right)
 			\right)+1\right)}{n}\notag\\
 		&\quad + 2 \cdot \inf_{f \in \mathcal{F}_n} \int |f(x)-m(x)|^2 {\PROB}_X (dx)
 	\end{align*}
 	for $n > 1$ and some constant $c_{14} > 0$, which does not depend on
 	$n$ or the parameters of the estimate.
 \end{lemma}
 
 \noindent
 {\bf Proof.}
 This result follows in a straightforward way from the proof of Theorem 1 in
 \cite{Bagirov2009}. A complete proof can be found in the supplement of Bauer and Kohler (2019).
 \hfill $\Box$
 
 Our next auxiliary result is an approximation result for
 $(p,C)$--smooth
 functions by very deep feedforward neural networks.
 \begin{lemma}
 	\label{le12}   Let $d \in \N$,
 	let $f:\Rd \rightarrow \R$ be $(p,C)$--smooth for some $p=q+s$,
 	$q \in \N_0$  and $s \in (0,1]$, and $C>0$. Let $M \in \N$ with $M>1$ sufficiently large, where 
 	\[M^{2p}\geq
 	c_{15}\cdot\left(\max\left\{2,\sup_{\substack{\bx\in[-2,2]^d \\
 			(l_1,\dots,l_d)\in\N^d \\ l_1+\dots+l_d\leq q}
 	}\left|\frac{\partial^{l_1+\dots +l_d}f}{\partial^{l_1}x^{(1)}\dots\partial^{l_d}x^{(d)}}(\bx)\right|\right\}\right)^{4(q+1)}
 	\]
 	must hold for some sufficiently large constant $c_{15}\geq1$.
 	Let $L,r\in\N$ such that
 	\begin{enumerate}[label=(\roman*)]
 		\item 
 		\begin{align*}
 			L\geq&5M^d+\left\lceil \log_{4}\left(M^{2p+4\cdot d\cdot(q+1)}\cdot e^{4\dot(q+1)\cdot(M^d-1)}\right) \right\rceil\\
 			&\cdot  \lceil \log_2(\max\{d,q\}+2)\rceil
 			+\lceil\log_4(M^{2p})\rceil
 		\end{align*}
 		\item
 		\[
 		r\geq132\cdot2^d \cdot \lceil e^{d}\rceil\cdot\binom{d+q}{d} \cdot\max\{q+1, d^2\}
 		\]
 	\end{enumerate}
 	hold.
 	Then there exists a feedforward neural network 
 	\[f_{net}\in\G_d(L,\bk)\]
 	with $\bk=(k_1,\dots,k_L)$ and $k_1=\dots=k_L=r$
 	such that
 	\begin{align*}
 		&\sup_{\bx \in [-2,2]^d} | f(\bx)-f_{net}(\bx)|\\
 		&\leq
 		c_{16}
 		\cdot\left(\max\left\{2,\sup_{\substack{\bx\in[-2,2]^d \\
 				(l_1,\dots,l_d)\in\N^d \\ l_1+\dots+l_d\leq q}
 		}\left|\frac{\partial^{l_1+\dots +l_d}f}{\partial^{l_1}x^{(1)}\dots\partial^{l_d}x^{(d)}}(\bx)\right|\right\}\right)^{4(q+1)}
 		\cdot M^{-2p}.
 	\end{align*}
 \end{lemma}
 
 \noindent
 {\bf Proof.}
 See Theorem 2 b) in Kohler and Langer (2019).
 An alternative proof of a closely related result can be found in
 \cite{Yarotsky2019}, see Theorem 4.1 therein.
 \hspace*{1cm} \hfill $\Box$
\end{document}